%% file: icml2017.tex
\documentclass{article}

\usepackage{times}
\usepackage{graphicx} 
\usepackage{subfigure} 
\usepackage{dsfont} 
\usepackage{bm} 
\usepackage{rotating} 
\usepackage{multirow}
\usepackage{a4wide}

\usepackage{natbib}

\usepackage{algorithm}
\usepackage{algorithmic}
\usepackage{amsmath}

\usepackage{hyperref}

\title{Scalable Multi-Class Gaussian Process Classification using Expectation Propagation}

\author{
Carlos Villacampa-Calvo\thanks{Equal Contributors}\\
Universidad Aut\'onoma de Madrid\\
Francisco Tom\'as y Valiente 11\\
28049, Madrid, Spain\\
\texttt{carlos.villacampa@uam.es} \\
\and
Daniel Hern\'andez-Lobato$^*$\\
Universidad Aut\'onoma de Madrid\\
Francisco Tom\'as y Valiente 11\\
28049, Madrid, Spain\\
\texttt{daniel.hernandez@uam.es} \\
}

\begin{document} 
	
\maketitle

	
	
	
	\begin{abstract} 
		This paper describes an expectation propagation (EP) method
		for multi-class classification with Gaussian processes that scales well 
		to very large datasets. In such a method the estimate of the log-marginal-likelihood 
		involves a sum across the data instances. This enables efficient training using 
		stochastic gradients and mini-batches. When this type of training is used,
		the computational cost does not depend on the number of data instances $N$. 
		Furthermore, extra assumptions in the approximate inference process make the 
		memory cost independent of $N$. The consequence is that the proposed EP method
		can be used on datasets with millions of instances. We compare 
		empirically this method with alternative approaches that approximate the 
		required computations using variational inference. The results show that 
		it performs similar or even better than these techniques, 
		which sometimes give significantly worse predictive distributions in terms of the 
		test log-likelihood. Besides this, the training process of the proposed approach 
		also seems to converge in a smaller number of iterations.
	\end{abstract}
	
	\section{Introduction} 
	
	
	Gaussian processes (GPs) are non-parametric models that can be used to 
	address multi-class classification problems \citep{rasmussen2005book}. 
	These models become more expressive as the number of data 
	instances $N$ grows. They are also very useful to introduce prior knowledge 
	in the learning problem, as many properties of the model are specified by a 
	covariance function. Moreover, GPs provide an estimate of the uncertainty 
	in the predictions made which may be critical in some applications. Nevertheless, in spite of these 
	advantages, GPs scale poorly to large datasets because their training cost is $\mathcal{O}(N^3)$, where 
	$N$ is the number of instances. An additional challenge is that exact inference in these models
	is generally intractable and one has to resort to approximate methods in practice.
	
	Traditionally, GP classification has received more attention in 
	the binary case 
	than in the multi-class setting \citep{kuss2005,nickish2008}. 
	The reason is that approximate inference is more challenging in the multi-class
	case where there is one latent function per class. To this one has to add 
	more complicated likelihood factors, which often have the form of softmax functions 
	or intractable Gaussian integrals. In spite of these difficulties, there have been several 
	works addressing multi-class GP classification 
	\citep{williams1998,Kim2006,girolami2006,chai2012variational,riihimaki2013nested}. 
	Nevertheless, most of the proposed methods do not scale well with the size of the training set.
	
	In the literature there have been some efforts to scale up GPs. These techniques often introduce
	a set of $M \ll N$ inducing points whose location is learnt alongside with the other model hyper-parameters.
	The use of inducing points in the model can be understood as an approximate GP prior with a low-rank 
	covariance structure \citep{quinonero2005unifying}. When inducing points are considered, the training 
	cost can be reduced to $\mathcal{O}(NM^2)$. This allows to address datasets with several thousands 
	of instances, but not millions. The reason is the difficulty of estimating the model
	hyper-parameters, which is often done by maximizing an estimate of the log-marginal-likelihood. Because 
	such an estimate does not involve a sum across the data instances, one cannot rely on efficient 
	methods for optimization based on stochastic gradients and mini-batches.
	
	A notable exception is the work of \citep{HensmanMG15} which uses variational inference to approximate
	the calculations. Such a method allows for stochastic optimization and can address datasets with millions of instances. 
	In this work we propose an alternative based on expectation propagation (EP) \citep{Minka01} and recent advances on 
	binary GP classification \citep{lobato2016}. The proposed approach also allows for efficient training
	using mini-batches. This leads to a training cost that is $\mathcal{O}(CM^3)$, where $C$ is 
	the number of classes. An experimental comparison with the variational approach and related methods
	from the literature shows that the proposed approach has benefits both in 
	terms of the training speed and the accuracy of the predictive distribution.
	
	\section{Scalable Multi-class Classification} 
	\label{sec:mgpc}
	
	Here we describe multi-class Gaussian process classification and
	the proposed method. Such a method uses the expectation 
	propagation algorithm whose original description is modified 
	to be more efficient both in terms of memory and computational costs. 
	For this, we  consider stochastic gradients to update the hyper-parameters 
	and an approximate likelihood that avoids one-dimensional quadratures. 
	
	\subsection{Multi-class Gaussian Process Classification}
	\label{subsec:multiclass_gps}
	
	We consider a dataset of $N$ instances in the form of a matrix of 
	attributes $\mathbf{X}=(\mathbf{x}_1,\ldots,\mathbf{x}_N)^\text{T}$ with 
	labels $\mathbf{y}=(y_1,\ldots,y_N)^\text{T}$, where $y_i\in \{1,\ldots,C\}$ and
	$C > 2$ is the total number of different classes.
	The task of interest is to predict the class label of a new data 
	instance $\mathbf{x}_\star$. 
	
	A typical approach in multi-class Gaussian process (GP)
	classification is to assume the following labeling rule for $y_i$ given
	$\mathbf{x}_i$: $y_i = \text{arg max}_k \quad f^k(\mathbf{x}_i)$, for 
	$k=1,\ldots,C$, where each $f^k(\cdot)$ is a non-linear latent function
	\citep{Kim2006}.
	Define $\mathbf{f}^k=(f^k(\mathbf{x}_1),\ldots,f^k(\mathbf{x}_N))^\text{T}\in\mathds{R}^N$
	and $\mathbf{f}_i=(f^1(\mathbf{x}_i),\ldots,f^C(\mathbf{x}_i))^\text{T} \in \mathds{R}^C$.
	The likelihood of $\mathbf{f}=(\mathbf{f}^1,\ldots,\mathbf{f}^C)^\text{T} \in \mathds{R}^{N\times C}$, 
	$p(\mathbf{y}|\mathbf{f})=\prod_{i=1}^N p(y_i|\mathbf{f}_i)$, is
	then a product of $N$ factors of the form:
	\begin{align}
	p(y_i|\mathbf{f}_i) = \prod_{k \neq y_i} \Theta\left(f^{y_i}(\mathbf{x}_i) - f^k(\mathbf{x}_i)\right)\,,
	\label{eq:likelihood}
	\end{align}
	where $\Theta(\cdot)$ is the Heaviside step function. This likelihood takes value 
	one if $\mathbf{f}$ can explain the observed data and zero otherwise.
	Potential classification errors can be easily introduced in (\ref{eq:likelihood}) by considering that each 
	$f^k$ has been contaminated with Gaussian noise with variance $\sigma_k^2$. That is, 
	$f^k(\mathbf{x}_i) = \hat{f}^k(\mathbf{x}_i) + \epsilon_i^k$, where $\epsilon^k_i \sim \mathcal{N}(0,\sigma^2_k)$. 
	
	In multi-class GP classification a GP prior is assumed for each function
	$f^k(\cdot)$ \citep{rasmussen2005book}. Namely, $f^k\sim \mathcal{GP}(0,c(\cdot,\cdot;\xi))$, 
	where $c(\cdot,\cdot;\xi^k)$ is some covariance function with hyper-parameters $\xi^k$.
	Often these priors are assumed to be independent. That is, $p(\mathbf{f})=\prod_{k=1}^C p(\mathbf{f}^k)$,
	where each $p(\mathbf{f}^k)$ is a multivariate Gaussian distribution.
	The task of interest is to make inference about $\mathbf{f}$ and for that Bayes'
	rule is used: $p(\mathbf{f}|\mathbf{y})=p(\mathbf{y}|\mathbf{f})p(\mathbf{f}) / p(\mathbf{y})$,
	where $p(\mathbf{y})$ is a normalization constant (the marginal likelihood) which can be maximized
	to find good hyper-parameters $\xi^k$, for $k=1,\ldots,C$. 
	However, because the likelihood in (\ref{eq:likelihood}) is non-Gaussian, evaluating $p(\mathbf{y})$
	and $p(\mathbf{f}|\mathbf{y})$ is intractable. Thus, these computations
	must be approximated. Often, one computes a Gaussian approximation to $p(\mathbf{f}|\mathbf{y})$ \citep{Kim2006}. 
	This results in a non-parametric classifier with training cost $\mathcal{O}(N^3)$, where $N$ is the 
	number of data instances.

	To reduce the computational cost of the method described a typical approach 
	is to consider a sparse representation for each GP.
	With this goal, one can introduce $C$ datasets of $M \ll N$ inducting points 
	$\overline{\mathbf{X}}^k=(\overline{\mathbf{x}}_1,\ldots,\overline{\mathbf{x}}_M^k)^\text{T}$, 
	with associated values $\overline{\mathbf{f}}^k=(f^k(\overline{\mathbf{x}}_1^k),\ldots,f^k(\overline{\mathbf{x}}_M^k))^\text{T}$
	for $k=1,\ldots,C$ \citep{Snelson2006,NaishGuzman2008}.
	Given each $\overline{\mathbf{X}}^k$ the prior for $\mathbf{f}^k$ is approximated as
	$p(\mathbf{f}^k) = \int p(\mathbf{f}^k|\overline{\mathbf{f}}^k) p(\overline{\mathbf{f}}^k|\overline{\mathbf{X}}^k) d \overline{\mathbf{f}}^k
	\approx \int [ \prod_{i=1}^N \allowbreak p(f_i^k(\mathbf{x}_i)|\overline{\mathbf{f}}^k)] 
	p(\overline{\mathbf{f}}^k|\overline{\mathbf{X}}^k) d \overline{\mathbf{f}}^k=p_\text{FITC}(\mathbf{f}^k|\overline{\mathbf{X}}^k)$, 
	in which the conditional Gaussian distribution
	$p(\mathbf{f}^k|\overline{\mathbf{f}}^k)$ has been approximated by the factorizing distribution 
	$\prod_{i=1}^N p(f_i^k(\mathbf{x}_i)|\overline{\mathbf{f}}^k)$.
	This approximation is known as the full independent training conditional
	(FITC) \citep{quinonero2005unifying}, and it leads to a Gaussian prior $p_\text{FITC}(\mathbf{f}^k|\overline{\mathbf{X}}^k)$ 
	with a low-rank covariance matrix. This allows for approximate inference with cost $\mathcal{O}(NM^2)$.
	The inducing points $\{\overline{\mathbf{X}}^k\}_{k=1}^C$ can be regarded as hyper-parameters and can be learnt
	by maximizing the estimate of the marginal likelihood $p(\mathbf{y})$.
	
	\subsection{Method Specification and Expectation Propagation}
	\label{sec:prop_method}
	
	The formulation of the previous section is limited because
	the estimate of the log-marginal-likelihood $\log p(\mathbf{y})$ cannot be expressed 
	as a sum across the data instances. This makes infeasible the use of efficient methods
	based on stochastic optimization for finding the model hyper-parameters. 
	
	A recent work focusing on the binary case has shown that it is possible to obtain an estimate 
	of $\log p(\mathbf{y})$ that involves a sum across the data instances if the values $\overline{\mathbf{f}}^k$ 
	associated to the inducing points are not marginalized \citep{lobato2016}.
	We follow that work and consider the posterior approximation 
	$p(\mathbf{f}|\mathbf{y}) \approx \int p(\mathbf{f}|\overline{\mathbf{f}}) q(\overline{\mathbf{f}}) d \overline{\mathbf{f}}$,
	where 
	$\overline{\mathbf{f}}=(\overline{\mathbf{f}}^1,\ldots,\overline{\mathbf{f}}^C)^\text{T}$,
	$p(\mathbf{f}|\overline{\mathbf{f}})=\prod_{k=1}^C p(\mathbf{f}^k|\overline{\mathbf{f}}^k)$,
	we have defined $p(\overline{\mathbf{f}}) = \prod_{k=1}^C p(\overline{\mathbf{f}}^k|\overline{\mathbf{X}}^k)$,
	and $q$ is a Gaussian approximation to $p(\overline{\mathbf{f}}|\mathbf{y})$.
	This distribution $q$ is obtained in three steps. First, we use on the exact posterior the FITC approximation:
	\begin{align}
	p(\overline{\mathbf{f}}|\mathbf{y}) & = 
	\frac{\int p(\mathbf{y}|\mathbf{f}) p(\mathbf{f}|\overline{\mathbf{f}}) d \mathbf{f} p(\overline{\mathbf{f}})}{p(\mathbf{y})}
	\nonumber \\
	& \approx 
	\frac{\int p(\mathbf{y}|\mathbf{f}) p_\text{FITC}(\mathbf{f}|\overline{\mathbf{f}}) d \mathbf{f} p(\overline{\mathbf{f}})}{p(\mathbf{y})}
	\nonumber \\
	& = 
	\frac{ [\prod_{i=1}^N \phi_i(\overline{\mathbf{f}})] p(\overline{\mathbf{f}})}{p(\mathbf{y})}
	\,,
	\label{eq:posterior}
	\end{align}
	where we have defined 
	$p_\text{FITC}(\mathbf{f}|\overline{\mathbf{f}})=\allowbreak \prod_{i=1}^N \allowbreak \prod_{k = 1}^C \allowbreak 
	p(f^k(\mathbf{x}_i) \allowbreak|\overline{\mathbf{f}}^k)
	\approx p(\mathbf{f}|\overline{\mathbf{f}}) = \prod_{k = 1}^C p(\mathbf{f}^k|\overline{\mathbf{f}}^k)$
	and 
	\begin{align}
	\phi_i(\overline{\mathbf{f}}) & = 
	\textstyle 	\int [\prod_{k\neq y_i} \Theta
	\left(f^{y_i}(\mathbf{x}_i) - f^k(\mathbf{x}_i)\right) ]
	\nonumber \\
	& \quad 
	\times
	\textstyle [\prod_{k = 1}^C p(f^k(\mathbf{x}_i)|\overline{\mathbf{f}}^k)] d \mathbf{f}_i
	\,,
	\label{eq:exact_factor_likelihood}
	\end{align}
	with $p(f^k(\mathbf{x}_i)|\overline{\mathbf{f}}^k) = \mathcal{N}(f^k(\mathbf{x}_i)|m_i^k, v_i^k)$,
	where 
	\begin{align}
	m_i^k & = \textstyle (\mathbf{k}_{\mathbf{x}_i\overline{\mathbf{X}}^k}^k)^\text{T} (\mathbf{K}_{\overline{\mathbf{X}}^k 
		\overline{\mathbf{X}}^k}^k)^{-1} \overline{\mathbf{f}}^k
	\,,
	\\
	s_i^k & = \textstyle \kappa_{\mathbf{x}_i \mathbf{x}_i}^k -
	(\mathbf{k}_{\mathbf{x}_i\overline{\mathbf{X}}^k}^k)^\text{T} (\mathbf{K}_{\overline{\mathbf{X}}^k 
		\overline{\mathbf{X}}^k}^k)^{-1} \mathbf{k}_{\mathbf{x}_i\overline{\mathbf{X}}^k}^k
	\,.
	\label{eq:means_and_variances}
	\end{align}
	In the previous expressions $\mathcal{N}(\cdot|\mu,\sigma^2)$ is the p.d.f. of a Gaussian 
	with mean $\mu$ and variance $\sigma^2$. Furthermore, {\scriptsize $\mathbf{k}_{\mathbf{x}_i\overline{\mathbf{X}}^k}^k$} is a 
	vector with the covariances between $f^k(\mathbf{x}_i)$ and {\small $\overline{\mathbf{f}}^k$};
	{\scriptsize $\mathbf{K}_{\overline{\mathbf{X}}^k \overline{\mathbf{X}}^k}^k$} is a $M \times M$ matrix with the cross covariances between 
	{\small $\overline{\mathbf{f}}^k$}; and, finally, $\kappa_{\mathbf{x}_i \mathbf{x}_i}^k$ is the prior variance of $f^k(\mathbf{x}_i)$.
	
	A practical difficulty is that the integral in (\ref{eq:exact_factor_likelihood}) is intractable.
	Although it can be evaluated using one-dimensional quadrature techniques \citep{lobato2011}, in this paper we 
	follow a different approach. For that, we note that (\ref{eq:exact_factor_likelihood}) is simply 
	the probability that $f^{y_i}(\mathbf{x}_i) > f^k(\mathbf{x}_i)$ for $k \neq y_i$, given 
	$\overline{\mathbf{f}}$. 
	Let $f^{y_i}_i = f^{y_i}(\mathbf{x}_i)$ and $f^k_i = f^k(\mathbf{x}_i)$.
	The second step consists in approximating (\ref{eq:exact_factor_likelihood}) as follows:
	{\small
		\begin{align}
		\textstyle p(\bigcap_{k \neq y_i} f^{y_i} > f^k) 
		= & p(f^{y_i} > f^1|\mathcal{S}_1) \times p(f^{y_i} > f^2|\mathcal{S}_2) \times \nonumber \\
		\cdots \times p(f^{y_i} > f^{y_i-1} & |\mathcal{S}_{y_i-1}) 
		\times p(f^{y_i} > f^{y_i+1}|\mathcal{S}_{y_i+1})  \times 
		\nonumber \\
		\cdots 
		\approx 
		\textstyle \prod_{k \neq y_i} & p(f^{y_i} > f^k) = \textstyle \prod_{k \neq y_i} \Phi(\alpha_i^k)
		\,,
		\label{eq:approx_original_factor}
		\end{align}
	}where $\mathcal{S}_j=\bigcap_{k \notin \{1,\ldots,j\}\cup\{y_i\}} f^{y_i} > f^k$,
	$\Phi(\cdot)$ is the c.d.f. of a standard Gaussian and {\small $\alpha_i^k = (m^{yi}_i - m^k_i) / \sqrt{s^{yi}_i + s^k_i}$},
	with $m^{yi}_i$, $m_i^k$, $s_i^{y_i}$ and $s_i^k$ defined in (\ref{eq:means_and_variances}).
	We have omitted in (\ref{eq:approx_original_factor}) the dependence on $\overline{\mathbf{f}}$ to improve the readability.
	The quality of this approximation is supported by the good experimental results obtained in Section \ref{sec:experiments}.
	When (\ref{eq:approx_original_factor}) is replaced in (\ref{eq:posterior}) we get an approximate posterior distribution
	in which we can evaluate all the likelihood factors:
	\begin{align}
	p(\overline{\mathbf{f}}|\mathbf{y}) & \approx
	\frac{ [\prod_{i=1}^N \prod_{k\neq y_k} \phi_i^k(\overline{\mathbf{f}})] p(\overline{\mathbf{f}})}{p(\mathbf{y})}
	\,,
	\label{eq:approx_posterior}
	\end{align}
	where we have defined $\phi_i^k(\overline{\mathbf{f}})=\Phi(\alpha_i^k)$. 
	
	The r.h.s. of (\ref{eq:approx_posterior}) is intractable due to the non-Gaussian form of the
	likelihood factors.  The third and last step uses expectation propagation (EP) \citep{Minka01} to 
	get a Gaussian approximation $q$ to (\ref{eq:approx_posterior}).  This approximation is obtained by replacing 
	each $\phi_i^k$ with an approximate Gaussian factor $\tilde{\phi}_i^k$:
	\begin{align}
	\textstyle \tilde{\phi}_i^k(\overline{\mathbf{f}}) & = \textstyle \tilde{s}_{i,k} \exp
	\left\{ 
	- \frac{1}{2} (\overline{\mathbf{f}}^{y_i})^\text{T} 
	\tilde{\mathbf{V}}_{i,k}^{y_i}
	\overline{\mathbf{f}}^{y_i} +  (\overline{\mathbf{f}}^{y_i})^\text{T} \tilde{\mathbf{m}}_{i,k}^{y_i}
	\right\} \times \nonumber \\
	& \quad 
	\textstyle
	\exp 
	\left\{
	- \frac{1}{2}(\overline{\mathbf{f}}^{k})^\text{T} 
	\tilde{\mathbf{V}}_{i,k}
	\overline{\mathbf{f}}^{k} +  (\overline{\mathbf{f}}^{k})^\text{T} \tilde{\mathbf{m}}_{i,k}
	\right\} 
	\,,
	\label{eq:approx_factor_ep}
	\end{align}
	where $\tilde{\mathbf{V}}_{i,k}^{y_i} = C_{i,k}^{1,y_i} \bm{\upsilon}_{i}^{y_i} (\bm{\upsilon}_{i}^{y_i})^\text{T}$,
	$\tilde{\mathbf{m}}_{i,k}^{y_i} = C_{i,k}^{2,y_i} \bm{\upsilon}_{i}^{y_i} $, 
	$\tilde{\mathbf{V}}_{i,k} = C_{i,k}^{1} \bm{\upsilon}_{i}^{k} (\bm{\upsilon}_{i}^{k})^\text{T}$,
	$\tilde{\mathbf{m}}_{i,k} = C_{i,k}^{2} \bm{\upsilon}_{i}^{k}$, and we have
	defined  {\small $\bm{\upsilon}_i^k = (\mathbf{k}_{\mathbf{x}_i\overline{\mathbf{X}}^k}^k)^\text{T} 
		(\mathbf{K}_{\overline{\mathbf{X}}^k \overline{\mathbf{X}}^k}^k)^{-1} $}. In
	(\ref{eq:approx_factor_ep}) $C_{i,k}^{1,y_i}$, $C_{i,k}^{2,y_i}$, 
	$C_{i,k}^{1}$, $C_{i,k}^{2}$ and $\tilde{s}_{i,k}$ are free parameters adjusted by EP.
	Because the precision matrices in (\ref{eq:approx_factor_ep}) are one-rank (see the supplementary material for details), 
	we only have to store in memory $\mathcal{O}(M)$ parameters for each $\tilde{\phi}_i^k$. The posterior approximation $q$
	is obtained by replacing in (\ref{eq:approx_posterior}) each exact factor $\phi_{i,k}$ by the corresponding approximate factor $\tilde{\phi}_{i,k}$.
	That is, $q(\overline{\mathbf{f}}) = \prod_{i=1}^N \prod_{k \neq y_i} \tilde{\phi}_i^k(\overline{\mathbf{f}}) p (\overline{\mathbf{f}}) / Z_q$,
	where $Z_q$ is a normalization constant that approximates the marginal likelihood $p(\mathbf{y})$.
	Because all the factors involved in the computation of $q$ are Gaussian, and we assume independence 
	among the latent functions of different classes in (\ref{eq:approx_factor_ep}), $q$ is a product of 
	$C$ multivariate Gaussians (on per class) on $M$ dimensions.
	
	In EP each $\tilde{\phi}_i^k$ is updated until convergence as follows:
	First, $\phi_i^k$ is removed from $q$ by computing $q^{\setminus i,k}\propto
	q / \tilde{\phi}_i^k$. Because the Gaussian family is closed under the product and division operations, 
	$q^{\setminus i,k}$ is also Gaussian with parameters given by the equations in \citep{Roweis99}.
	Then, the Kullback-Leibler divergence between $Z_{i,k}^{-1} \phi_i^k 
	q^{\setminus i,k}$ and $q$, \emph{i.e}, $\text{KL}[Z_{i,k}^{-1} \phi_i^k q^{\setminus i,k}|q]$, 
	is minimized with respect to $q$, where $Z_{i,k}$ is the normalization constant of
	$\phi_i^k q^{\setminus i,k}$. This is done by matching the moments of $Z_{i,k}^{-1} \phi_i^k q^{\setminus i,k}$.
	These moments can be obtained from the derivatives of $Z_{i,k}$ with respect to the parameters
	of $q^{\setminus i,k}$ \citep{matthias2006}. After updating $q$, the new approximate factor is $\tilde{\phi}_{i,k}=Z_{i,k}
	q / q^{\setminus i,k}$. We update all the approximate factors at the same time,
	and reconstruct $q$ afterwards by computing the product of all the $\tilde{\phi}_i^k$ and the prior, as in \citep{lobato2011}.
	
	The EP approximation to the marginal likelihood is the normalization constant of $q$, $Z_q$. 
	The log of its value is:
	\begin{align}
	\log Z_q &= g(\bm{\theta}) - g(\bm{\theta}_\text{prior}) + \textstyle \sum_{i=1}^N \textstyle \sum_{k \neq y_k} \log \tilde{s}_{i,k}\,,
	\label{eq:log_zq}
	\end{align}
	where $\log \tilde{s}_{i,k} = \log Z_{i,k} + g(\bm{\theta}^{\setminus i,k}) - g(\bm{\theta})$; $\bm{\theta}$, 
	$\bm{\theta}^{\setminus i,k}$, and $\bm{\theta}_\text{prior}$ are the natural parameters of
	$q$, $q^{\setminus i,k}$ and the prior, respectively;
	and $g(\bm{\theta})$ is the log-normalizer of a multi-variate Gaussian distribution with natural parameters $\bm{\theta}$.
	
	It is possible to show that if EP converges, the gradient of $\log Z_q$ w.r.t the parameters of each $\tilde{\phi}_{i,k}$
	is zero. Thus, the gradient of $\log Z_q$ w.r.t. a hyper-parameter 
	$\xi_j^k$ of the $k$-th covariance function (including the inducing points) is:
	{\small
		\begin{align}
		\frac{\partial \log Z_q}{\partial \xi_j^k} &= (\bm{\eta}^\text{T}-\bm{\eta}_\text{prior}^\text{T})
		\frac{\partial \bm{\theta}_\text{prior}}{\partial \xi_j^k} + \sum_{i=1}^N \sum_{k \neq y_i} \frac{\log Z_{i,k}}{\partial \xi_j^k}\,,
		\label{eq:gradient}
		\end{align}
	}where $\bm{\eta}$ and $\bm{\eta}_\text{prior}$ are the expected sufficient statistics under $q$ and 
	the prior, respectively. Importantly, only the direct dependency of $\log Z_{i,k}$ on $\xi_j^k$
	has to be taken into account. See \citep{matthias2006}. The dependency through $\bm{\theta}^{\setminus i,k}$,
	\emph{i.e.}, the natural parameters of $q^{\setminus i,k}$ can be ignored.
	
	After obtaining $q$ and finding the model hyper-parameters by maximizing $\log Z_q$, one can get an approximate 
	predictive distribution for the label $y_\star$ of a new instance $\mathbf{x}_\star$:
	\begin{align}
	p(y_\star|\mathbf{x}_\star, \mathbf{y}) &= \textstyle \int p(y_\star|\mathbf{f}_\star,\overline{\mathbf{f}})
	q(\overline{\mathbf{f}}) d \overline{\mathbf{f}} d \mathbf{f}_\star
	\,,
	\label{eq:predictive}
	\end{align}
	where we have defined $\mathbf{f}_\star=(f^1(\mathbf{x}_\star),\ldots,f^C(\mathbf{x}_\star))^\text{T}$,
	and $\int p(y_\star|\mathbf{f}_\star,\overline{\mathbf{f}}) d \mathbf{f}_\star$ has the same form as the likelihood factor
	in (\ref{eq:exact_factor_likelihood}). The resulting integral in (\ref{eq:predictive}) is again intractable. However, 
	it can be approximated using a one-dimensional quadrature. See the supplementary material.
	
	Because some simplifications occur when computing the derivatives of $\log Z_q$ w.r.t the inducing points,
	the total training time of EP is $\mathcal{O}(NM^2)$ while the total memory 
	cost is $\mathcal{O}(NMC)$ \citep{SnelsonThesis07}.
	
	\subsection{Scalable Expectation Propagation}
	
	Traditionally, for finding the model hyper-parameters with EP one re-runs EP until convergence 
	(using the previous solution as the starting point), after each gradient ascent update of the 
	hyper-parameters. The reason for this is that (\ref{eq:gradient}) is only true if EP has converged (\emph{i.e.}, 
	the approximate factors do not change any more). This approach is particularly inefficient initially,
	when there are strong changes to the model hyper-parameters, and EP may require several iterations to 
	converge. Recently, a more efficient method has been proposed in \citep{lobato2016}. In that work
	the authors suggest to update both the approximate factors and the model hyper-parameters at the same time.
	Because we do not wait for EP to converge, one should ideally add to (\ref{eq:gradient}) extra terms to get 
	the gradient. These terms account for the mismatch between the moments of $Z_{i,k}^{-1}\phi_i^k q^{\setminus i,k}$
	and $q$. However, according to \citep{lobato2016} these extra terms can be ignored and one can simply
	use (\ref{eq:gradient}) for an inner update of the hyper-parameters.
	
	\begin{figure}[ht]
		\begin{center}
			\centerline{\includegraphics[width=0.75 \columnwidth]{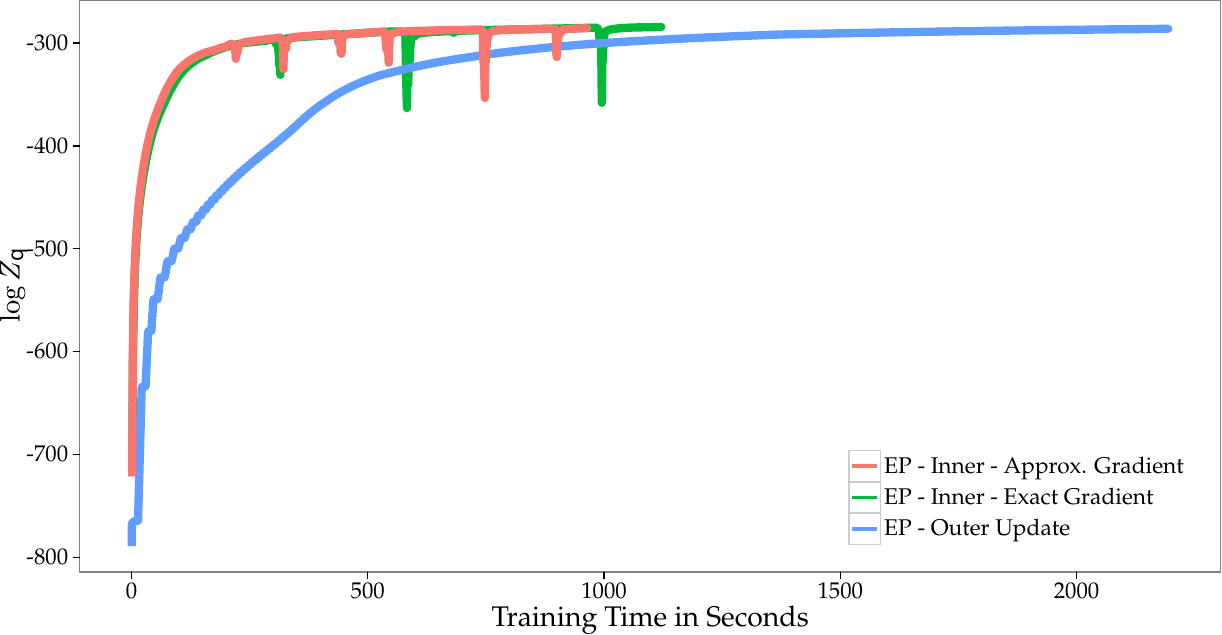}}
			\caption{{\small Estimate of $p(\mathbf{y})$ on the Vehicle dataset as a function of the training 
					time for the proposed EP method when considering three different schemes to update the model hyper-parameters.}} 
			\label{fig:exact_vs_approx_grad}
		\end{center}
	\end{figure}

	Figure \ref{fig:exact_vs_approx_grad} shows, for the Vehicle dataset from UCI repository \citep{Lichman2013}, the 
	estimate of the marginal likelihood $\log Z_q$ with respect to the training time, for 250 updates of the hyper-parameters, and $M = N / 5$. 
	We compare three methods: (i) re-running EP until convergence each time and using (\ref{eq:gradient}) to update 
	the hyper-parameters (EP-outer); (ii) updating at the same time the approximate factors $\tilde{\phi}_i^k$ and the 
	hyper-parameters with (\ref{eq:gradient}) (EP-inner-approx); and (iii) the same approach as the previous one,
	but using the exact gradient for the update instead of (\ref{eq:gradient}) (EP-inner-exact).
	All approaches successfully maximize $\log Z_q$. However, the inner updates are 
	more efficient as they do not wait until EP converges. Moreover, using the approximate gradient 
	is faster (it is cheaper to compute), and it gives almost the same results as the exact gradient.

	\subsubsection{Stochastic Expectation Propagation}
	
\begin{figure}[ht]
		\begin{center}
			\centerline{\includegraphics[width=0.75\columnwidth]{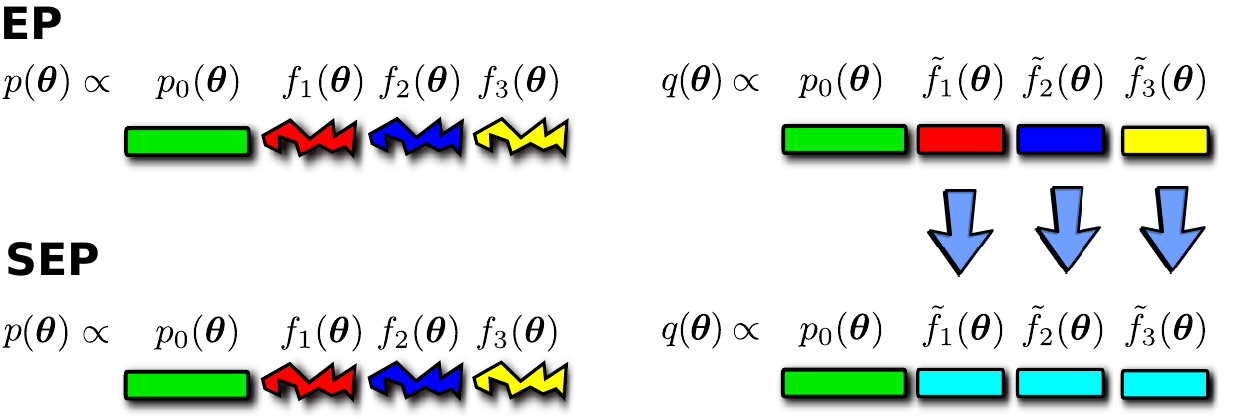}}
			\caption{{\small (top) EP approximation of a distribution over the variable 
					$\theta$ with complicated likelihood factors, but tractable prior. (bottom) SEP 
					approximation of the same distribution.  }
			} 
			\label{fig:sep}
		\end{center}
	\end{figure}

	The memory cost of EP can be significantly reduced by a technique
	called stochastic EP (SEP) \citep{Li2015}. In SEP all the approximate factors $\tilde{\phi}_i^k$
	are tied. This means that instead of storing their individual parameters, what
	is stored is their product, \emph{i.e.}, $\tilde{\phi}=\prod_{i=1}^N \prod_{k\neq y_k} \tilde{\phi}_i^k$. 
	A consequence of this is that we no longer have direct access to their individual parameters. 
	This only affects the computation of the cavity distribution $q^{\setminus i,k}$ which now is obtained
	in an approximate way $q^{\setminus i,k}\propto q / \tilde{\phi}^\frac{1}{n}$, where $n$ is the total
	number of factors and $\tilde{\phi}^{\frac{1}{n}}$ approximates each individual factor. 
	Thus, SEP reduces the memory costs of EP by a factor of $n$.
	All the other steps are carried out as in the original EP algorithm, including the 
	computation of $\log Z_q$ and its gradients.  Figure \ref{fig:sep} shows the differences between 
	EP and SEP on a toy example. When SEP is used in the proposed method, 
	the memory cost is reduced to $\mathcal{O}(CM^2)$.
	
	\subsubsection{Training Using Mini-batches}
	
	Both the estimate of the log-marginal-likelihood in (\ref{eq:log_zq}) and its gradient in (\ref{eq:gradient}) 
	contain a sum across the data instances. This allows to write an EP algorithm that
	processes mini-batches of data, as in \citep{lobato2016}. For this, the data are split in mini-batches $\mathcal{M}_j$ of 
	size $S \ll N$, where $N$ is the number of instances. Given a mini-batch $\mathcal{M}_j$,
	we process all the approximate factors corresponding to that mini-batch, \emph{i.e.}, 
	$\{\{\tilde{\phi}_i^k\}_{k\neq y_i} : (\mathbf{x}_i, y_i) \in \mathcal{M}_j\}$. Then, we update the model 
	hyper-parameters using a stochastic approximation of (\ref{eq:gradient}):
	{\small
		\begin{align}
		\frac{\partial \log Z_q}{\partial \xi_j^k} & \approx (\bm{\eta}^\text{T}-\bm{\eta}_\text{prior}^\text{T})
		\frac{\partial \bm{\theta}_\text{prior}}{\partial \xi_j^k} + \rho 
		\sum_{i\in \mathcal{M}_j} \sum_{k \neq y_i} \frac{\log Z_{i,k}}{\partial \xi_j^k}
		\,,
		\label{eq:gradient_st}
		\end{align}
	}where $\rho=N/|\mathcal{M}_j|$. We reconstruct $q$ after each update of the approximate
	factors and each update of the hyper-parameters. When using mini-batches of data, we update
	more frequently $q$ and the hyper-parameters. The consequence is that the training cost 
	is $\mathcal{O}(CM^3)$, assuming a constant number of updates until convergence.
	This training scheme can handle datasets with millions of instances.
	
	\section{Related Work}
	\label{sec:related}
	
	The likelihood used in (\ref{eq:likelihood}) was first considered for multi-class Gaussian
	process classification in \citep{Kim2006}. That work considers full non-parametric GP
	priors, which lead to a training cost that is $\mathcal{O}(CN^3)$. The consequence
	is that it can only address small classification problems.
	It is, however, straight forward to replace the non-parametric GP priors with the 
	FITC approximate priors $p_\text{FITC}(\mathbf{f}^k|\overline{\mathbf{X}}^k)$ \citep{quinonero2005unifying}.
	These priors are obtained by marginalizing the latent variables $\overline{\mathbf{f}}^k$ associated to 
	the inducing points $\overline{\mathbf{X}}^k$, as indicated in Section \ref{subsec:multiclass_gps}. This allows 
	to address datasets with a few thousand instances. This is precisely the approach followed 
	in \citep{NaishGuzman2008} to address binary GP classification problems. We refer to such an approach
	as the generalized FITC approximation (GFITC). Nevertheless, such an approach cannot use stochastic
	optimization. The reason is that the estimate of the log-marginal-likelihood (needed for 
	hyper-parameter estimation) does not contain a sum across the instances.
	Thus, GFITC cannot scale well to very large datasets. Nevertheless, unlike the proposed 
	approach, it can run expectation propagation over the exact likelihood factors in (\ref{eq:likelihood}).
	In GFITC we follow the traditional approach and run EP until convergence before updating the 
	hyper-parameters.
	
	Multi-class GP classification for potentially huge datasets has also been considered in \citep{hensman2015}
	using variational inference (VI). However, such an approach cannot use the likelihood in (\ref{eq:likelihood}) since its 
	logarithm is not well defined (note that it takes value zero for some values of $\mathbf{f}_i$). As an alternative, 
	\citet{hensman2015} have considered the robust likelihood of \citep{lobato2011}:
	{\small
		\begin{align}
		p(y_i|\mathbf{f}_i) = (1-\epsilon) \textstyle \prod_{k \neq y_i} \Theta\left(f^{y_i}(\mathbf{x}_i) - f^k(\mathbf{x}_i)\right) + \frac{\epsilon}{C}\,,
		\label{eq:likelihood_vi}
		\end{align}
	}where $\epsilon$ is the probability of a labeling error (in that case, $y_i$ is chosen at random from 
	the potential class labels). In \citep{hensman2015} it is suggested to set $\epsilon =10^{-3}$. 
	
	We now describe the VI approach in detail.
	Using (\ref{eq:likelihood_vi}) and the definitions of Section \ref{sec:mgpc}, we know that 
	$p(\mathbf{y}|\overline{\mathbf{f}}) = \int p(\mathbf{y}|\mathbf{f}) p(\mathbf{f}|\overline{\mathbf{f}}) d \mathbf{f}$. 
	If we take the log and use Jensen's inequality we get the bound $\log p(\mathbf{y}|\overline{\mathbf{f}}) 
	\geq \mathds{E}_{p(\mathbf{f}|\overline{\mathbf{f}})}[\log p(\mathbf{y}|\mathbf{f})]$. 
	Consider now a Gaussian approximation $q$ to $p(\overline{\mathbf{f}}|\mathbf{y})$. Then,
	\begin{align}
	\log p(\mathbf{y}) = & \textstyle \int q(\overline{\mathbf{f}}) p(\mathbf{y}|\overline{\mathbf{f}}) p(\overline{\mathbf{f}})
	/ q(\overline{\mathbf{f}}) d \overline{\mathbf{f}}
	\nonumber \\
	& \geq \mathds{E}_{q(\overline{\mathbf{f}})}[\log p(\mathbf{y}|\overline{\mathbf{f}})]  - 
	\text{KL}[q(\overline{\mathbf{f}})||p(\overline{\mathbf{f}})]
	\,,
	\label{eq:bound}
	\end{align}
	where we have used Jensen's inequality and $\text{KL}$ is the Kullback Leibler divergence.
	If we use the first bound we get
	{\small
		\begin{align}
		\log p(\mathbf{y}) & \geq 
		\mathds{E}_{q(\overline{\mathbf{f}})}[\mathds{E}_{p(\mathbf{f}|\overline{\mathbf{f}})} [\log p(\mathbf{y}|\mathbf{f})] ]
		\text{KL}[q(\overline{\mathbf{f}})||p(\overline{\mathbf{f}})]
		\nonumber \\
		& \geq \mathds{E}_{q(\mathbf{f})}[ \log p(\mathbf{y}|\mathbf{f}) ]
		- \text{KL}[q(\overline{\mathbf{f}})||p(\overline{\mathbf{f}})]
		\nonumber \\
		& \geq \textstyle 
		\sum_{i=1}^N \mathds{E}_{q(\mathbf{f}_i)} [\log p(y_i|\mathbf{f}_i)] - \text{KL}[q(\overline{\mathbf{f}})||p(\overline{\mathbf{f}})]
		\,,
		\label{eq:lower_bound_vi}
		\end{align}
	}where $q(\mathbf{f})=\int p(\mathbf{f}|\overline{\mathbf{f}})q(\overline{\mathbf{f}}) d \overline{\mathbf{f}}$  and 
	the corresponding marginal over $\mathbf{f}_i=(f^1(\mathbf{x}_i),\ldots,f^C(\mathbf{x}_i))^\text{T}$
	is $q(\mathbf{f}_i)=\prod_{k=1}^C \mathcal{N}(f^k(\mathbf{x}_i)|\hat{m}^k_i,\hat{s}^k_i)$.
	Note that $q(\mathbf{f}_i)$ is Gaussian because $q(\mathbf{f})$ involves a Gaussian convolution. 
	As in the proposed
	approach, $q(\overline{\mathbf{f}})$ is assumed to be a Gaussian factorizing over the latent 
	functions $\overline{\mathbf{f}}^1,\ldots,\overline{\mathbf{f}}^C$. 
	However, its mean and covariance parameters, \emph{i.e.}, $\{\mathbf{m}^k\}_{k=1}^C$ and $\{\mathbf{S}^k\}_{k=1}^C$
	are found by maximizing (\ref{eq:lower_bound_vi}).
	The parameters of $q(\mathbf{f}_i)$ are:
	{\small
		\begin{align}
		\hat{m}_i^k & = (\mathbf{k}_{\mathbf{x}_i\overline{\mathbf{X}}^k}^k)^\text{T} (\mathbf{K}_{\overline{\mathbf{X}}^k 
			\overline{\mathbf{X}}^k}^k)^{-1} \mathbf{m}^k
		\,,
		\\
		\hat{s}_i^k & = \kappa_{\mathbf{x}_i \mathbf{x}_i}^k -
		(\mathbf{k}_{\mathbf{x}_i\overline{\mathbf{X}}^k}^k)^\text{T} (\mathbf{K}_{\overline{\mathbf{X}}^k 
			\overline{\mathbf{X}}^k}^k)^{-1} \mathbf{k}_{\mathbf{x}_i\overline{\mathbf{X}}^k}^k
		\nonumber \\
		& + (\mathbf{k}_{\mathbf{x}_i\overline{\mathbf{X}}^k}^k)^\text{T} (\mathbf{K}_{\overline{\mathbf{X}}^k 
			\overline{\mathbf{X}}^k}^k)^{-1} \mathbf{S}^k (\mathbf{K}_{\overline{\mathbf{X}}^k 
			\overline{\mathbf{X}}^k}^k)^{-1}
		\mathbf{k}_{\mathbf{x}_i\overline{\mathbf{X}}^k}^k
		\,.
		\end{align} 
	}\citet{hensman2015} consider Markov chain Monte Carlo (MCMC) to sample the hyper-parameters.
	Here we simply maximize (\ref{eq:lower_bound_vi}) to find the hyper-parameters and the 
	inducing points. The reason for this is that in very large datasets MCMC is not expected to give 
	much better results. We refer to the described approach as VI. The objective in (\ref{eq:lower_bound_vi})
	contains a sum across the data instances. Thus, VI also allows for stochastic optimization 
	and it results in the same cost as the proposed approach. However, the expectations in 
	(\ref{eq:lower_bound_vi}) must be approximated using one-dimensional quadratures.
	This is a drawback with respect to the proposed method which is free of any quadrature.
	Finally, there are some methods related to the VI approach just described.
	\citet{dezfouli2015} assume that $q$ can be a mixture of Gaussians, 
	and \citet{chai2012variational} uses a soft-max likelihood (but
	does not consider stochastic optimization). Both works need to introduce extra approximations.
	
	In the literature there are other research works addressing multi-class Gaussian process 
	classification.  Some examples include \citep{williams1998,girolami2006,lobato2011,Henao2012,
		riihimaki2013nested}. These works employ expectation propagation, variational inference
	or the Laplace approximation to approximate the computations. Nevertheless, the corresponding
	estimate of the log-marginal-likelihood cannot be expressed as a sum across the data instances. This
	avoids using efficient techniques for optimization based on stochastic gradients.
	Thus, one cannot address very large datasets with these methods.
	
	\section{Experiments} \label{sec:experiments}
	
	We evaluate the performance of the method proposed in Section \ref{sec:prop_method}.
	We consider two versions of it. A first one, using expectation propagation (EP). A second, 
	using the memory efficient stochastic EP (SEP). EP and SEP are compared with the methods 
	described in Section \ref{sec:related}. Namely, GFITC and VI. All methods are
	codified in the R language (the source code is in the supplementary
	material), and they consider the same initial values for the model hyper-parameters
	(including the inducing points, that are chosen at random from the training instances). The
	hyper-parameters are optimized by maximizing the estimate of the marginal likelihood. A Gaussian
	covariance function with automatic relevance determination, an amplitude parameter and 
	an additive noise parameter is employed.
	
	\subsection{Performance on Datasets from the UCI Repository}
	\label{subsec:uci}
	
	We evaluate the performance of each method on 8 datasets from the UCI repository 
	\citep{Lichman2013}. The characteristics of the datasets are displayed in Table \ref{tab:charac}. 
	We use batch training in each method (\emph{i.e.}, we go through all the data to compute the 
	gradients). Batch training does not scale to large datasets. However, it is preferred on small datasets
	like the ones considered here. We use 90\% of the data for training and 10\% for 
	testing, expect for \emph{Satellite} which is fairly big, where we use 20\% for training and 80\% for 
	testing. In \emph{Waveform}, which is synthetic, we generate 1000 instances and split them in 
	30\% for training and 70\% for testing. Finally, in \emph{Vowel} we consider only 
	the points that belong to the 6 first classes. All methods are trained for
	250 iterations using gradient ascent (GFITC and VI use l-BFGS, EP and SEP
	use an adaptive learning rate described in the supplementary material). We consider three values for $M$, the number of inducing points.
	Namely $5\%$, $10\%$ and $20\%$ of the number of training instances. We report averages over 
	20 repetitions of the experiments. 
	
	\begin{table}[ht]
		\begin{center}
		\label{tab:charac}
		\caption{Characteristics of the datasets from the UCI Repository.}
		{\small
			\begin{tabular}{lccc}
				\hline
				{\bf Dataset} & {\bf \#Instances} & {\bf \#Attributes} & {\bf \#Classes} \\
				\hline
				Glass & 214 & 9 & 6 \\
				New-thyroid & 215 & 5 & 3 \\
				Satellite & 6435 & 36 & 6 \\
				Svmguide2 & 391 & 20 & 3 \\
				Vehicle & 846 & 18 & 4 \\
				Vowel & 540 & 10 & 6 \\
				Waveform & 1000 & 21 & 3 \\
				Wine & 178 & 13 & 3 \\
				\hline
			\end{tabular}
		}
		\end{center}
	\end{table}
	
	Table \ref{tab:ll_uci} shows, for each value of $M$, the average negative test log-likelihood of 
	each method with the corresponding error bars (test errors are shown in the supplementary material). 
	The average training time of each method is also 
	displayed. The best method (the lower the better) for each dataset is highlighted in bold face.
	We observe that the proposed approach, EP, obtains very similar results to those of GFITC, 
	and sometimes it obtains the best results. The memory efficient version of EP, SEP, seems to provide 
	similar results without reducing the performance. Regarding the computational cost, SEP is the fastest method
	(between 2 and 3 times faster than GFITC). VI is slower as a consequence of the quadratures required 
	by this method. VI also gives much worse results in some datasets, 
	\emph{e.g.}, \emph{Glass}, \emph{Svmguide2} and \emph{Waveform}. This is related to the optimization of
	$\mathds{E}_{q(\mathbf{f}_i)}[\log p(y_i|\mathbf{f}_i)]$ in  (\ref{eq:lower_bound_vi}), instead of 
	$\log \mathds{E}_{q(\mathbf{f}_i)}[p(y_i|\mathbf{f}_i)]$, which is closer to the data log-likelihood.
	In the EP objective in (\ref{eq:log_zq}), $\sum_{k \neq y_i} \log Z_{i,k}$ is probably more similar to 
	$\log \mathds{E}_{q(\mathbf{f}_i)}[p(y_i|\mathbf{f}_i)]$. This explains the much better results obtained by EP and SEP.
	
	\begin{table}[htb]
		\begin{center}
			\caption{\small Average negative test log likelihood for each method and average training time in seconds on UCI repository datasets.}
			\label{tab:ll_uci}
			{\small
				\begin{tabular}{@{\hspace{0mm}}l@{\hspace{.5mm}}|@{\hspace{.1mm}}l@{\hspace{.1mm}}
						|@{\hspace{.1mm}}c@{{\tiny $\pm$}}c@{\hspace{.1mm}}
						|@{\hspace{.1mm}}c@{{\tiny $\pm$}}c@{\hspace{.1mm}}
						|@{\hspace{.1mm}}c@{{\tiny $\pm$}}c@{\hspace{.1mm}}
						|@{\hspace{.1mm}}c@{{\tiny $\pm$}}c@{\hspace{.1mm}}
					}
					\hline
					& \bf{Problem}  & \multicolumn{2}{c}{\scriptsize \bf GFITC} & \multicolumn{2}{c}{\bf EP} &\multicolumn{2}{c}{\bf SEP} & \multicolumn{2}{c}{\bf VI}  \\
					\hline
					\parbox[t]{2mm}{\multirow{8}{*}{\rotatebox[origin=c]{90}{{\scriptsize ${\bf M = 5\%}$}}}}
					& Glass                & \bf{ 0.61 } & \bf{ 0.05 } & 0.78 & 0.06 & 0.77 & 0.07 & 2.45 & 0.14           \\
					& New-thyroid          & \bf{ 0.06 } & \bf{ 0.01 } & 0.11 & 0.03 & 0.06 & 0.01 & 0.09 & 0.02           \\
					& Satellite            & 0.33 & 0.00 & \bf{ 0.31 } & \bf{ 0.00 } & 0.33 & 0.00 & 0.61 & 0.01                    \\
					& Svmguide2            & \bf{ 0.63 } & \bf{ 0.06 } & 0.63 & 0.06 & 0.67 & 0.06 & 1.03 & 0.08           \\
					& Vehicle              & \bf{ 0.32 } & \bf{ 0.01 } & 0.34 & 0.02 & 0.34 & 0.02 & 0.76 & 0.05           \\
					& Vowel                & \bf{ 0.16 } & \bf{ 0.01 } & 0.25 & 0.01 & 0.25 & 0.01 & 0.41 & 0.05           \\
					& Waveform             & 0.42 & 0.01 & \bf{ 0.36 } & \bf{ 0.00 } & 0.39 & 0.01 & 0.89 & 0.02              \\
					& Wine                 & 0.08 & 0.02 & \bf{ 0.07 } & \bf{ 0.01 } & 0.08 & 0.01 & 0.08 & 0.02           \\
					\hline
					& {\bf Avg. Time} & 131 & 3.11 & 53.8 & 0.19 & {\bf 48.5} & {\bf 0.97} & 157 & 0.59 \\
					\hline
					\parbox[t]{2mm}{\multirow{8}{*}{\rotatebox[origin=c]{90}{{\scriptsize ${\bf M = 10\%}$}}}}
					& Glass                & \bf{ 0.58 } & \bf{ 0.05 } & 0.74 & 0.06 & 0.79 & 0.07 & 2.18 & 0.14          \\
					& New-thyroid          & 0.07 & 0.01 & 0.06 & 0.01 & 0.06 & 0.01 & \bf{ 0.05 } & \bf{ 0.01 }          \\
					& Satellite            & 0.34 & 0.00 & \bf{ 0.30 } & \bf{ 0.00 } & 0.34 & 0.00 & 0.58 & 0.01                          \\
					& Svmguide2            & \bf{ 0.67 } & \bf{ 0.05 } & 0.67 & 0.05 & 0.74 & 0.07 & 0.90 & 0.10            \\
					& Vehicle              & \bf{ 0.33 } & \bf{ 0.01 } & 0.33 & 0.02 & 0.34 & 0.02 & 0.72 & 0.04          \\
					& Vowel                & \bf{ 0.14 } & \bf{ 0.01 } & 0.19 & 0.01 & 0.19 & 0.01 & 0.30 & 0.04           \\
					& Waveform             & 0.42 & 0.01 & \bf{ 0.36 } & \bf{ 0.01 } & 0.41 & 0.01 & 0.85 & 0.01             \\
					& Wine                 & 0.07 & 0.01 & \bf{ 0.06 } & \bf{ 0.01 } & 0.07 & 0.01 & 0.07 & 0.01          \\
					\hline
					& {\bf Avg. Time} & 264 & 6.91 & 102 & 0.64 & {\bf 96.6} & {\bf 1.99} & 179 & 0.78 \\
					\hline
					\parbox[t]{2mm}{\multirow{8}{*}{\rotatebox[origin=c]{90}{{\scriptsize ${\bf M = 20\%}$}}}}
					& Glass              & \bf{ 0.6 } & \bf{ 0.07 } & 0.75 & 0.06 & 0.81 & 0.07 & 2.30 & 0.15             \\
					& New-thyroid        & 0.07 & 0.01 & 0.06 & 0.01 & \bf{ 0.05 } & \bf{ 0.01 } & 0.05 & 0.01           \\
					& Satellite          & 0.34 & 0.01 & \bf{ 0.30 } & \bf{ 0.00 } & 0.36 & 0.00 & 0.53 & 0.01                        \\
					& Svmguide2          & 0.67 & 0.05 & \bf{ 0.65 } & \bf{ 0.06 } & 0.74 & 0.07 & 0.94 & 0.08           \\
					& Vehicle            & 0.33 & 0.01 & \bf{ 0.33 } & \bf{ 0.02 } & 0.34 & 0.02 & 0.63 & 0.04           \\
					& Vowel              & \bf{ 0.12 } & \bf{ 0.01 } & 0.16 & 0.01 & 0.18 & 0.01 & 0.15 & 0.03           \\
					& Waveform           & 0.43 & 0.01 & \bf{ 0.37 } & \bf{ 0.01 } & 0.45 & 0.01 & 0.80 & 0.01               \\
					& Wine               & 0.07 & 0.01 & \bf{ 0.05 } & \bf{ 0.01 } & 0.06 & 0.01 & 0.06 & 0.02           \\
					\hline
					& {\bf Avg. Time} & 683 & 17.3 & 228 & 0.78 & {\bf 216} & {\bf 2.88} & 248 & 0.66 \\
					\hline
				\end{tabular}
			}
		\end{center}
	\end{table}

	\begin{figure*}[htb]
		\begin{center}
			\begin{tabular}{@{\hspace{1mm}}l@{\hspace{1mm}}c@{\hspace{1mm}}c@{\hspace{1mm}}c@{\hspace{1mm}}c@{\hspace{1mm}}c@{\hspace{1mm}}c@{\hspace{1mm}}c@{\hspace{1mm}}c@{\hspace{1mm}}c}
				& {\scriptsize $M=1$} & {\scriptsize $M = 2$} &{\scriptsize  $M = 4$} & {\scriptsize $M = 8$} & {\scriptsize $M = 16$} & {\scriptsize $M = 32$} & {\scriptsize $M = 64$} & {\scriptsize $M = 128$} & {\scriptsize $M = 256$} \\
				\rotatebox{90}{\hspace{0.40cm}{{\scriptsize {\bf GFITC}}}} &
				\includegraphics[width = 1.5cm]{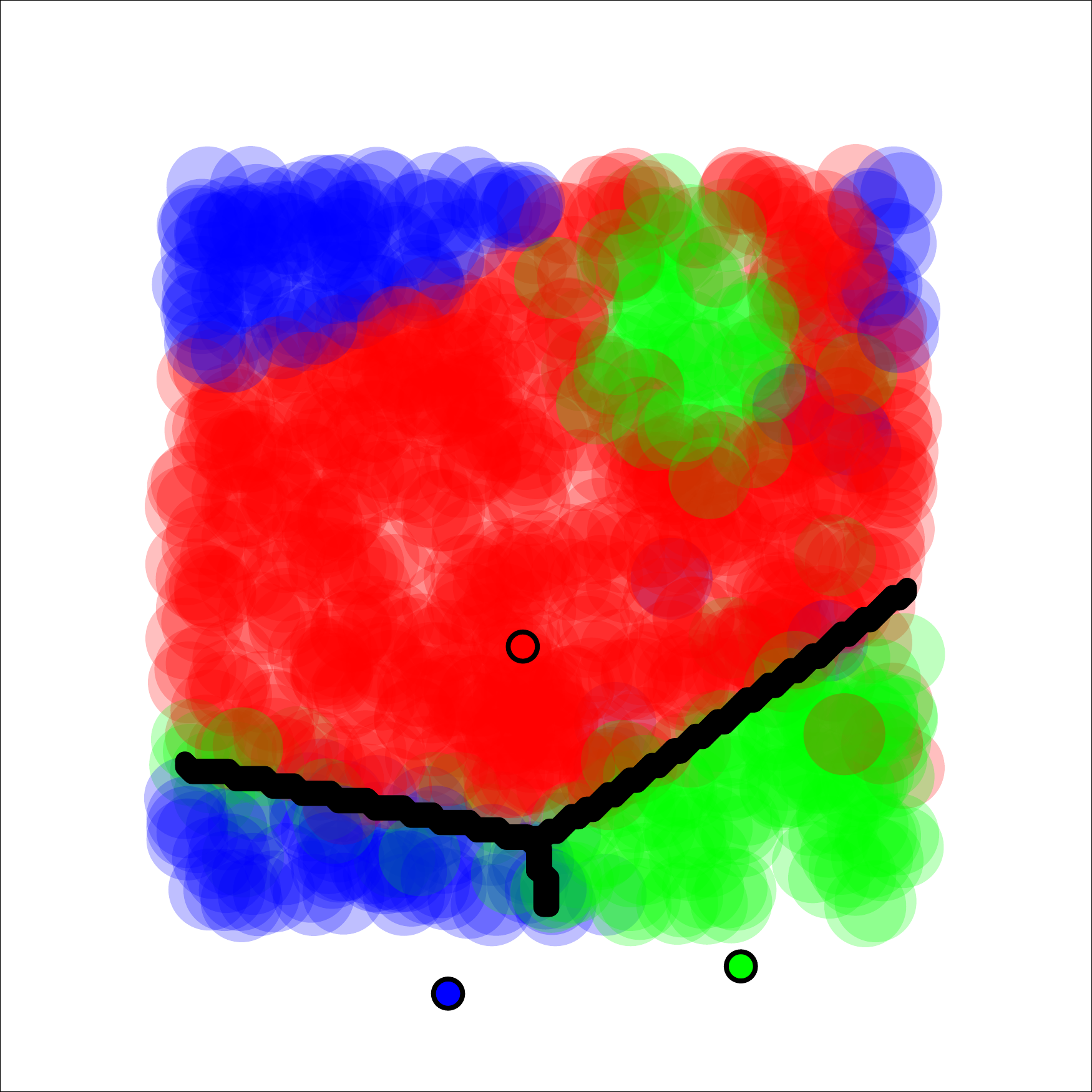}  &
				\includegraphics[width = 1.5cm]{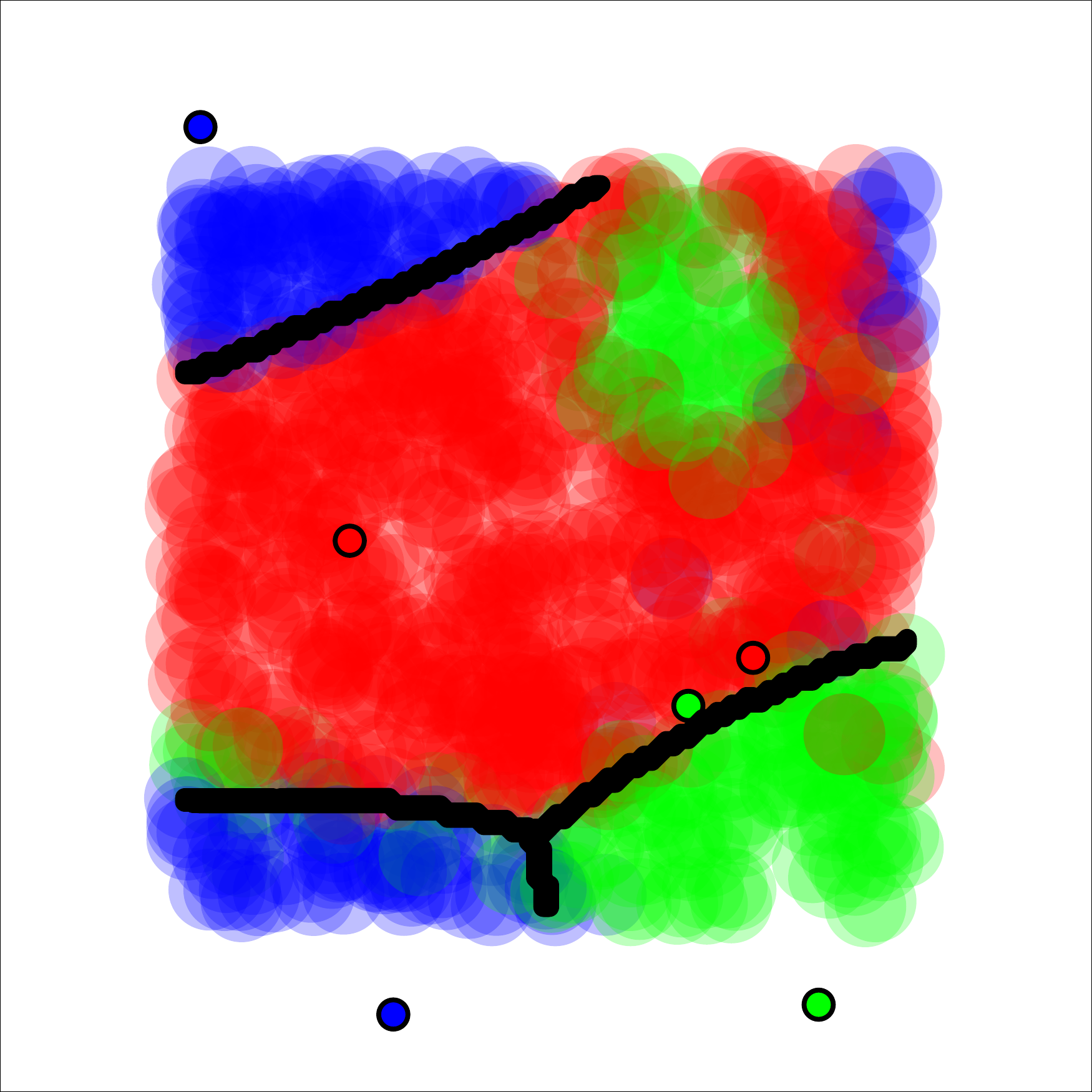}  &
				\includegraphics[width = 1.5cm]{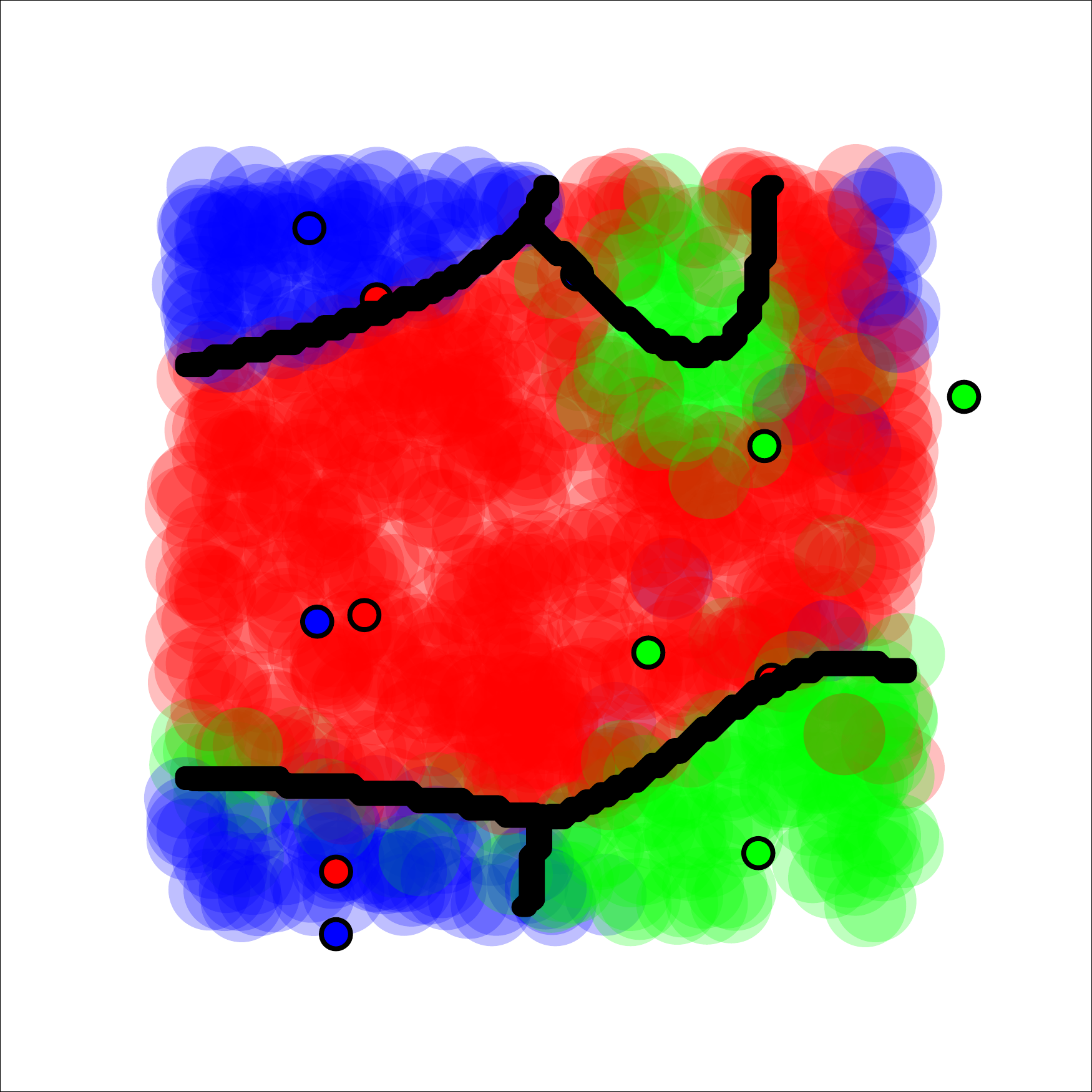}  &
				\includegraphics[width = 1.5cm]{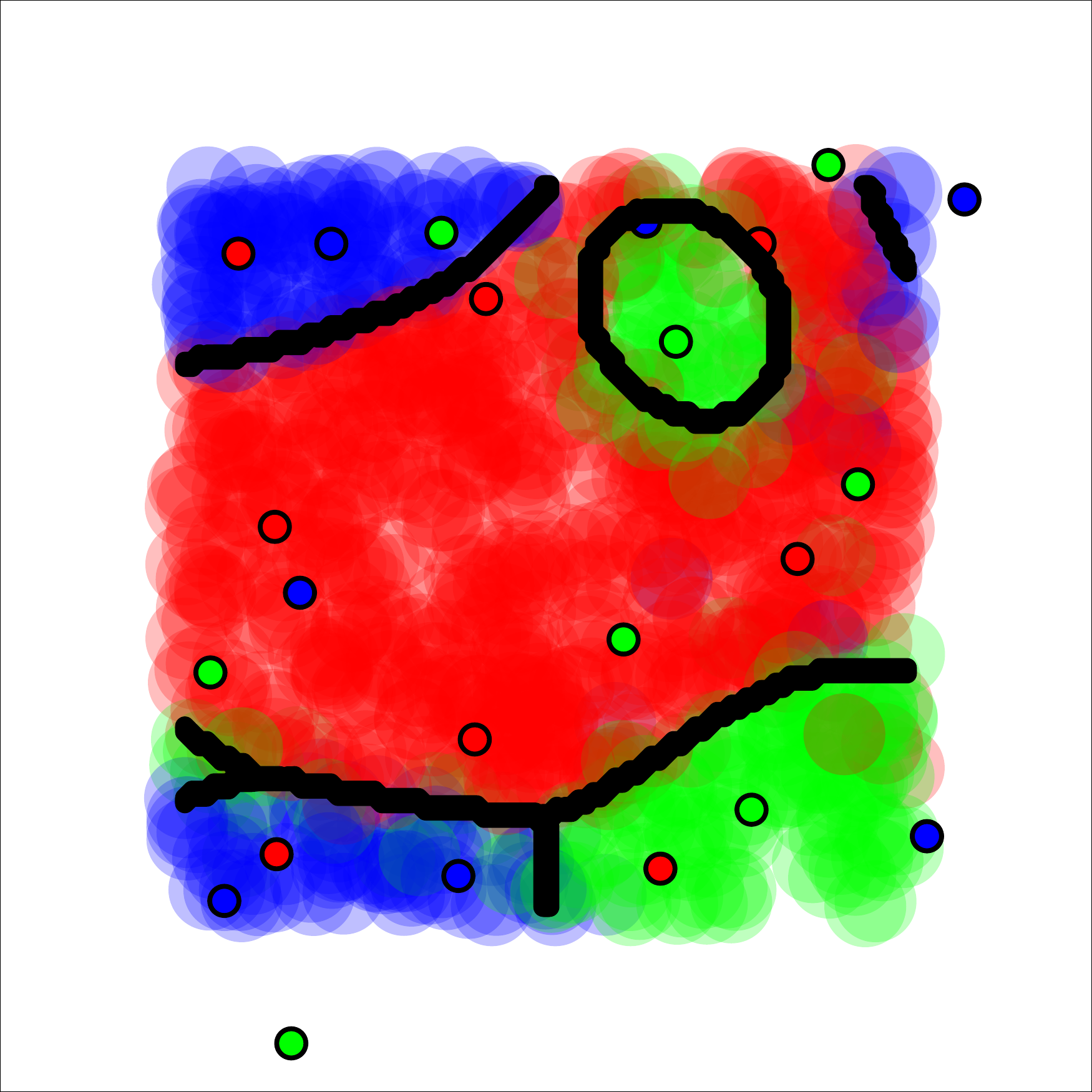}  &
				\includegraphics[width = 1.5cm]{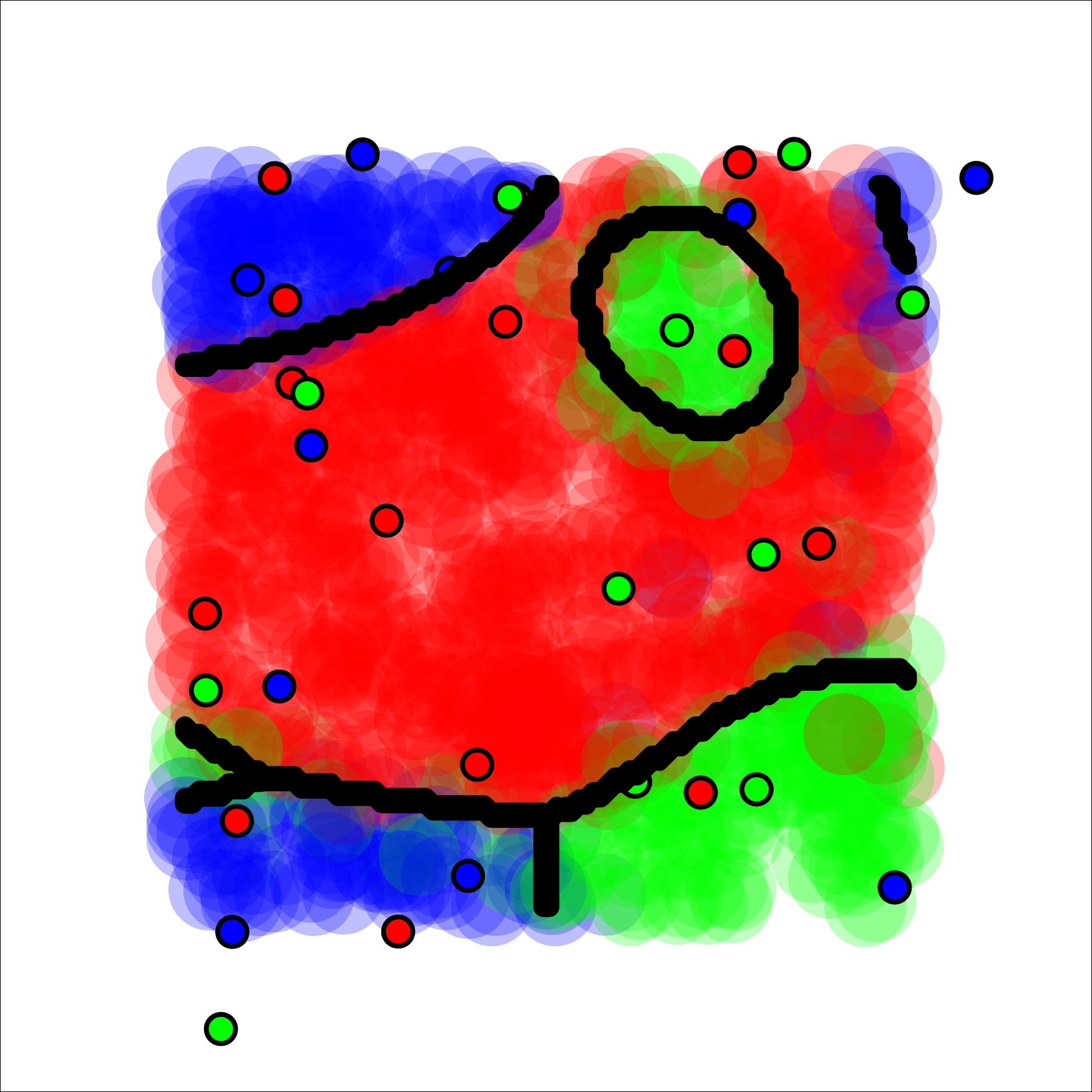}  &
				\includegraphics[width = 1.5cm]{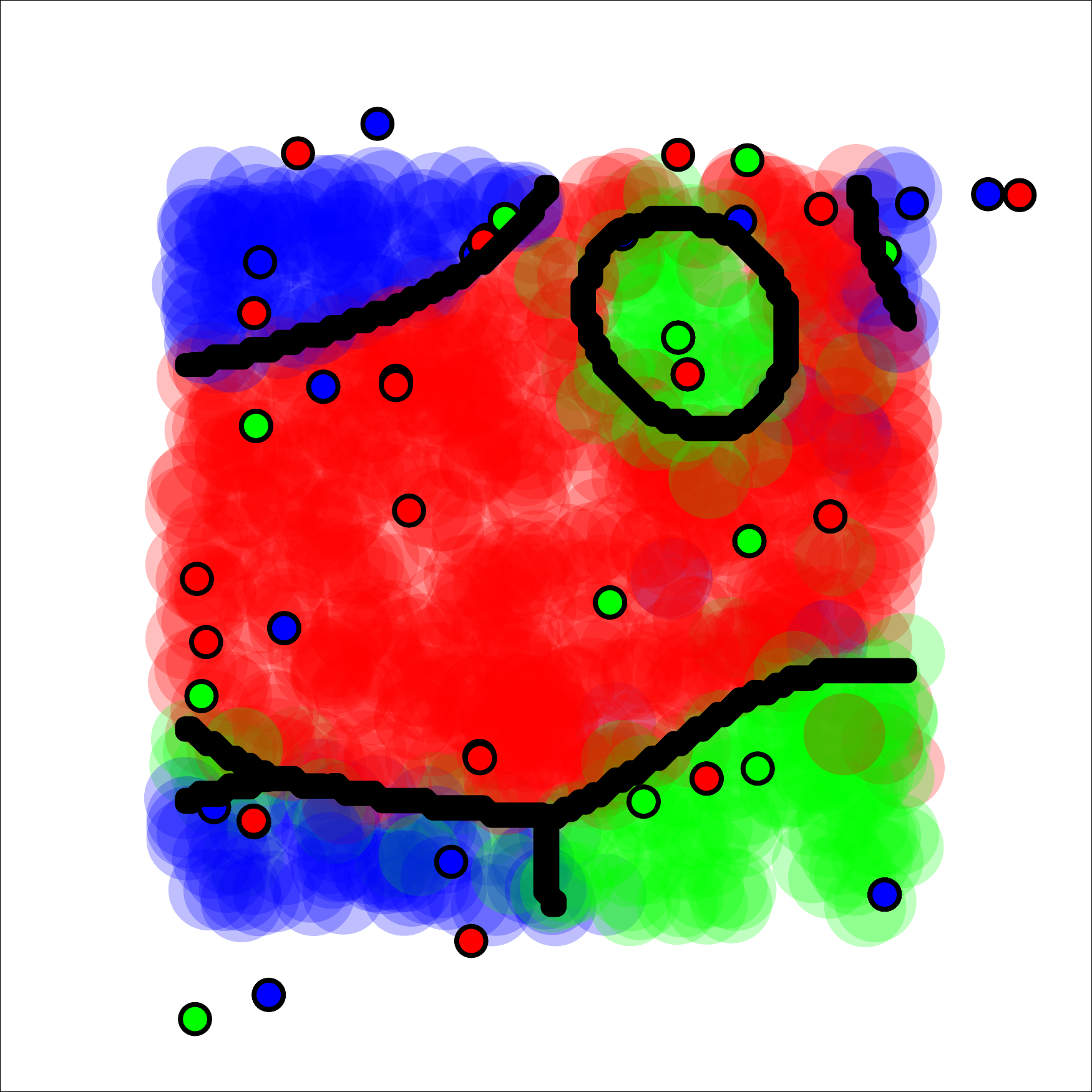}  &
				\includegraphics[width = 1.5cm]{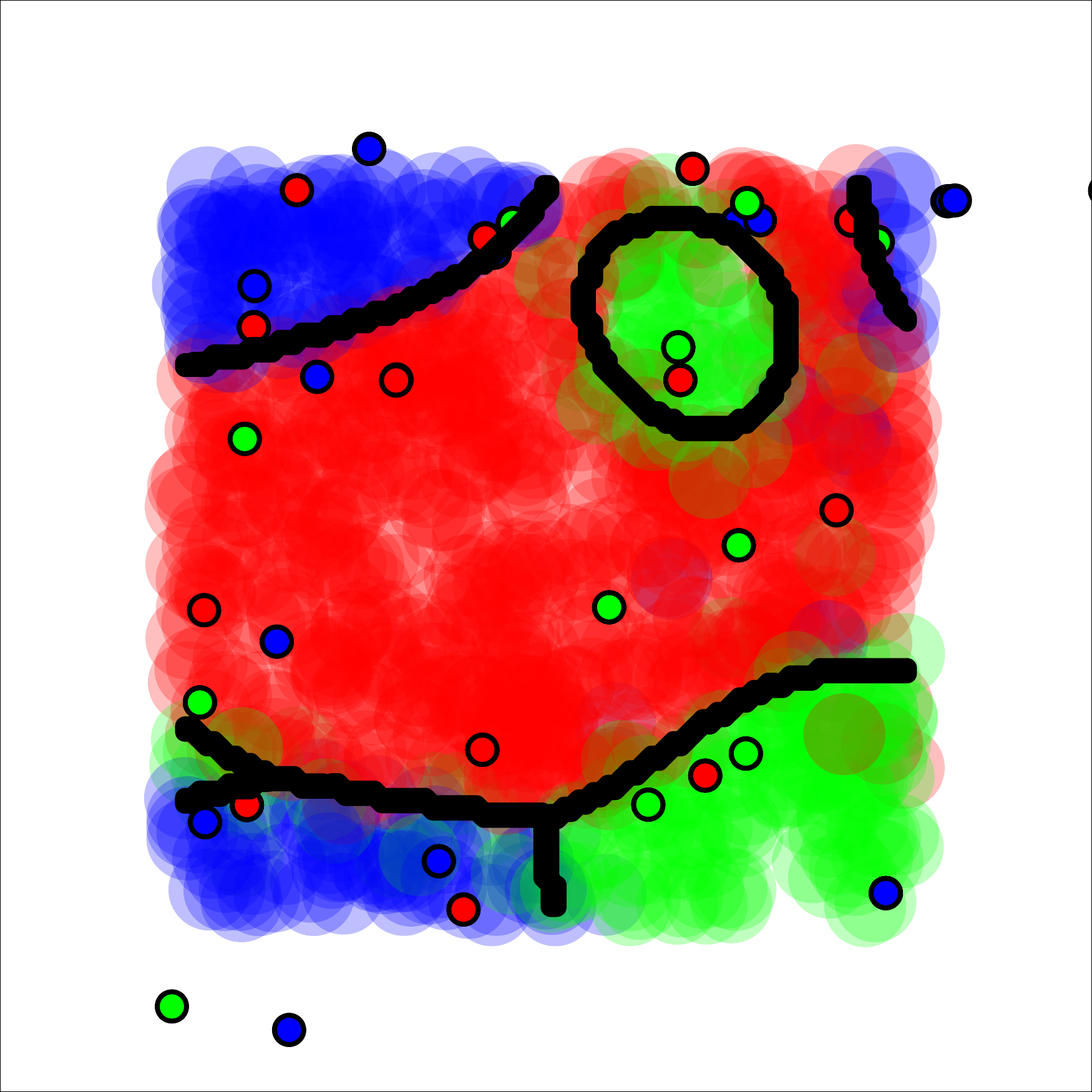}  &
				\includegraphics[width = 1.5cm]{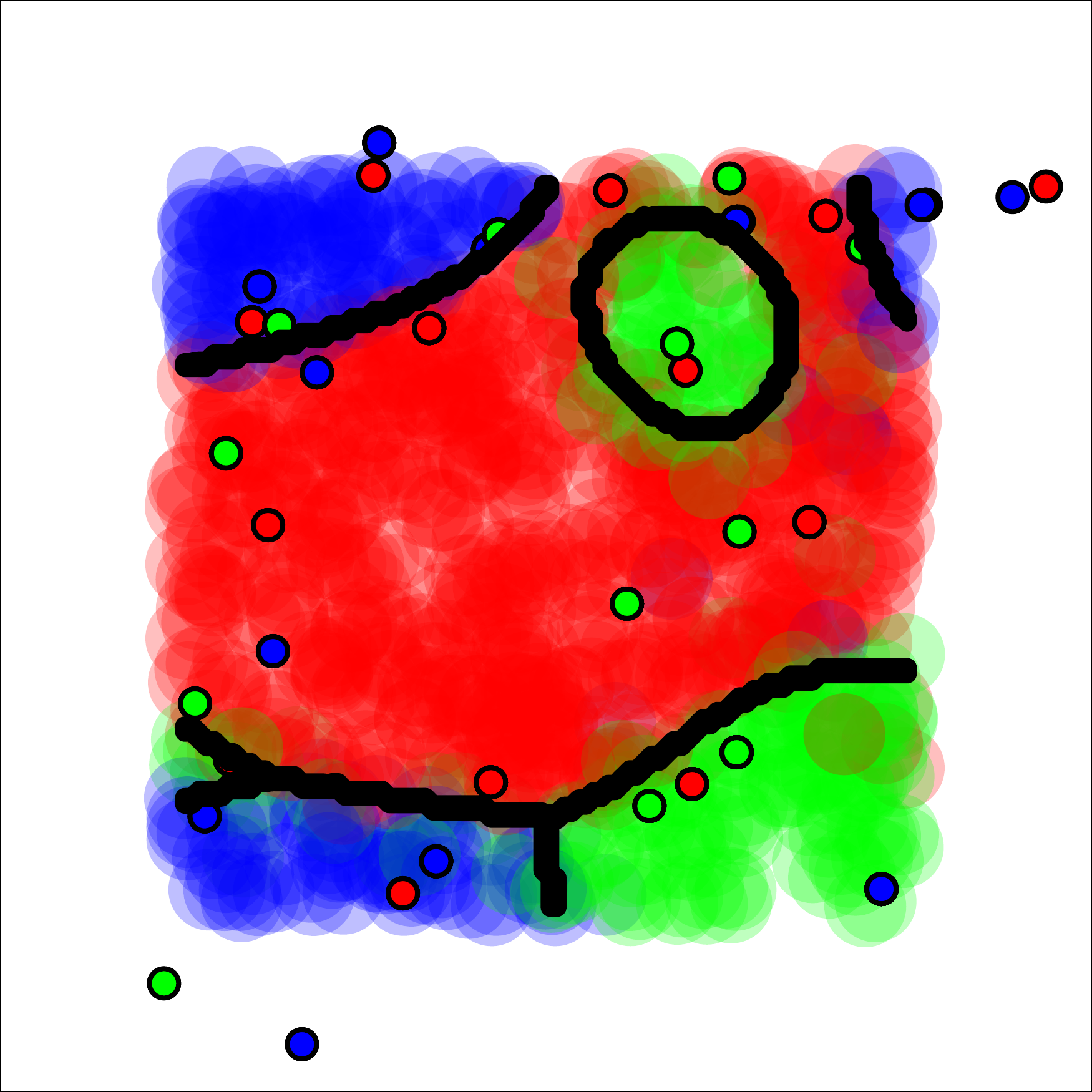} & 
				\includegraphics[width = 1.5cm]{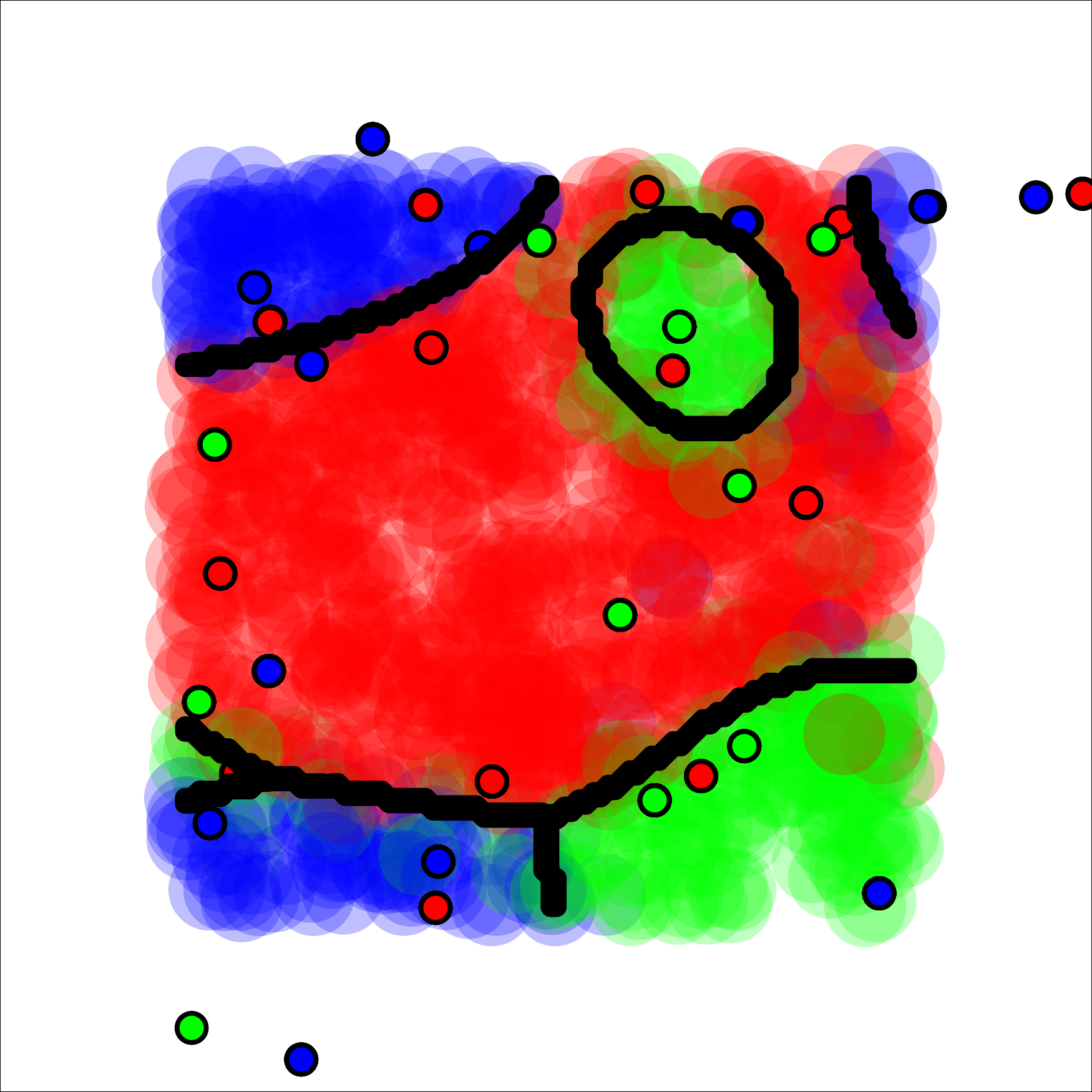}  
				\\
				
				\rotatebox{90}{\hspace{0.70cm}{{\scriptsize {\bf EP}}}} &
				\includegraphics[width = 1.5cm]{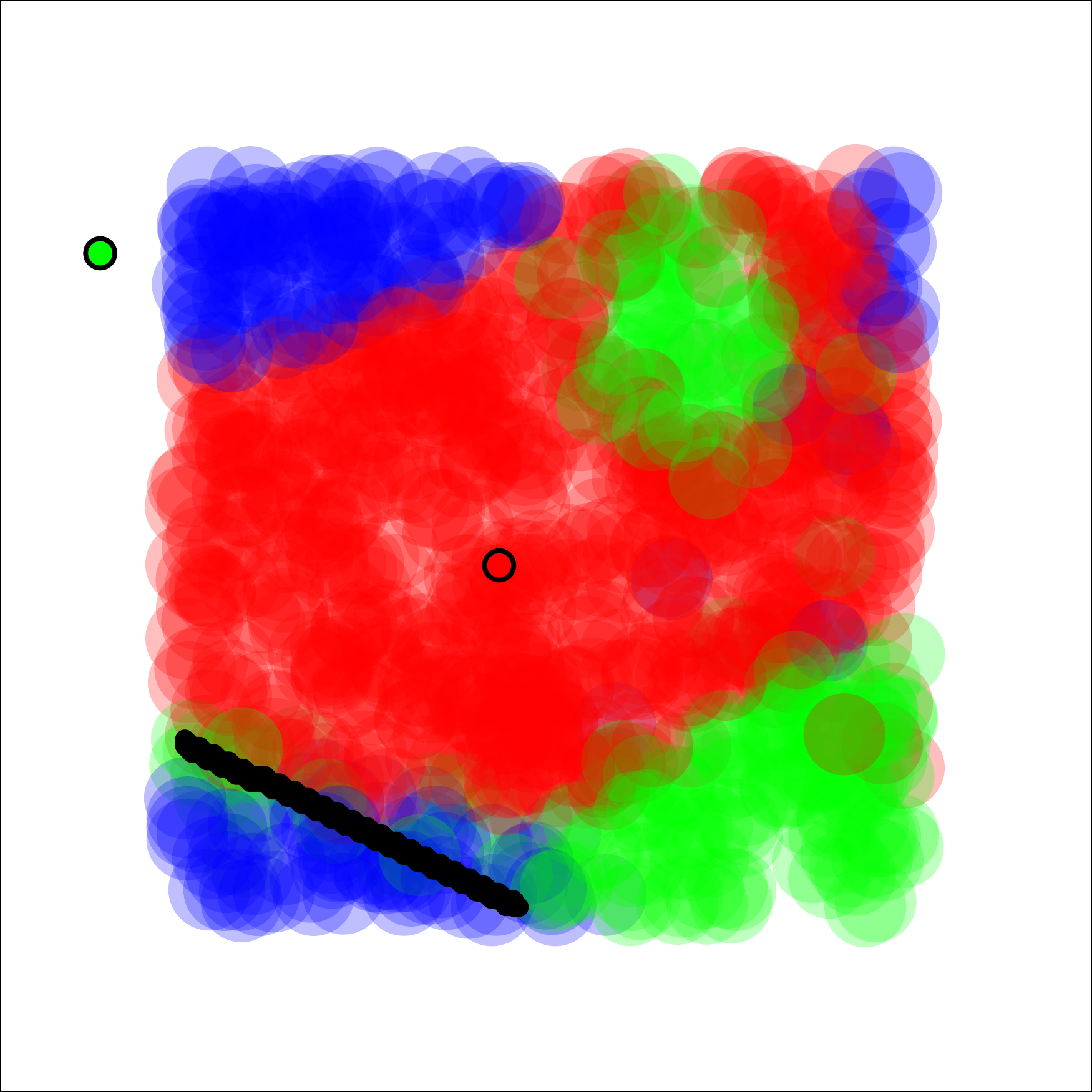}  &
				\includegraphics[width = 1.5cm]{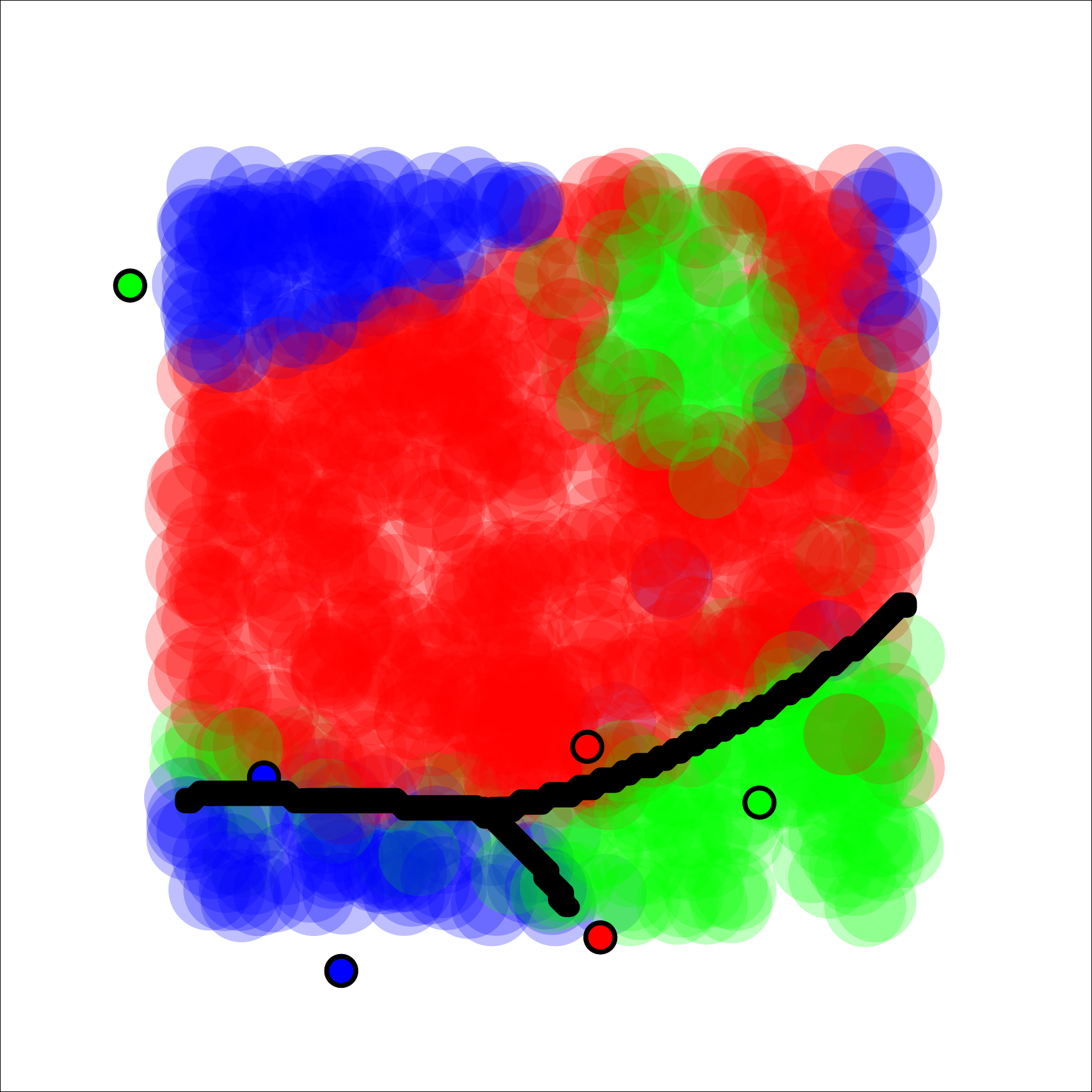}  &
				\includegraphics[width = 1.5cm]{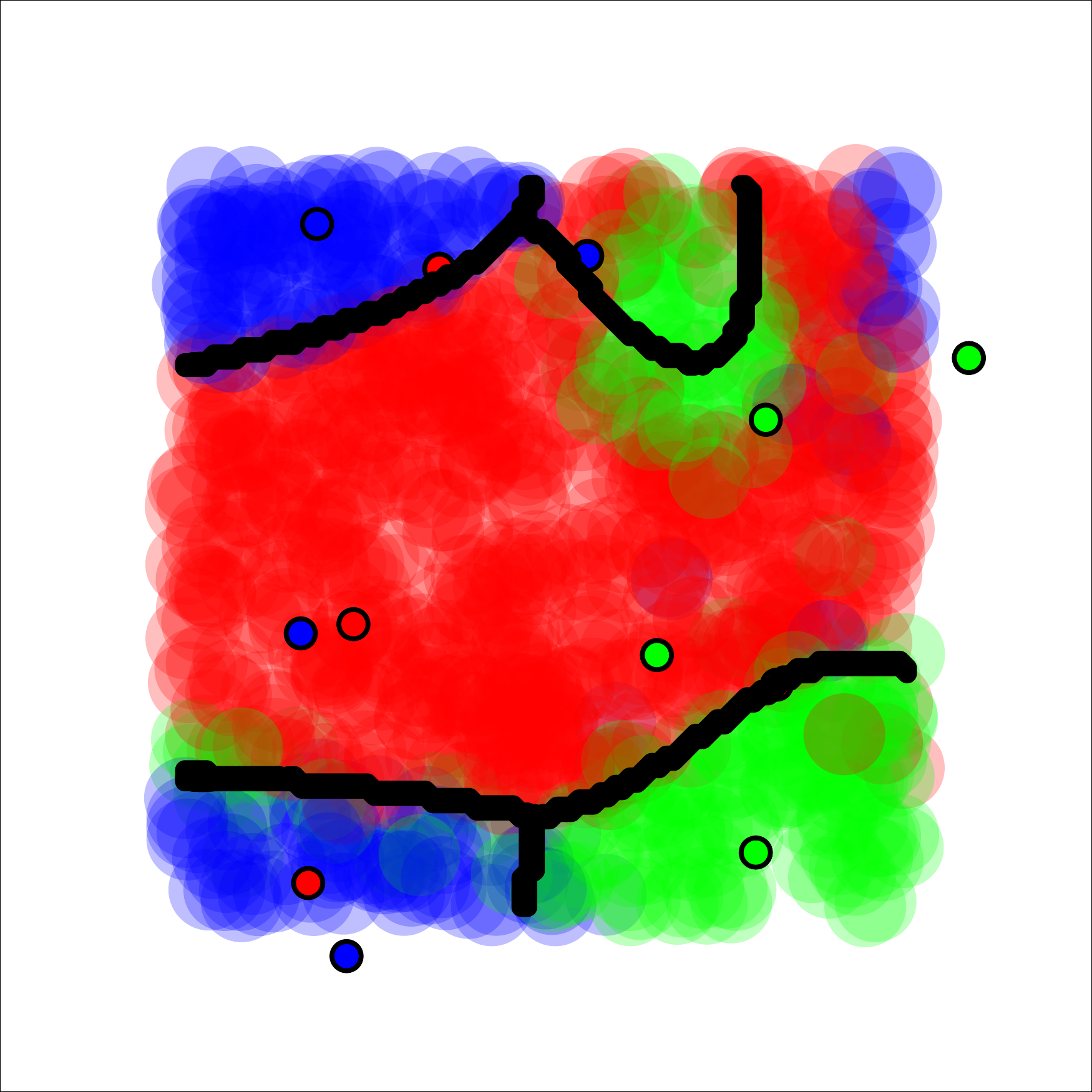}  &
				\includegraphics[width = 1.5cm]{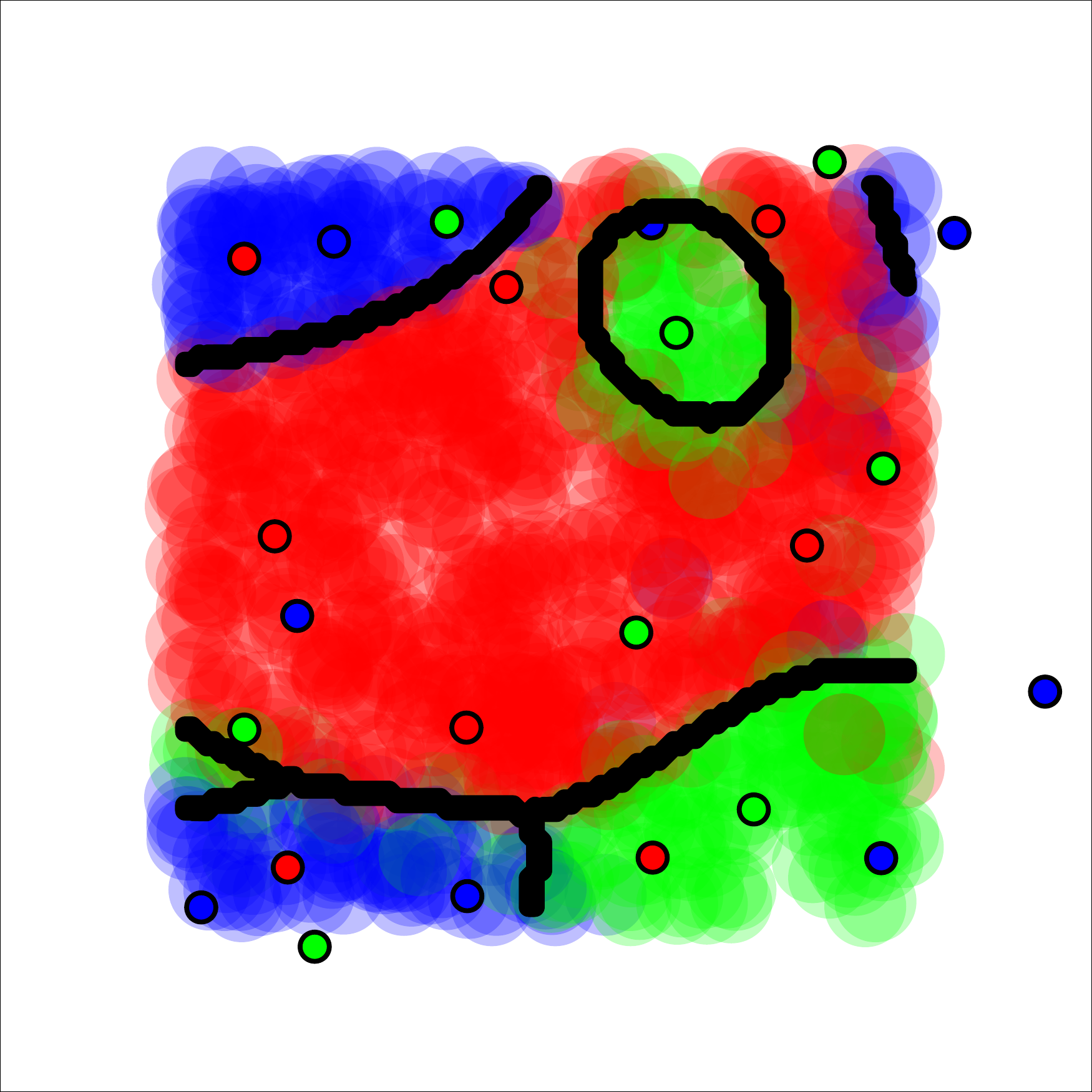}  &
				\includegraphics[width = 1.5cm]{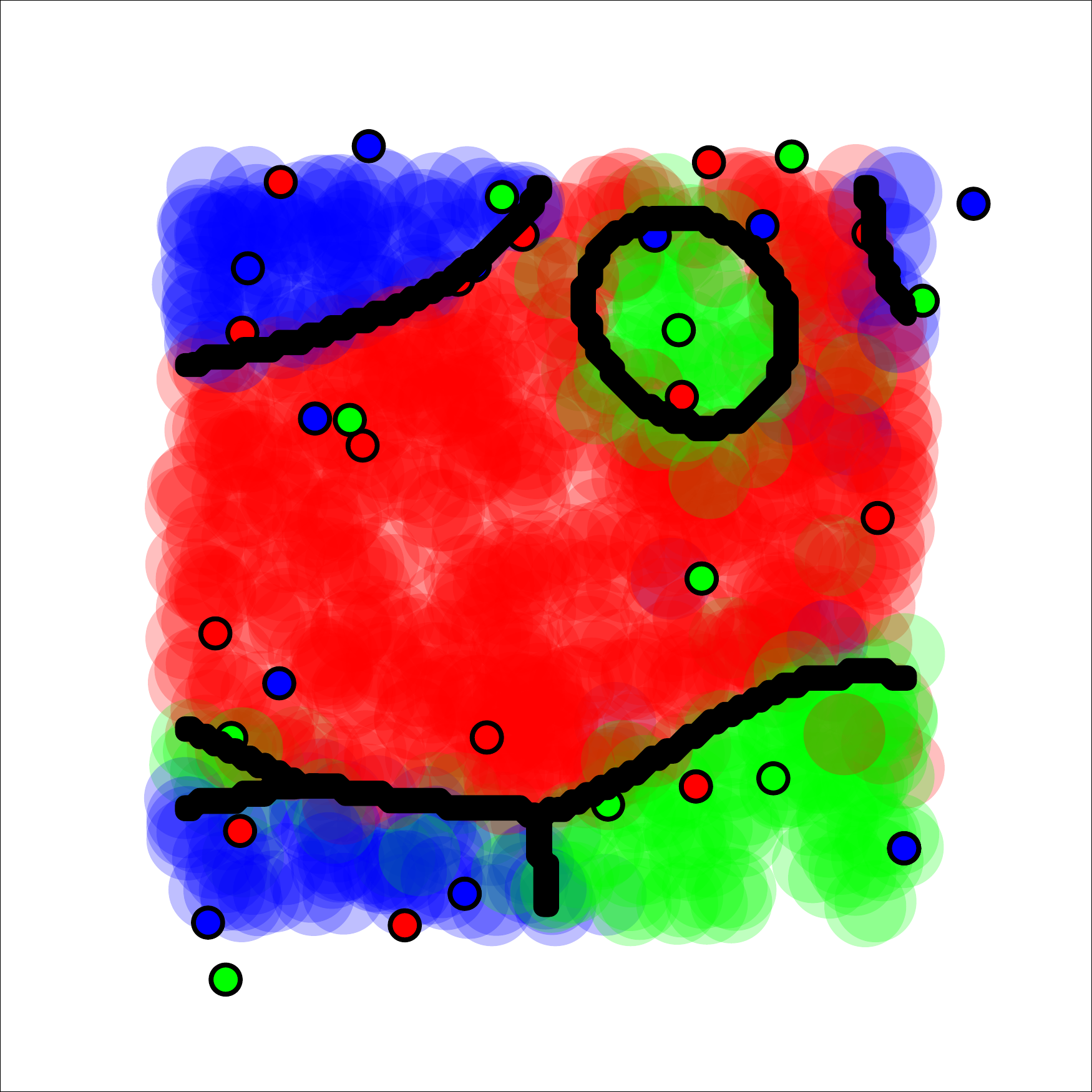}  &
				\includegraphics[width = 1.5cm]{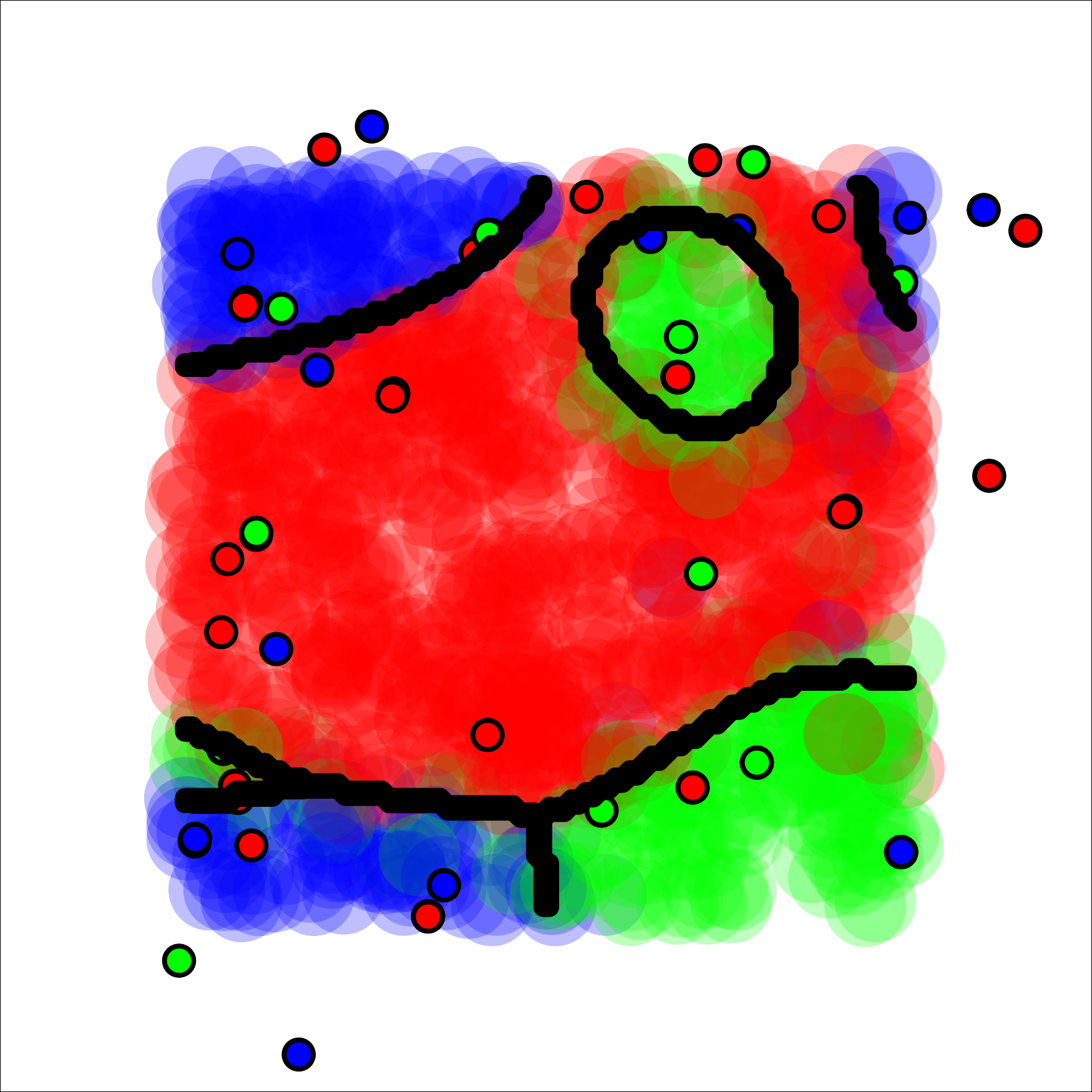}  &
				\includegraphics[width = 1.5cm]{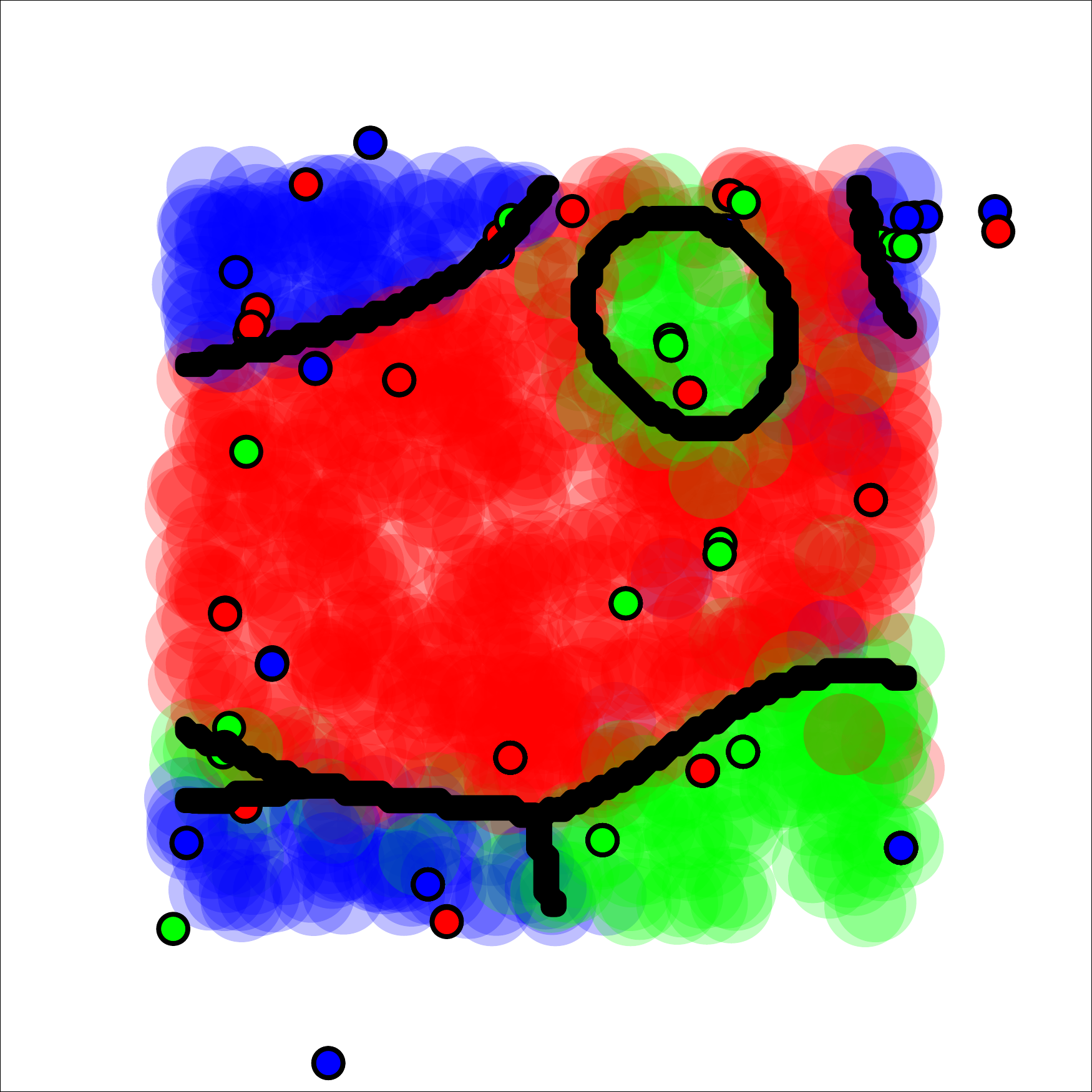}  &
				\includegraphics[width = 1.5cm]{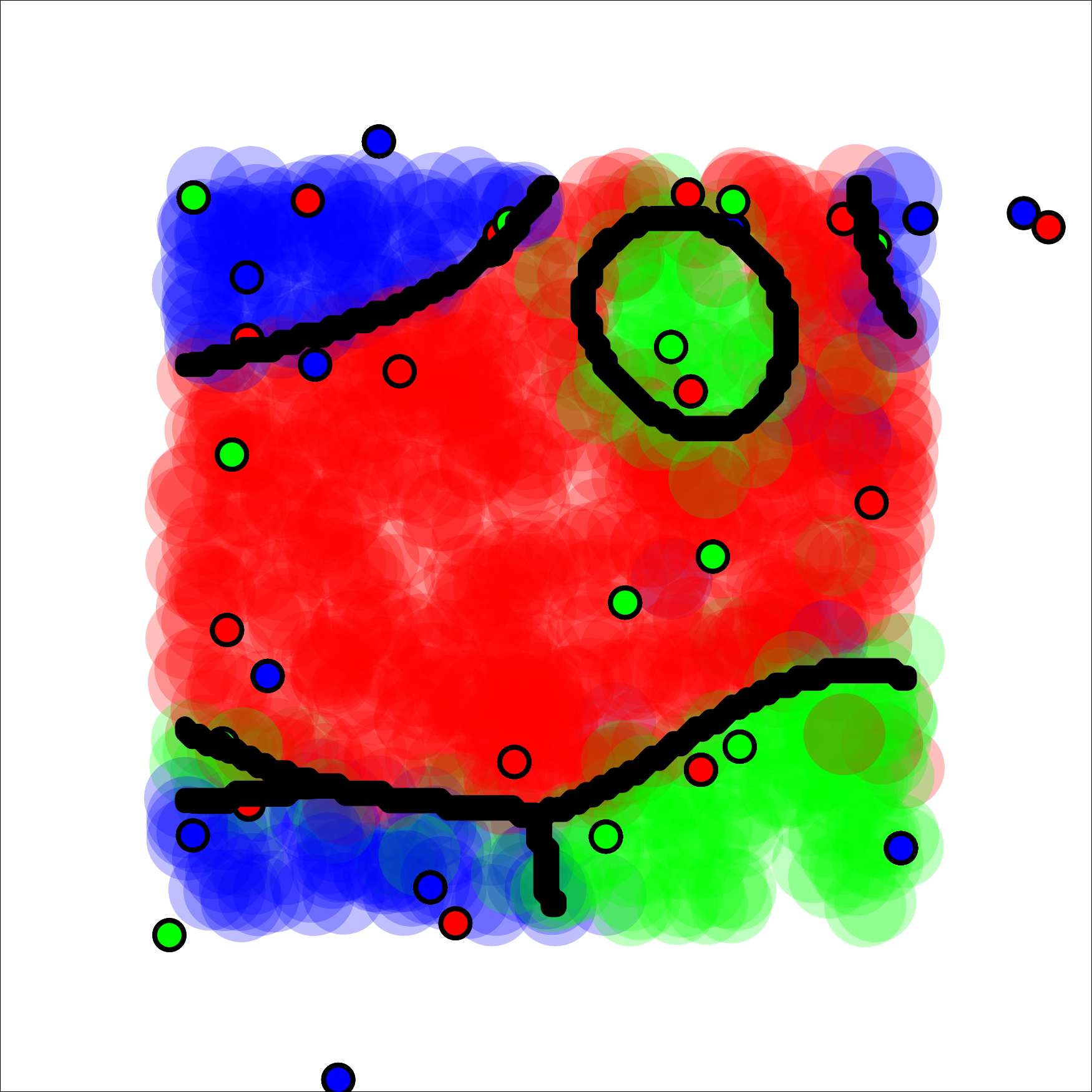} &
				\includegraphics[width = 1.5cm]{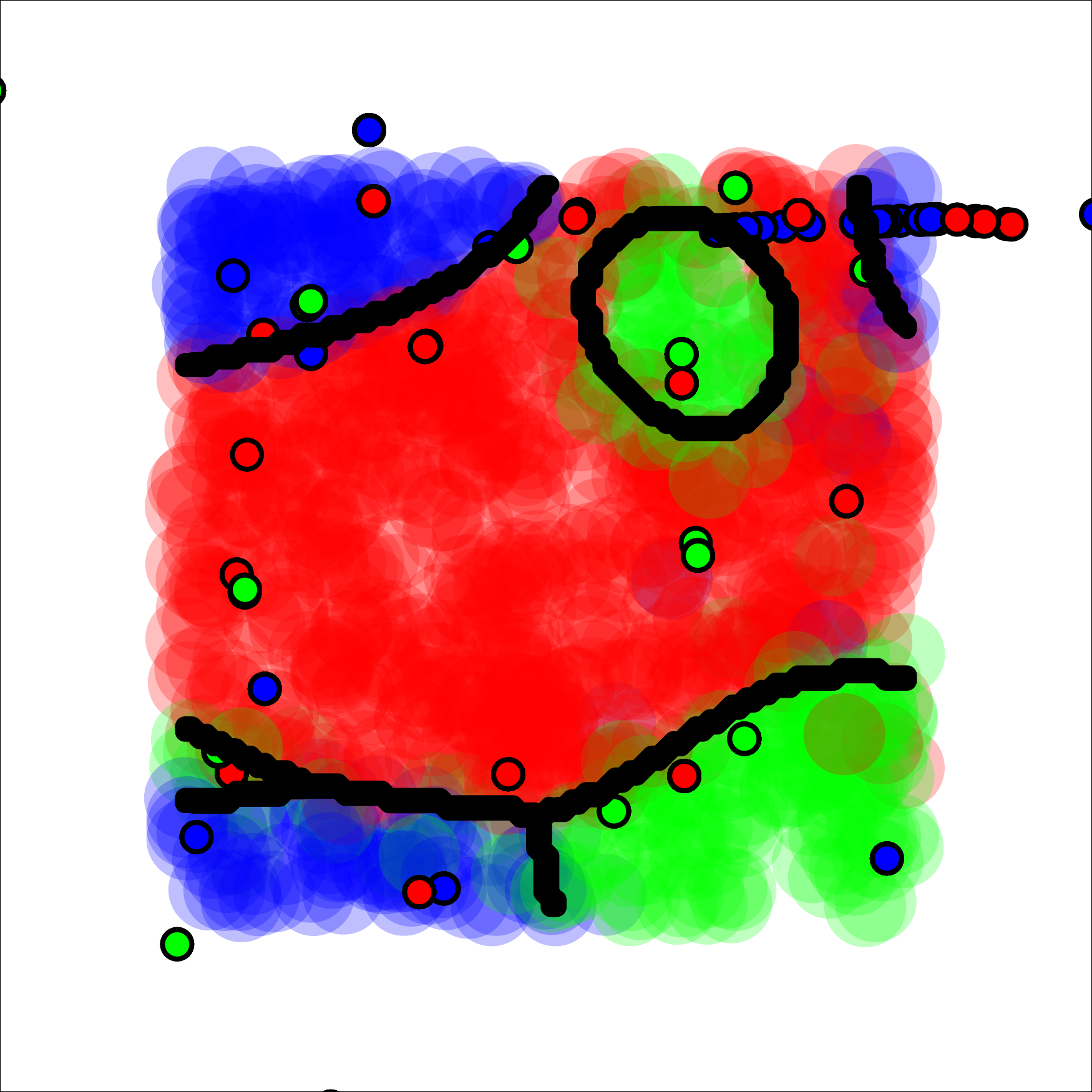}  
				\\
				
				\rotatebox{90}{\hspace{0.55cm}{{\scriptsize {\bf SEP}}}} &
				\includegraphics[width = 1.5cm]{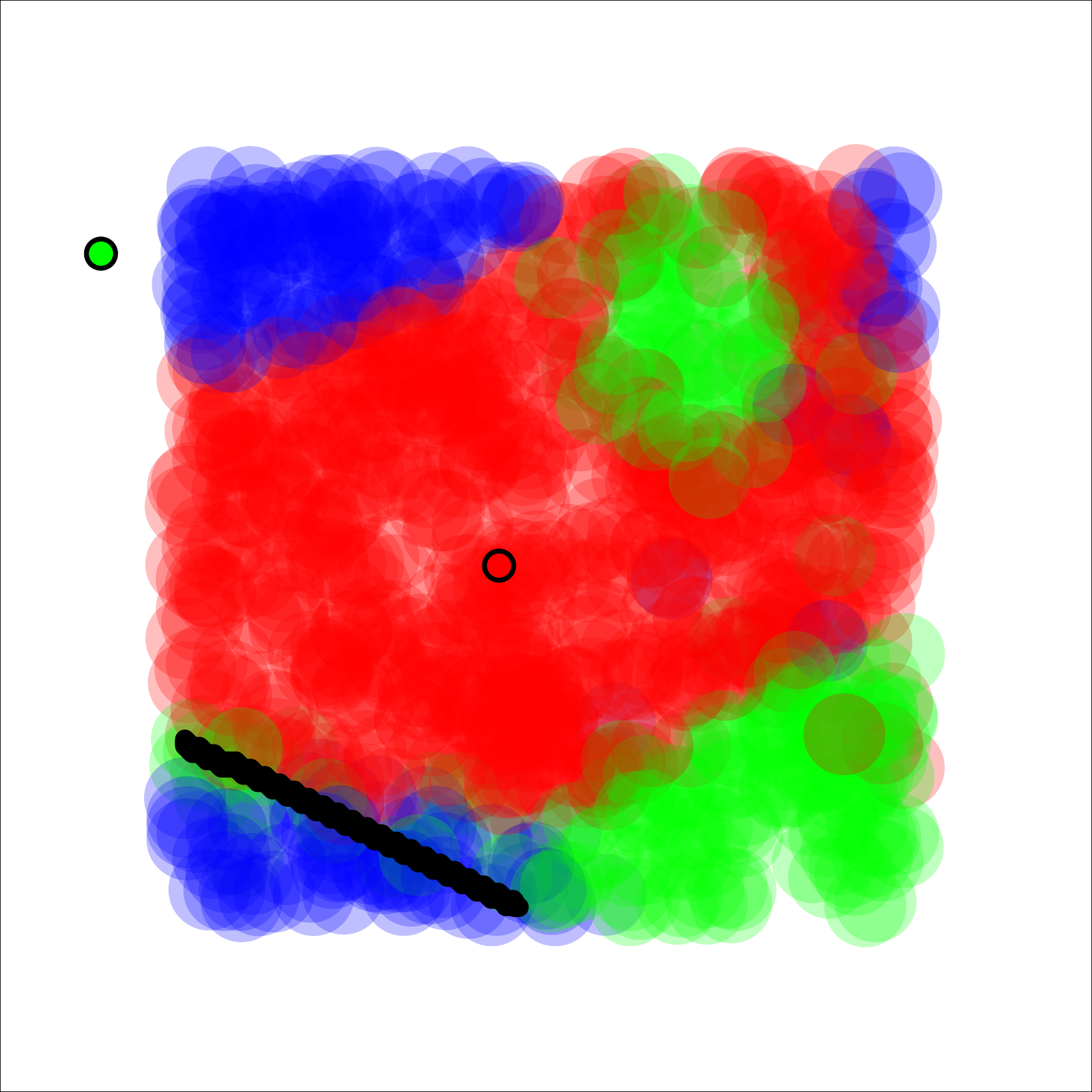}  &
				\includegraphics[width = 1.5cm]{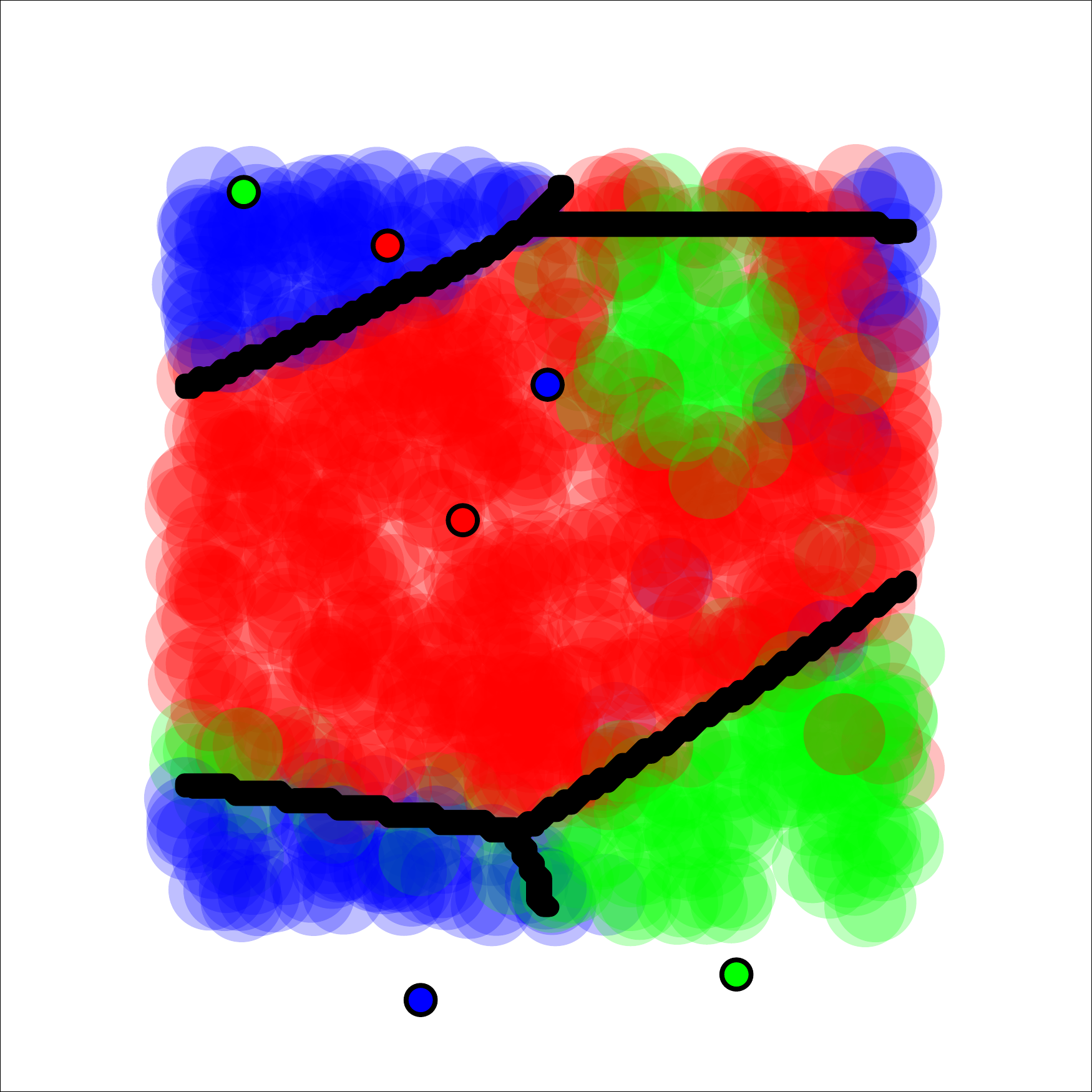}  &
				\includegraphics[width = 1.5cm]{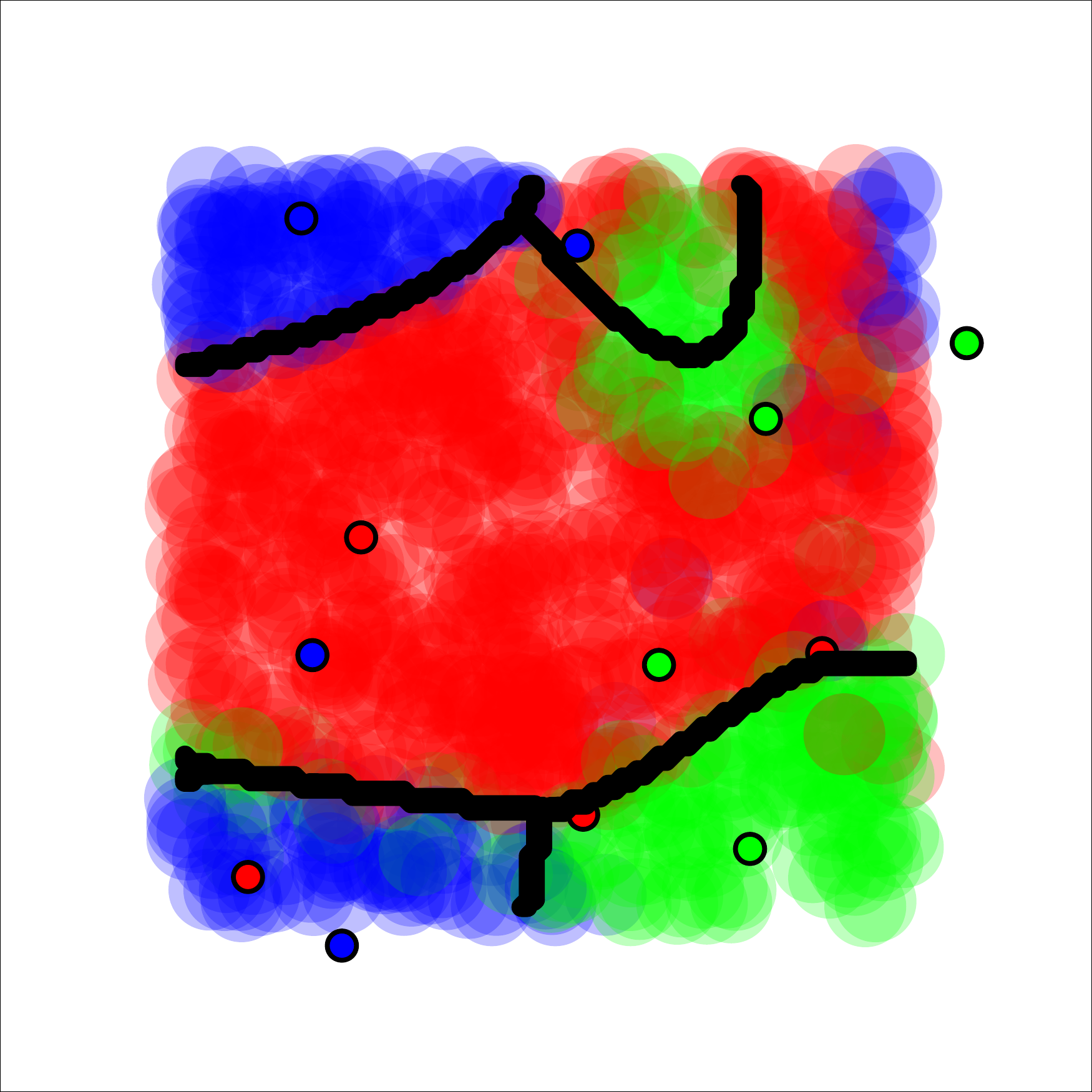}  &
				\includegraphics[width = 1.5cm]{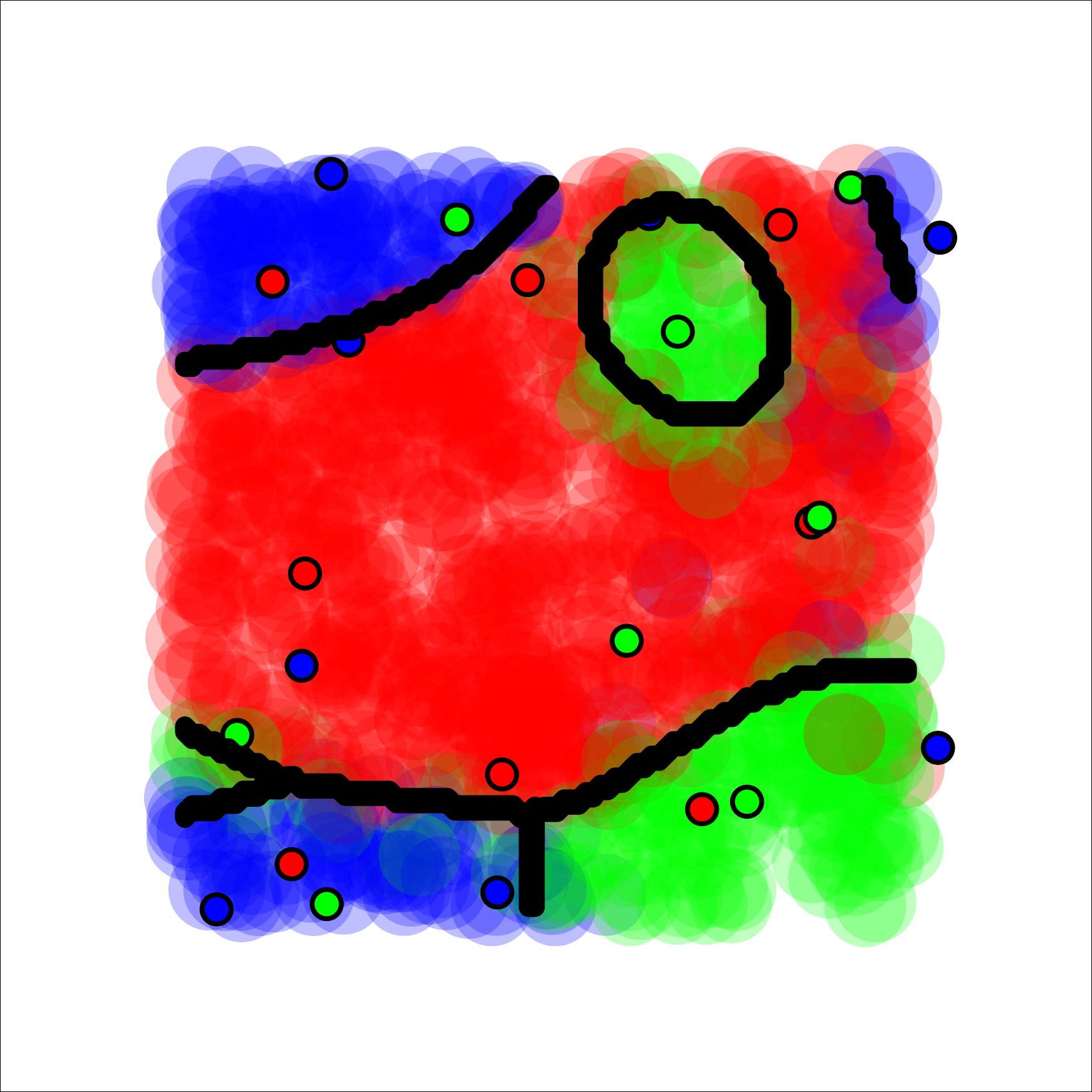}  &
				\includegraphics[width = 1.5cm]{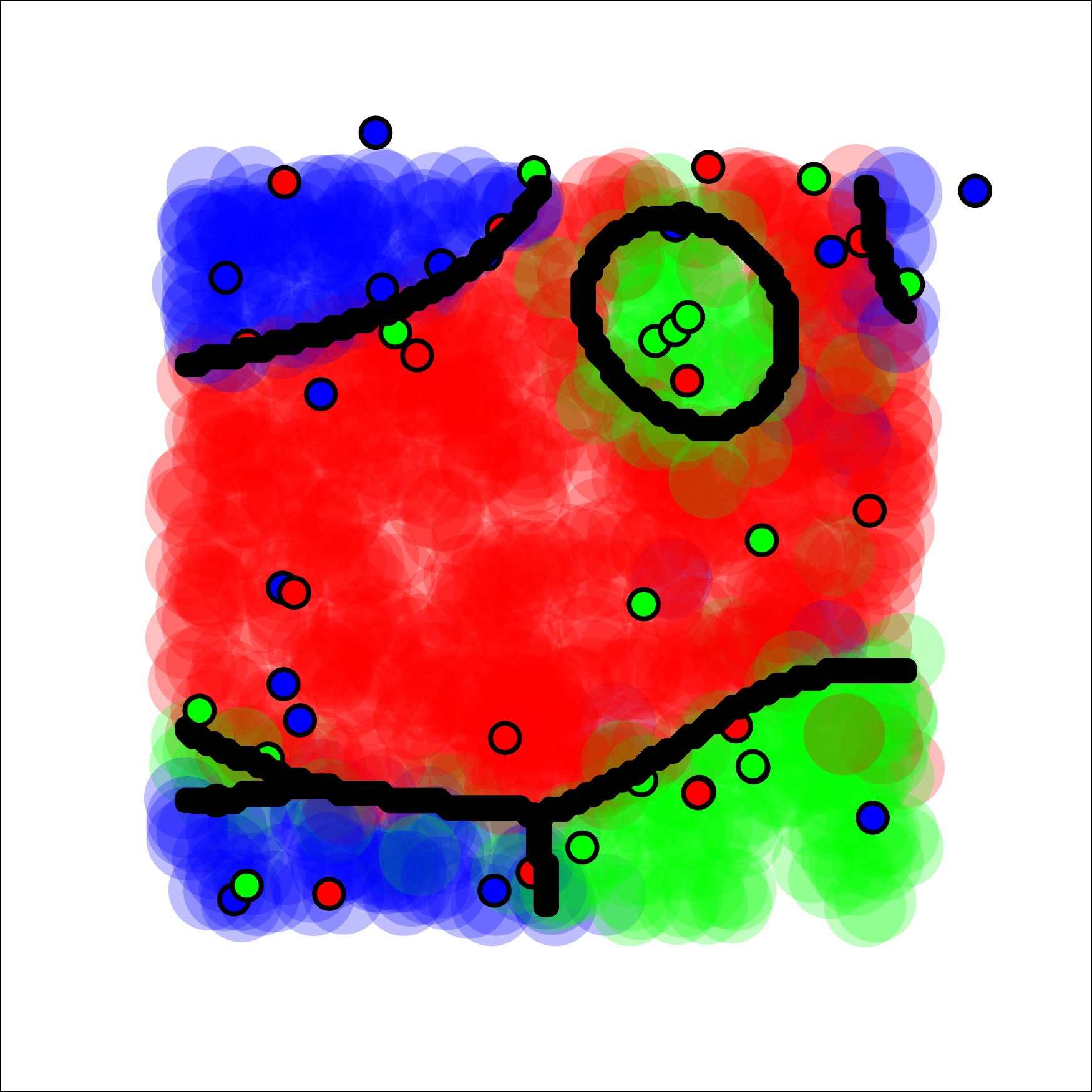}  &
				\includegraphics[width = 1.5cm]{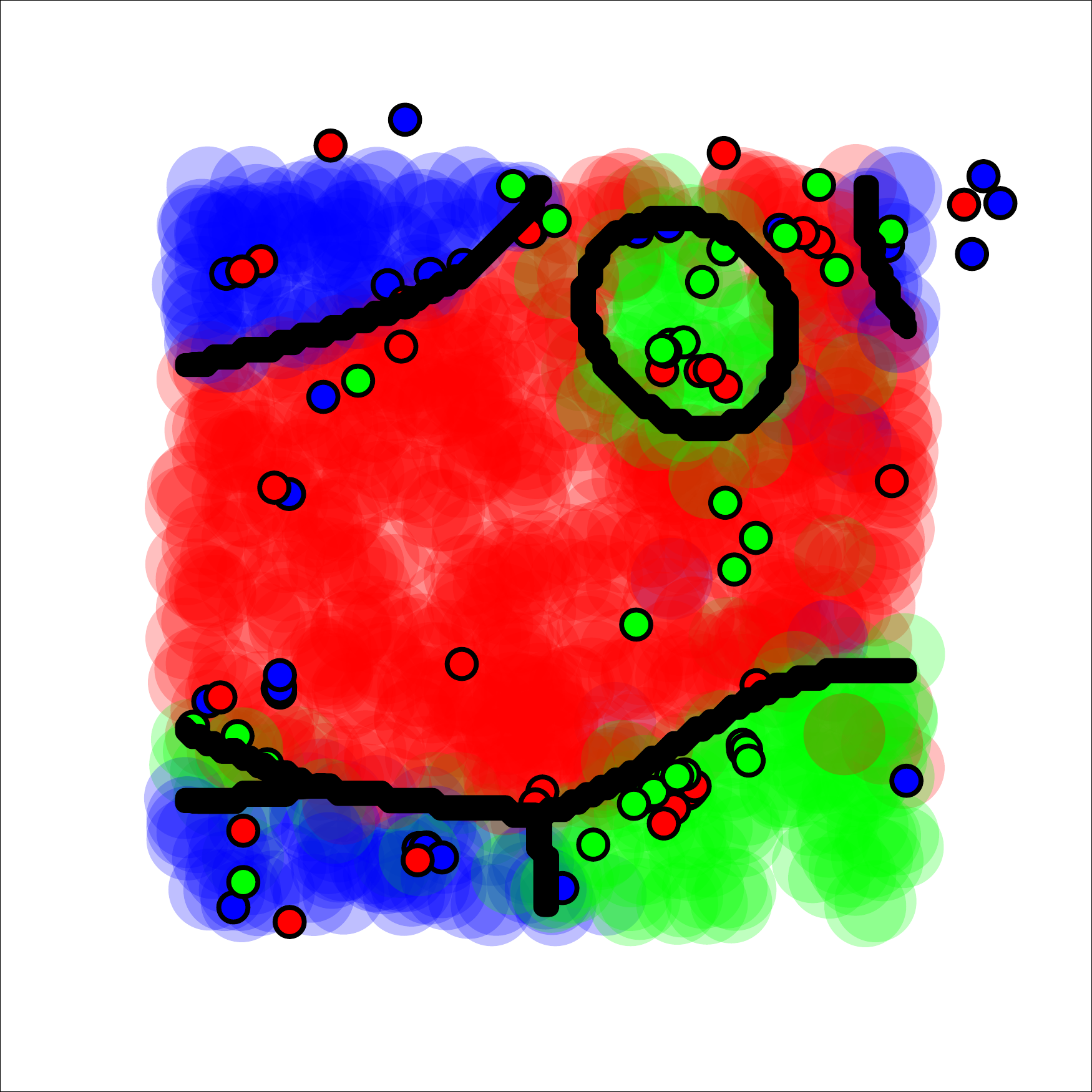}  &
				\includegraphics[width = 1.5cm]{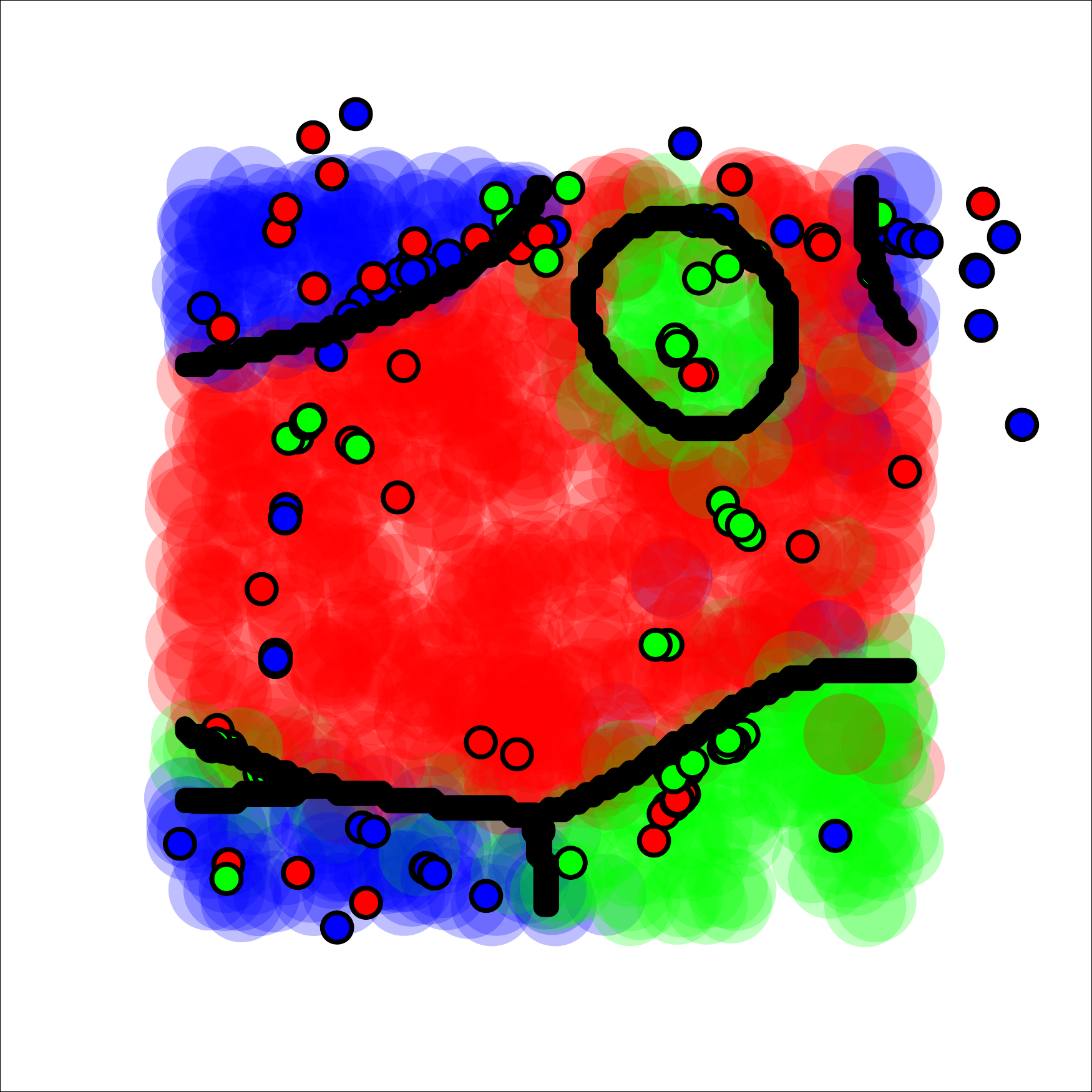}  &
				\includegraphics[width = 1.5cm]{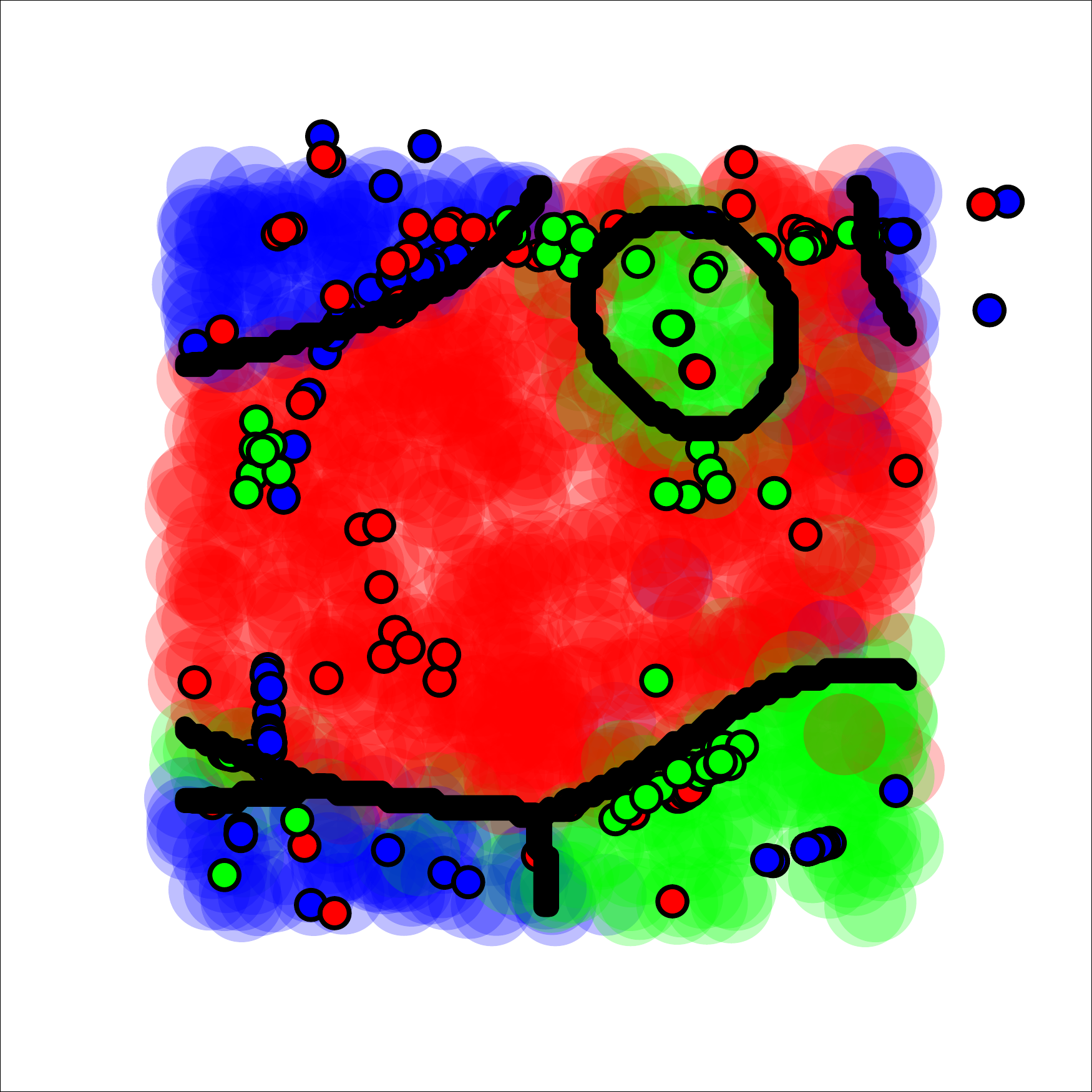} &
				\includegraphics[width = 1.5cm]{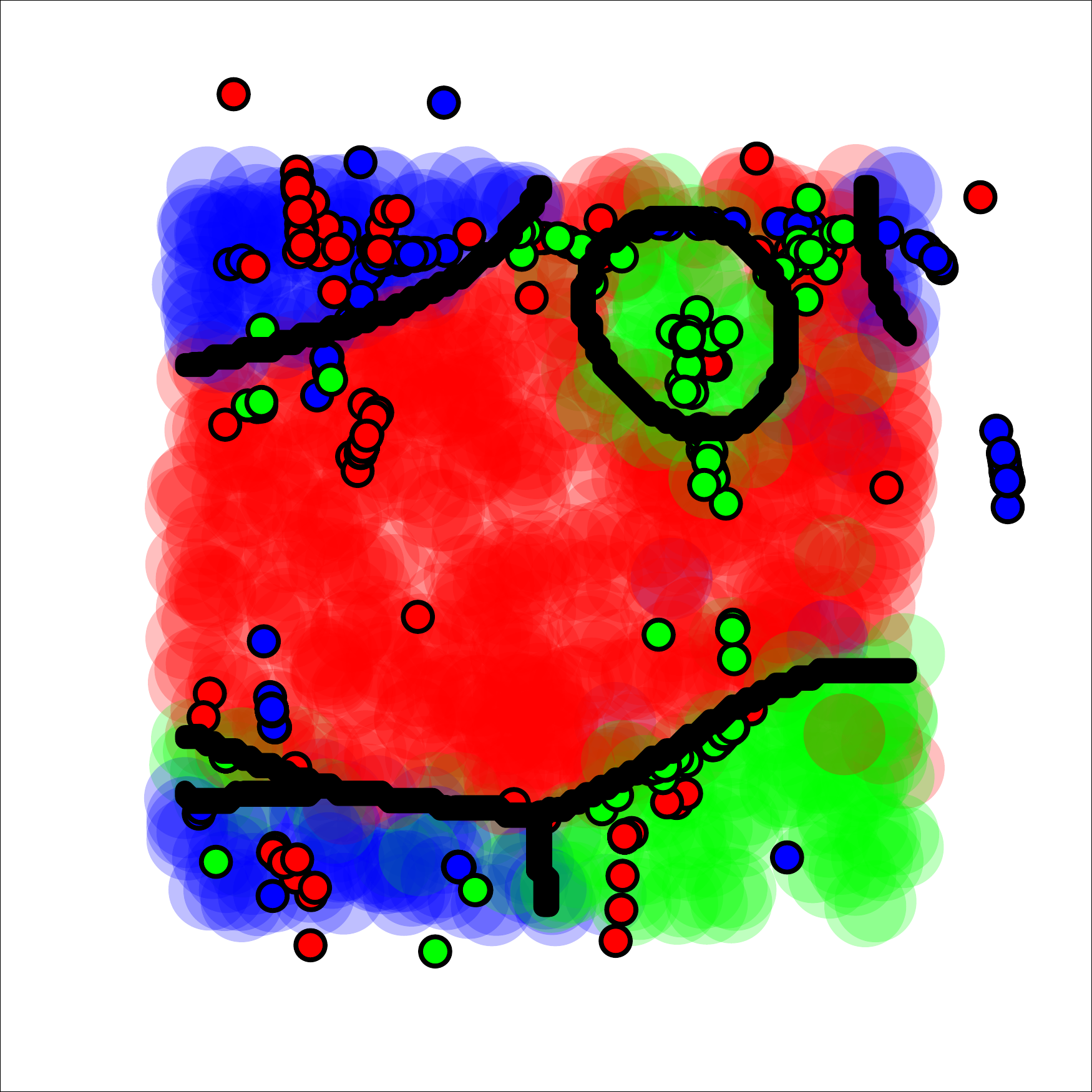}  
				\\
				
				\rotatebox{90}{\hspace{0.70cm}{{\scriptsize {\bf VI}}}} &
				\includegraphics[width = 1.5cm]{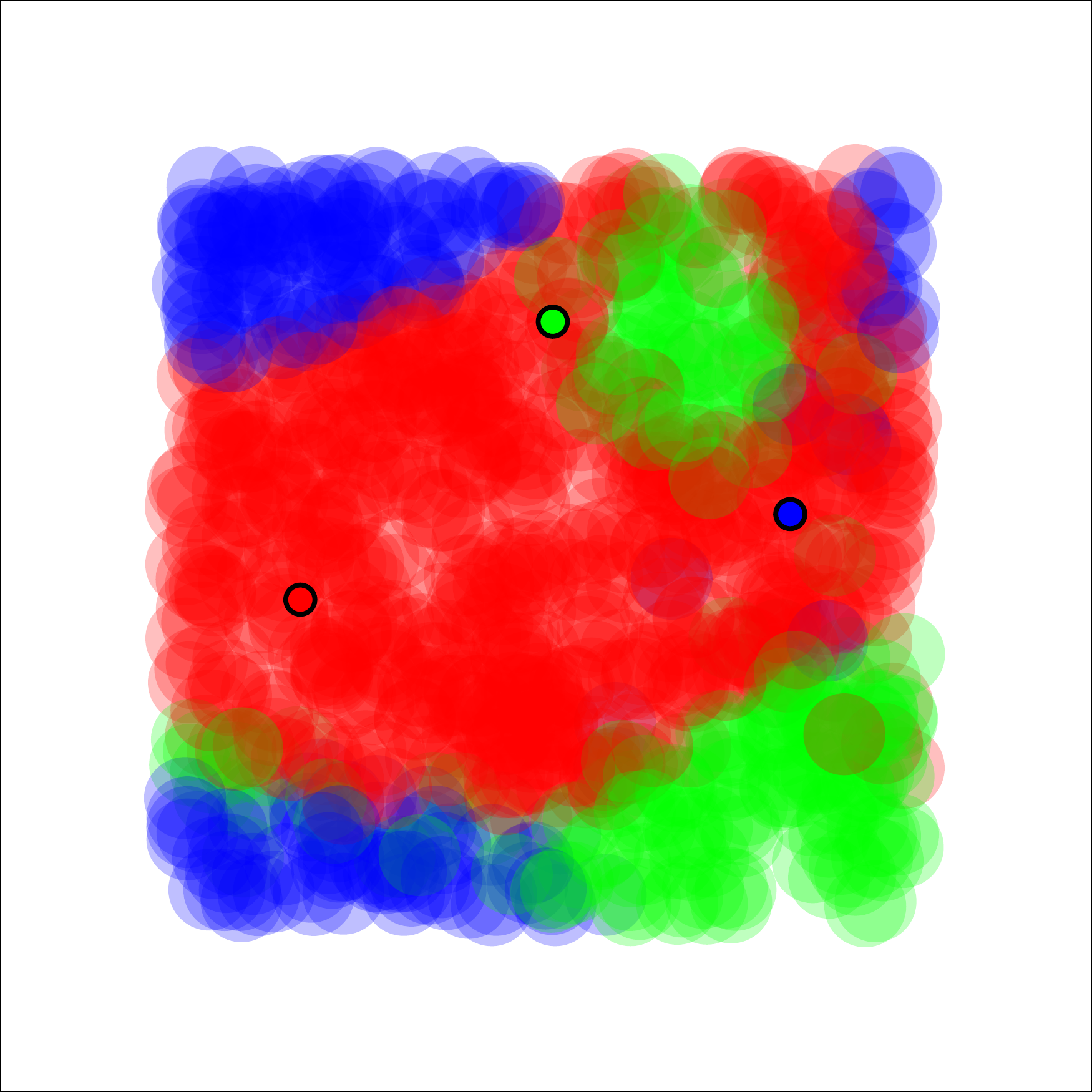}  &
				\includegraphics[width = 1.5cm]{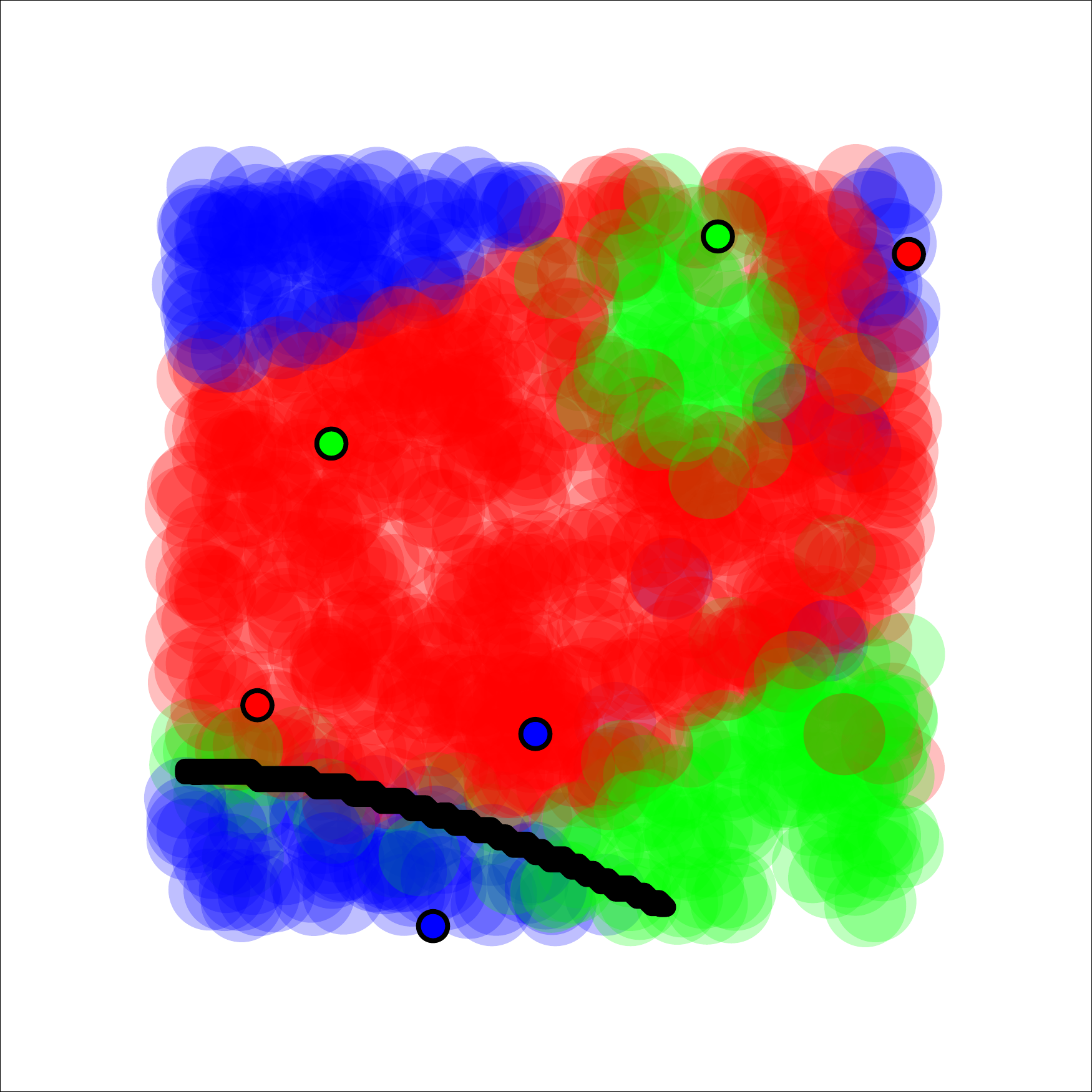}  &
				\includegraphics[width = 1.5cm]{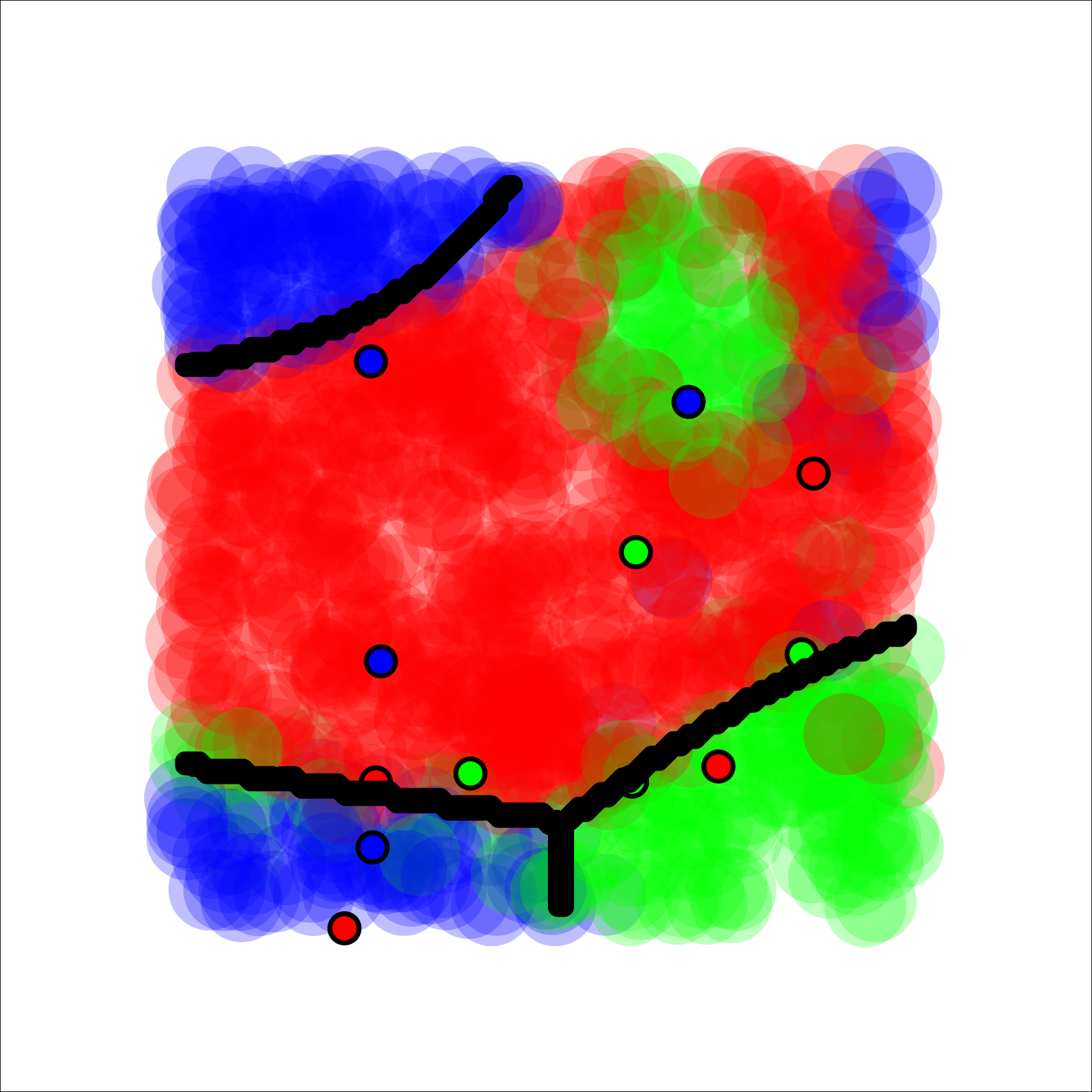}  &
				\includegraphics[width = 1.5cm]{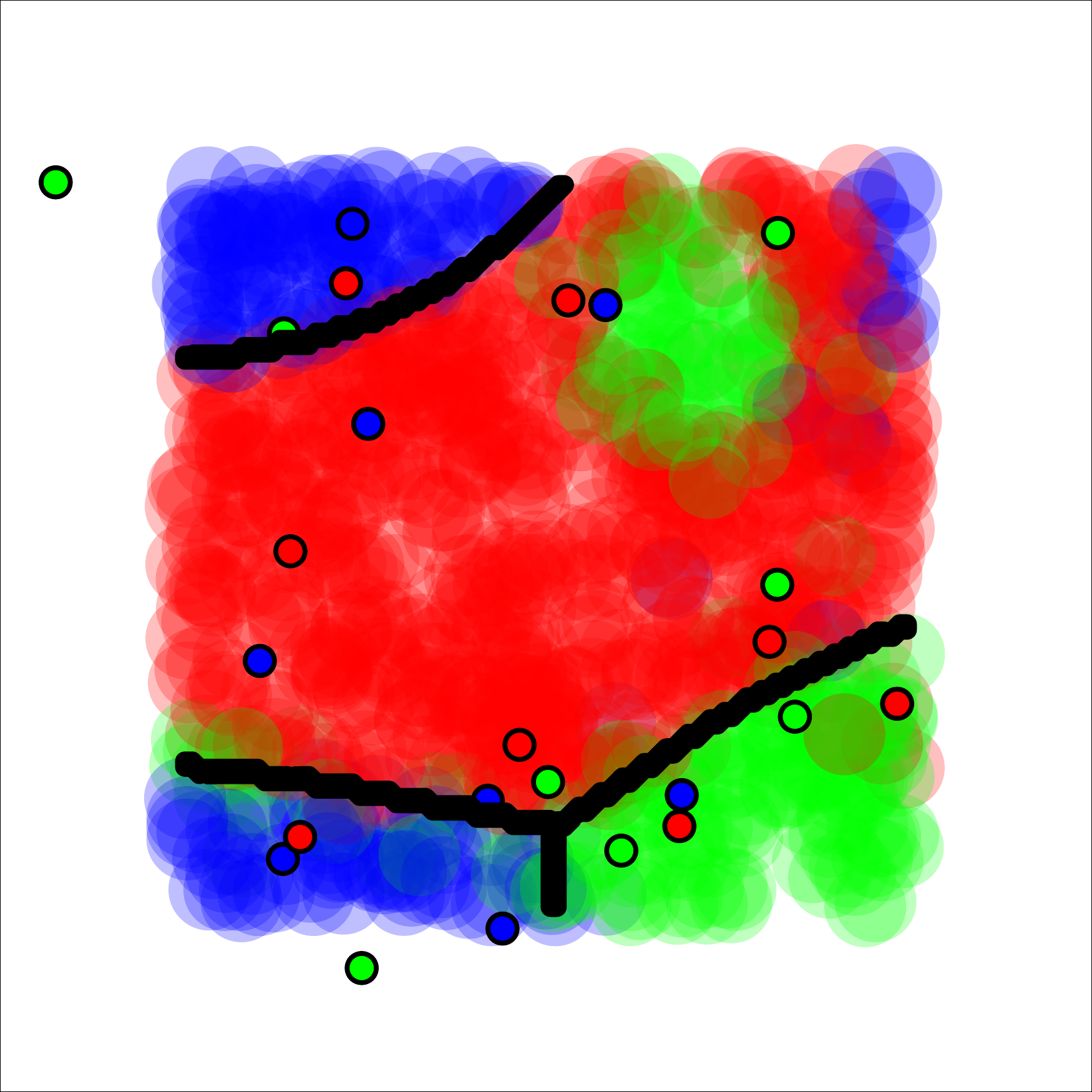}  &
				\includegraphics[width = 1.5cm]{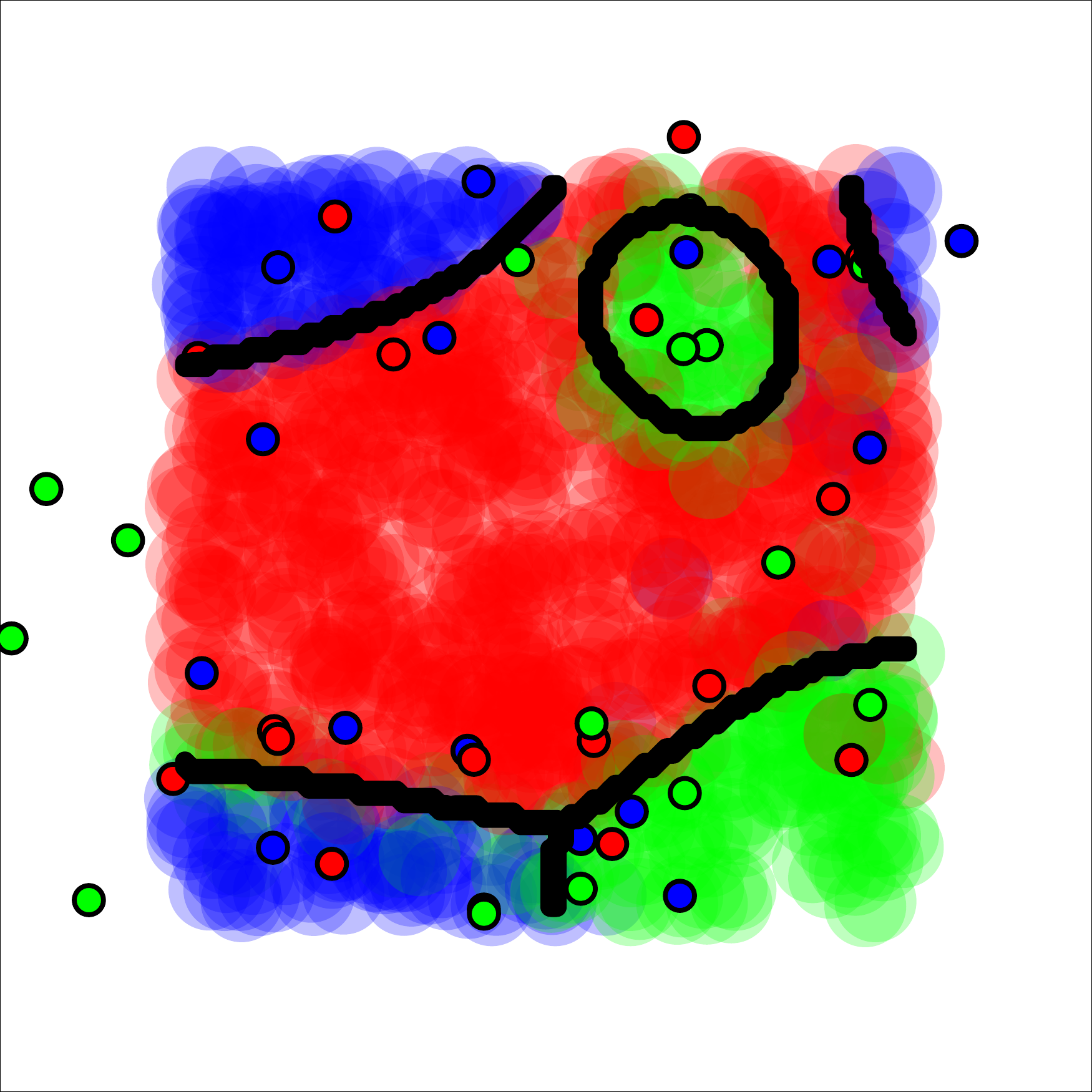}  &
				\includegraphics[width = 1.5cm]{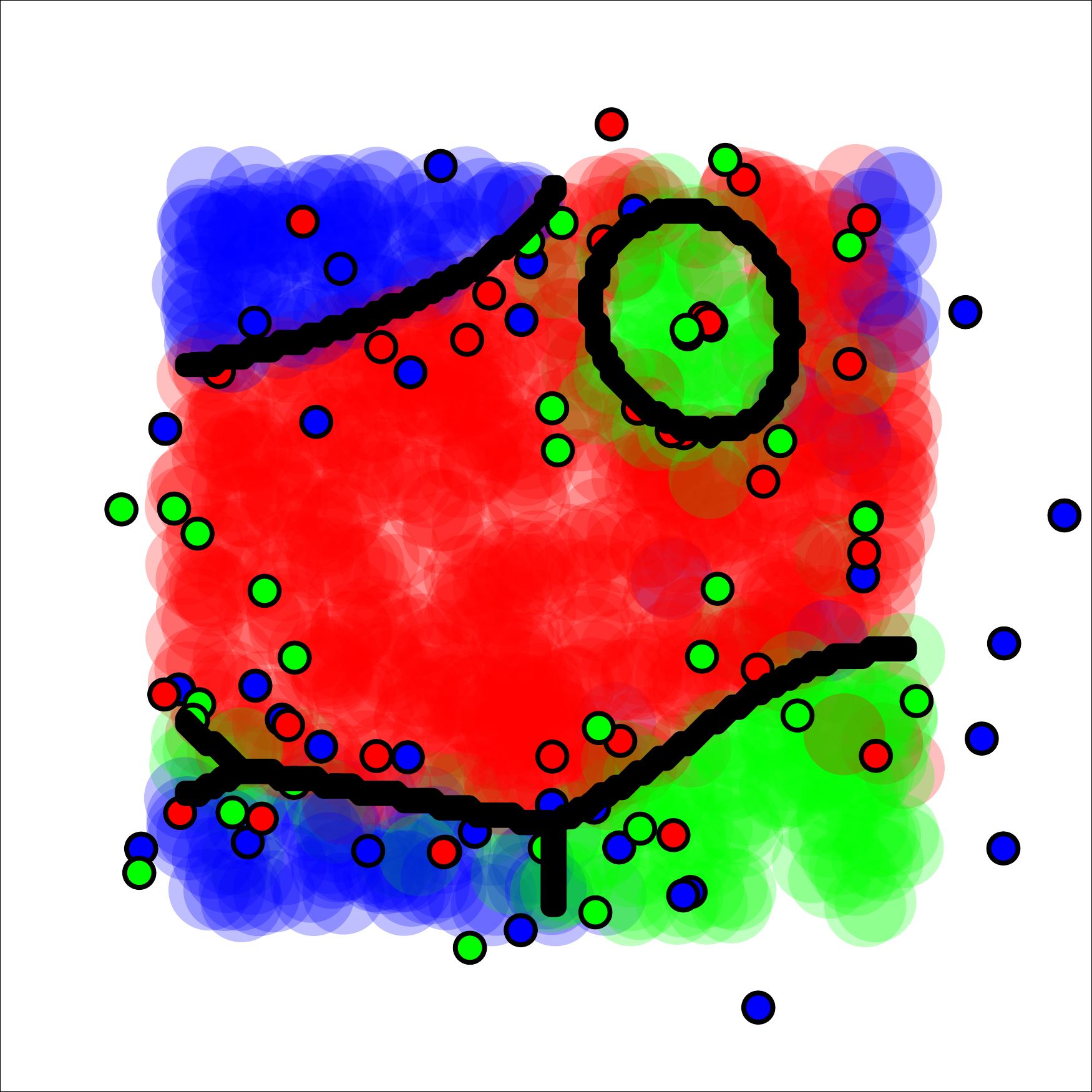}  &
				\includegraphics[width = 1.5cm]{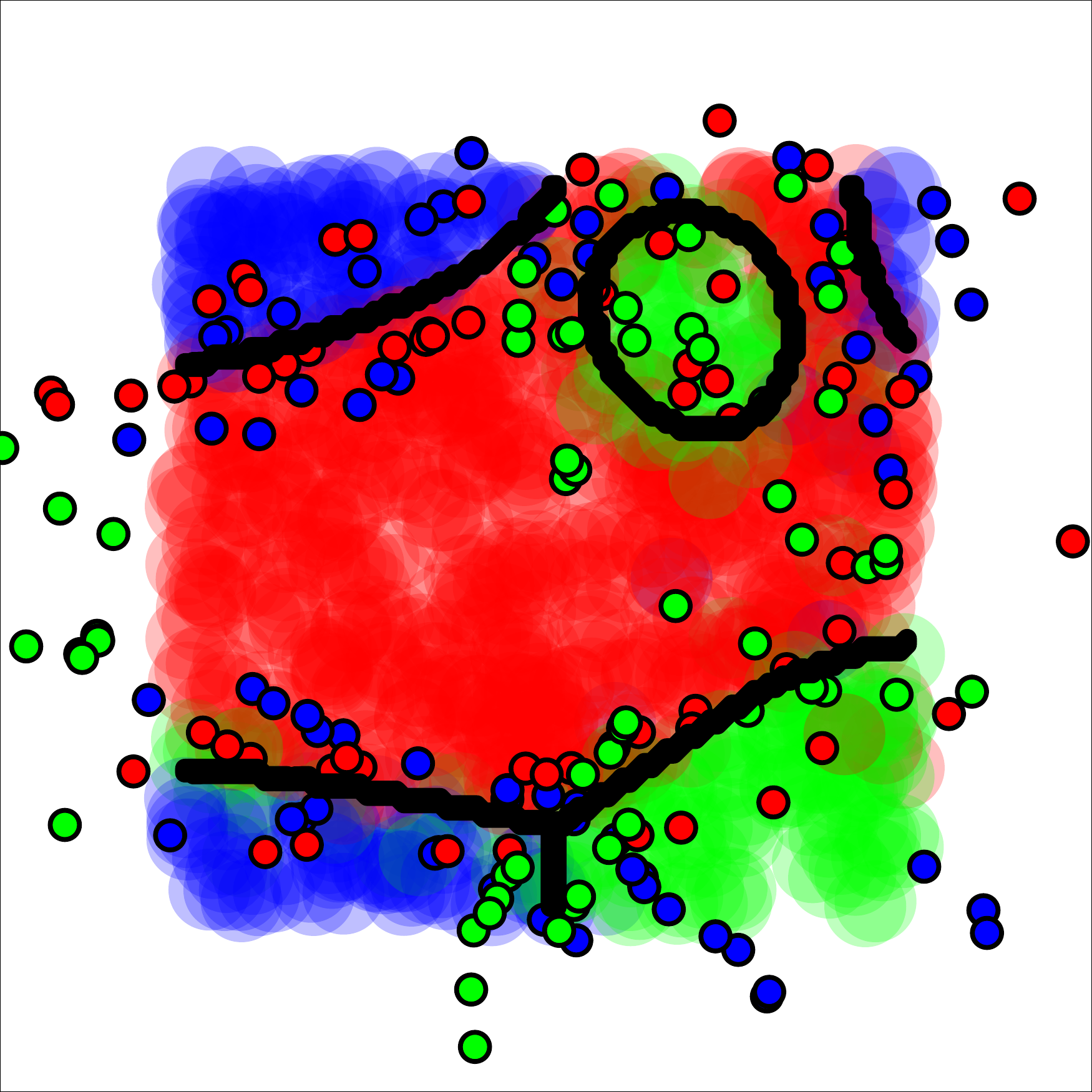}  &
				\includegraphics[width = 1.5cm]{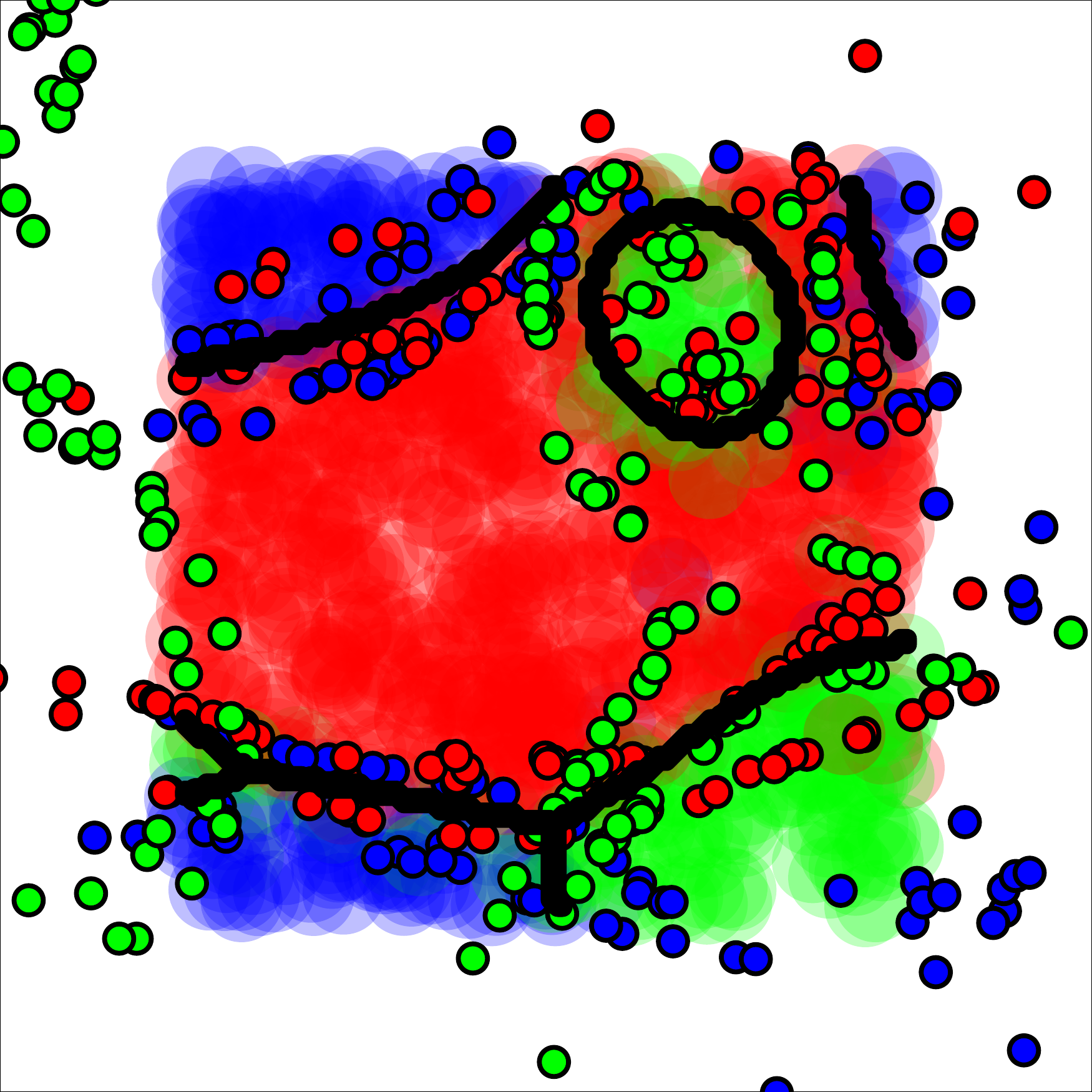} &
				\includegraphics[width = 1.5cm]{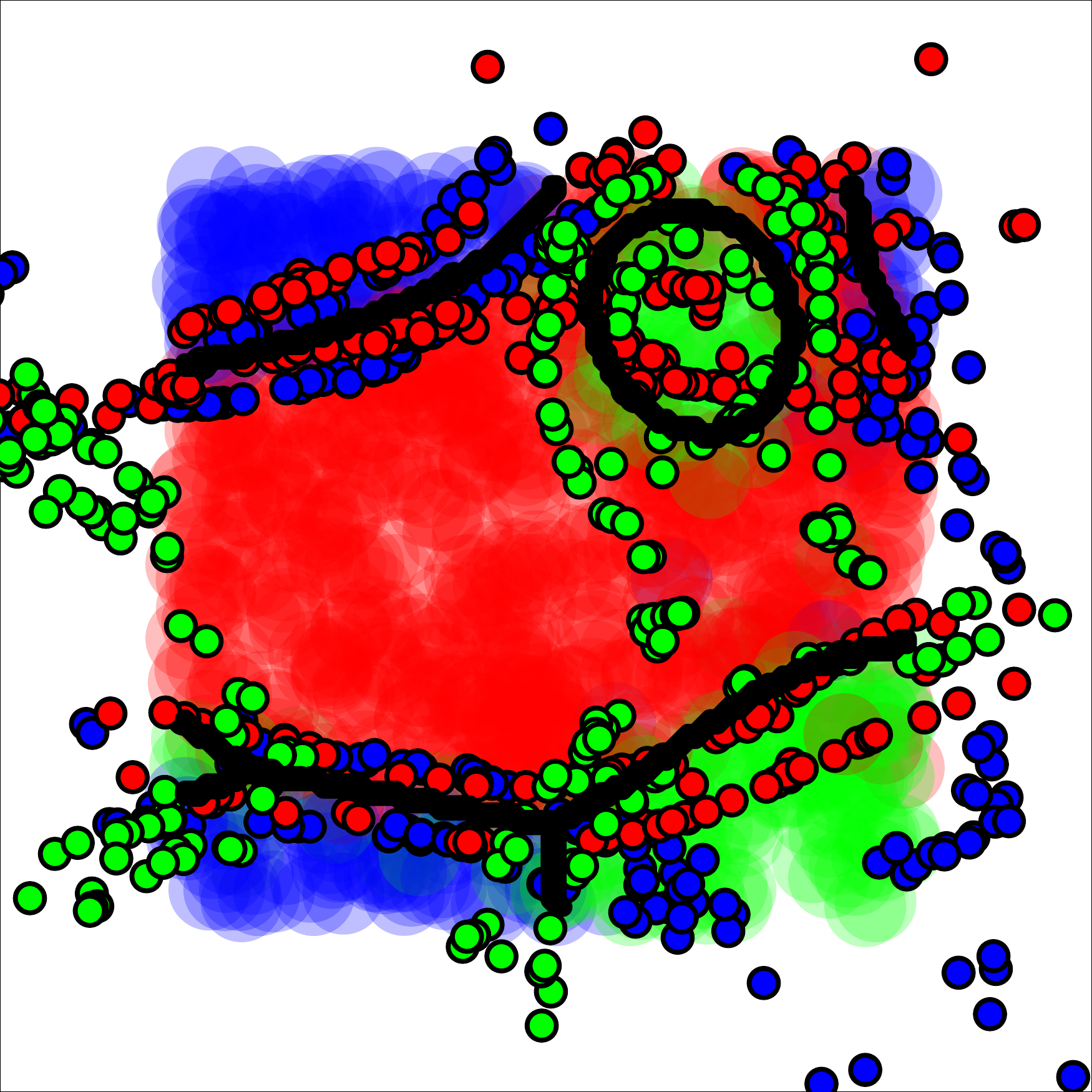}  
				\\

			\end{tabular}
		\end{center}
		\caption{ Decision boundaries and location of the inducing points after training for each method.
			GFITC, EP and SEP seem to place the inducing points one on top of each other. By contrast, VI
			prefers to place them near the decision boundaries. Best seen in color.
		}
		\label{fig:inducing}
	\end{figure*}

	\subsection{Analysis of Inducing Point Learning}
	
	We generate a synthetic two dimensional problem with three
	classes by sampling the latent functions from the GP prior 
	and applying the rule $y_i = \text{arg max}_k \, f^k(\mathbf{x}_i)$. 
	The distribution of $\mathbf{x}_i$ is uniform in the box $[-2.5,2.5]\times[-2.5,2.5]$.
	We consider 1000 training instances and a growing number of inducing points, \emph{i.e.}, 
	$M = 1$ to $M = 256$. The initial location of the inducing points is chosen at 
	random and it is the same for all the methods. We are interested in
	the location of the inducing points after training. Thus, we set the other 
	hyper-parameters to their true values (specified before generating the data) and we keep them
	fixed. All methods but VI are trained using batch methods during 2000 iterations. VI is 
	trained using stochastic gradients for 2000 epochs (the batch version often gets 
	stuck in local optima). We use ADAM with the default settings \citep{kingma2015}, 
	and $100$ as the mini-batch size.
	
	Figure \ref{fig:inducing} shows the location learnt by each method for the inducing points.
	Blue, red and green points represent the training data, black lines are decision 
	boundaries and black border points are the inducing points. As we increase the number of 
	inducing points the methods become more accurate. However, GFITC, EP and SEP 
	identify decision boundaries that are better with a smaller number of inducing points.
	VI fails in this task. This is probably because VI updates the inducing-points with
	a bad estimate of $q$ during the initial iterations. VI uses gradient steps to 
	update $q$, which is less efficient than the EP updates (free of any learning rate).
	GFITC, EP and SEP overlap the inducing points, which can be seen as a pruning 
	mechanism (if two inducing points are equal, it is like having only one). This has already 
	been observed in regression problems \citep{bauer2016}.  By contrast, VI places the inducing 
	points near the decision boundaries. This agrees with previous results on binary classification \citep{HensmanMG15}.
	
	\subsection{Performance as a Function of the Training Time}
	
\begin{figure}[!h]
		\begin{center}
			\centerline{\includegraphics[width=0.75 \columnwidth]{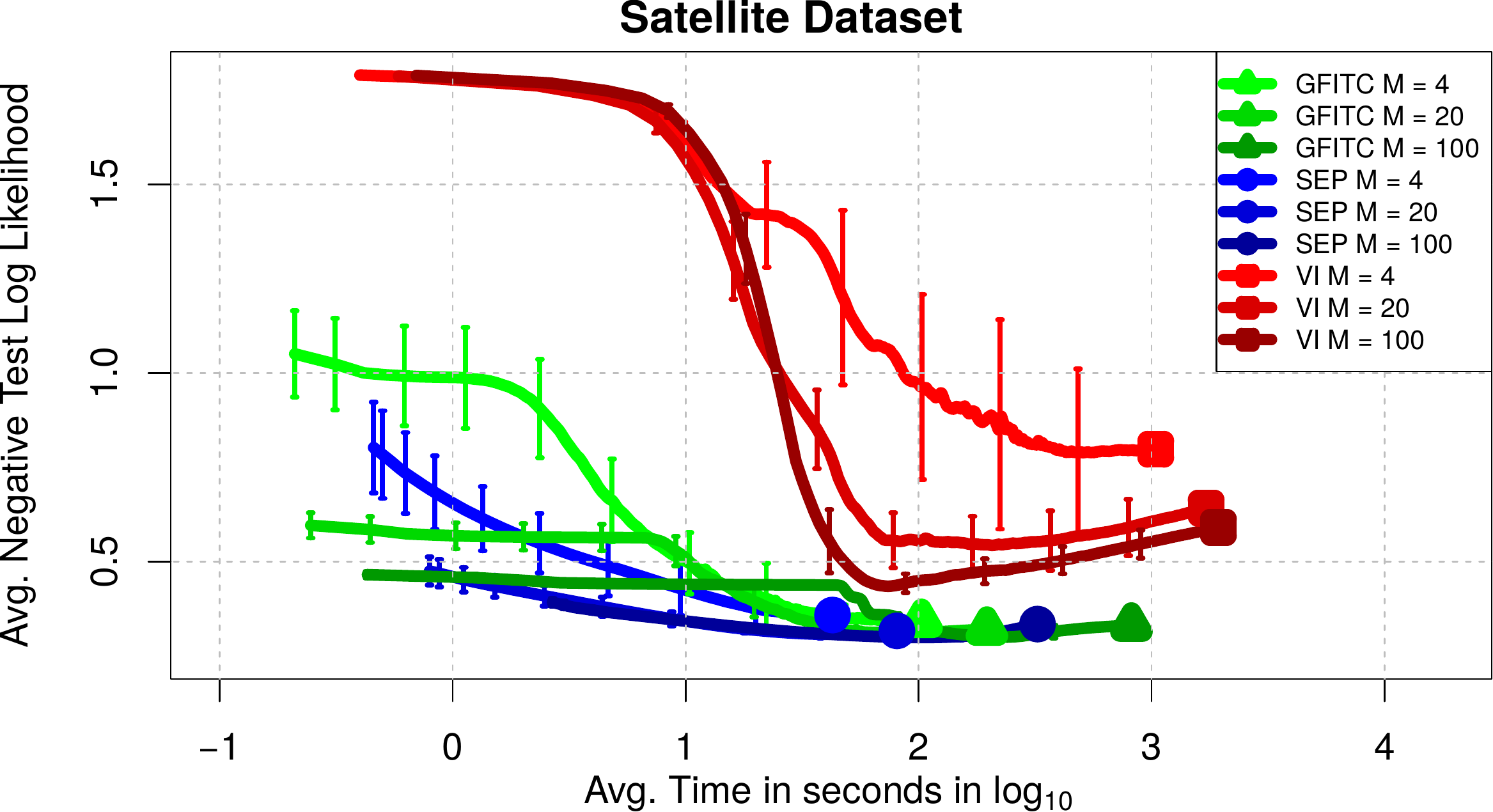}}
			\caption{{\small Negative test log-likelihood for GFITC, SEP and VI on \emph{Satellite} as a function
					of the training time. Best seen in color.}} 
			\label{fig:nll_time}
		\end{center}
	\end{figure}

	Figure \ref{fig:nll_time} shows the negative test log-likelihood of each method as a 
	function of the training time on the Satellite dataset (EP results are not shown 
	since it performs equal to SEP). Training is done as in Section \ref{subsec:uci}. We consider 
	a growing number of inducing points $M = 4, 20, 100$ and report averages over 100 repetitions 
	of the experiments. In all methods we use batch training. 
	We observe that SEP is the method with the best performance at the lowest cost.
	Again, it is faster than GFITC because it optimizes $q$ 
	and the hyper-parameters at the same time, while GFITC waits until EP has converged 
	to update the hyper-parameters. VI is not very efficient 
	for small values of $M$, due to the quadratures.
	It also takes more time to get a good estimate of $q$, which is updated by gradient descent and is
	less efficient than the EP updates. Similar results are obtained in terms of the test error. 
	See the supplementary material. However, in that case VI does not overfit the training data.
	
	\begin{figure*}[htb]
		\begin{center}
			\begin{tabular}{cc}
				\includegraphics[width=0.45\columnwidth]{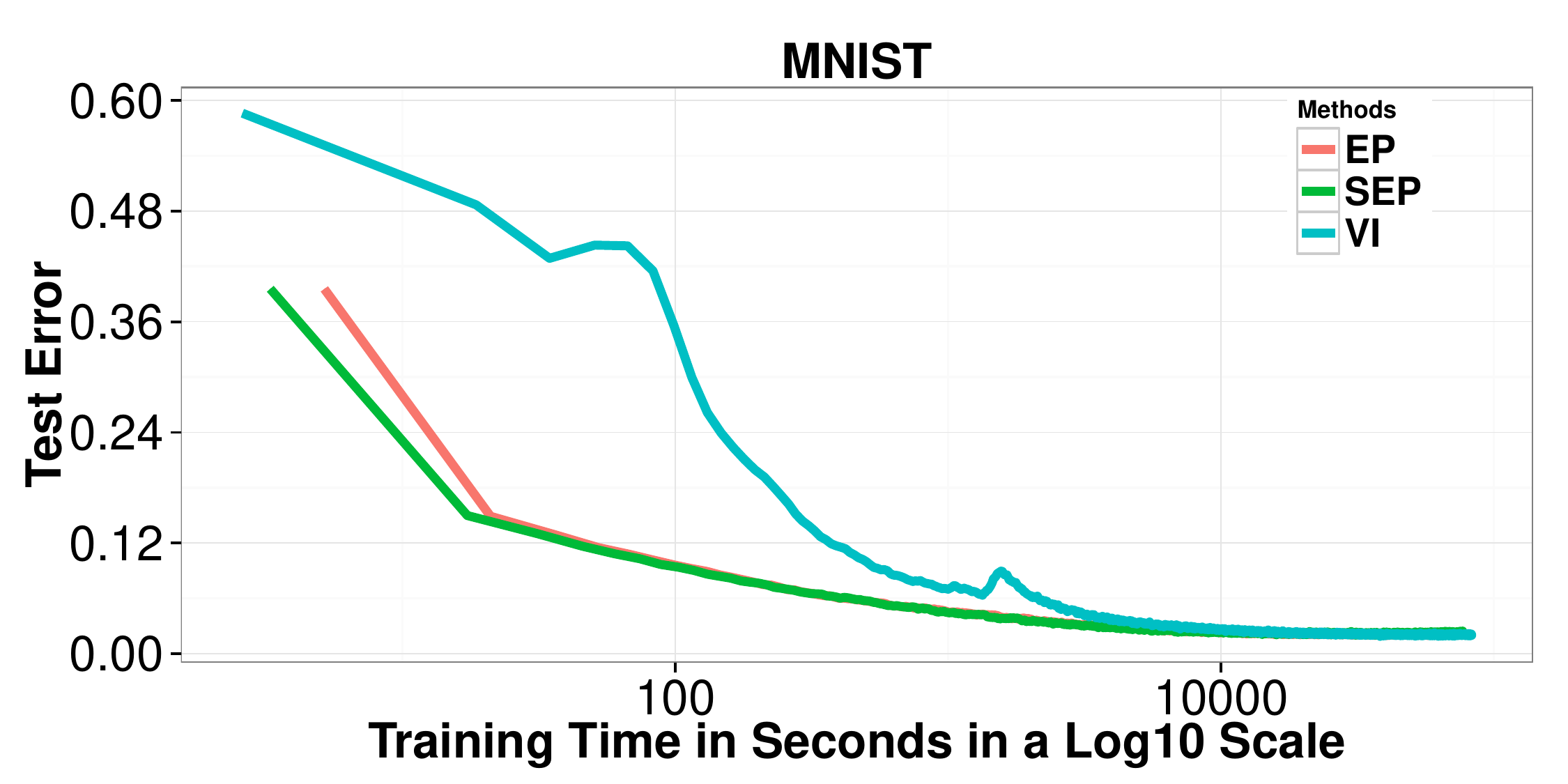} &
				\includegraphics[width=0.45\columnwidth]{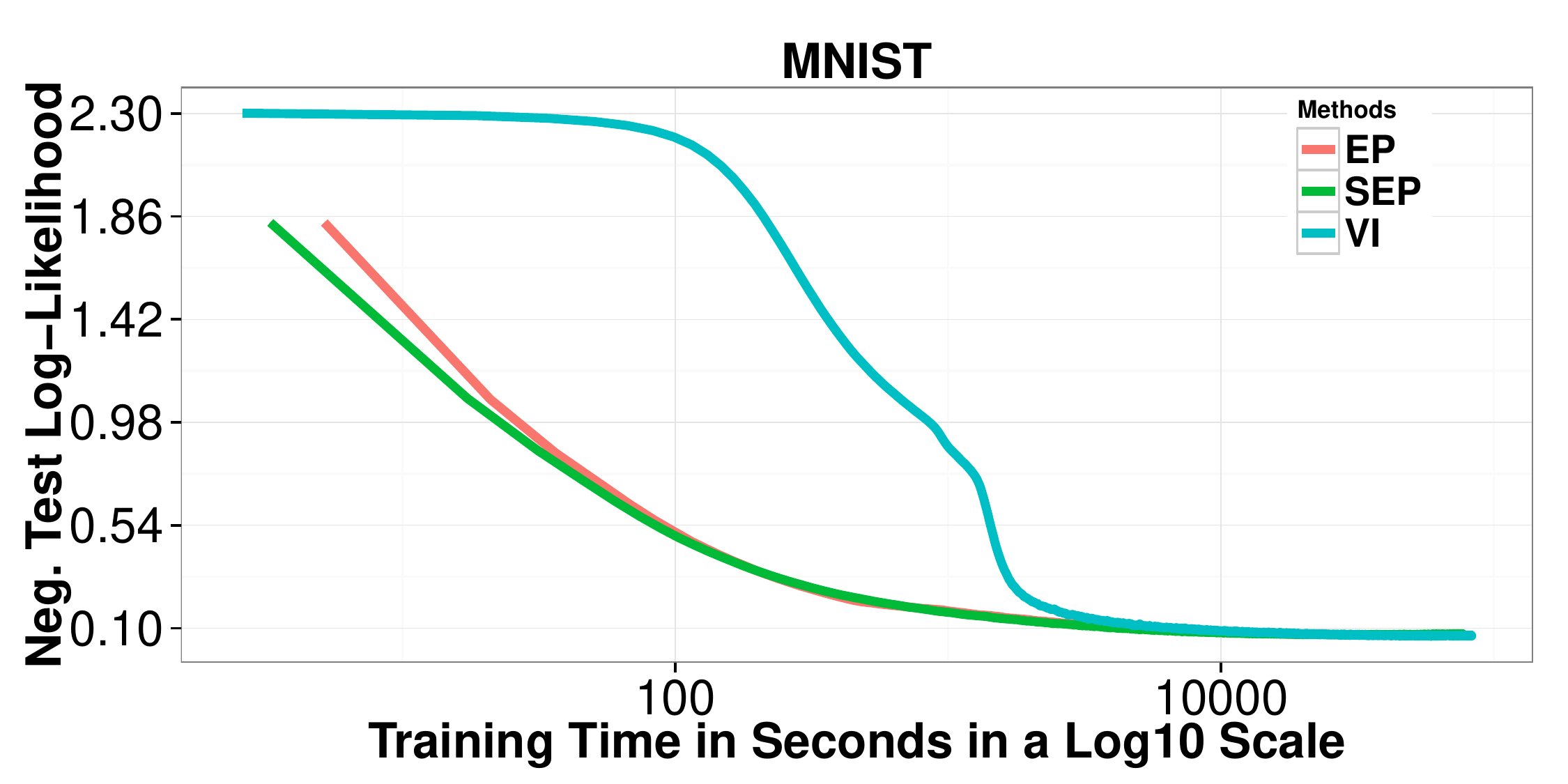}  \\
				\includegraphics[width=0.45\columnwidth]{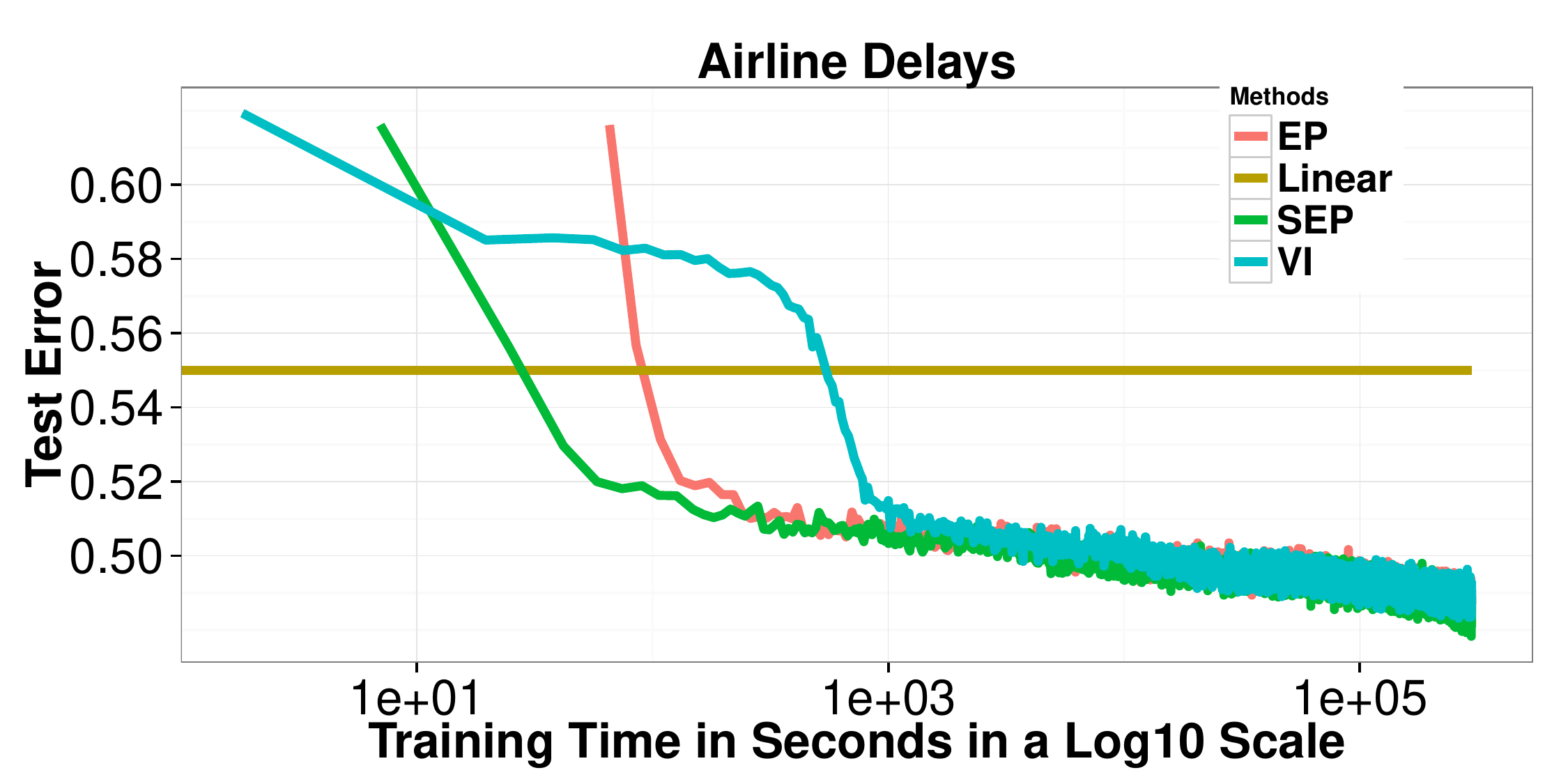} &
				\includegraphics[width=0.45\columnwidth]{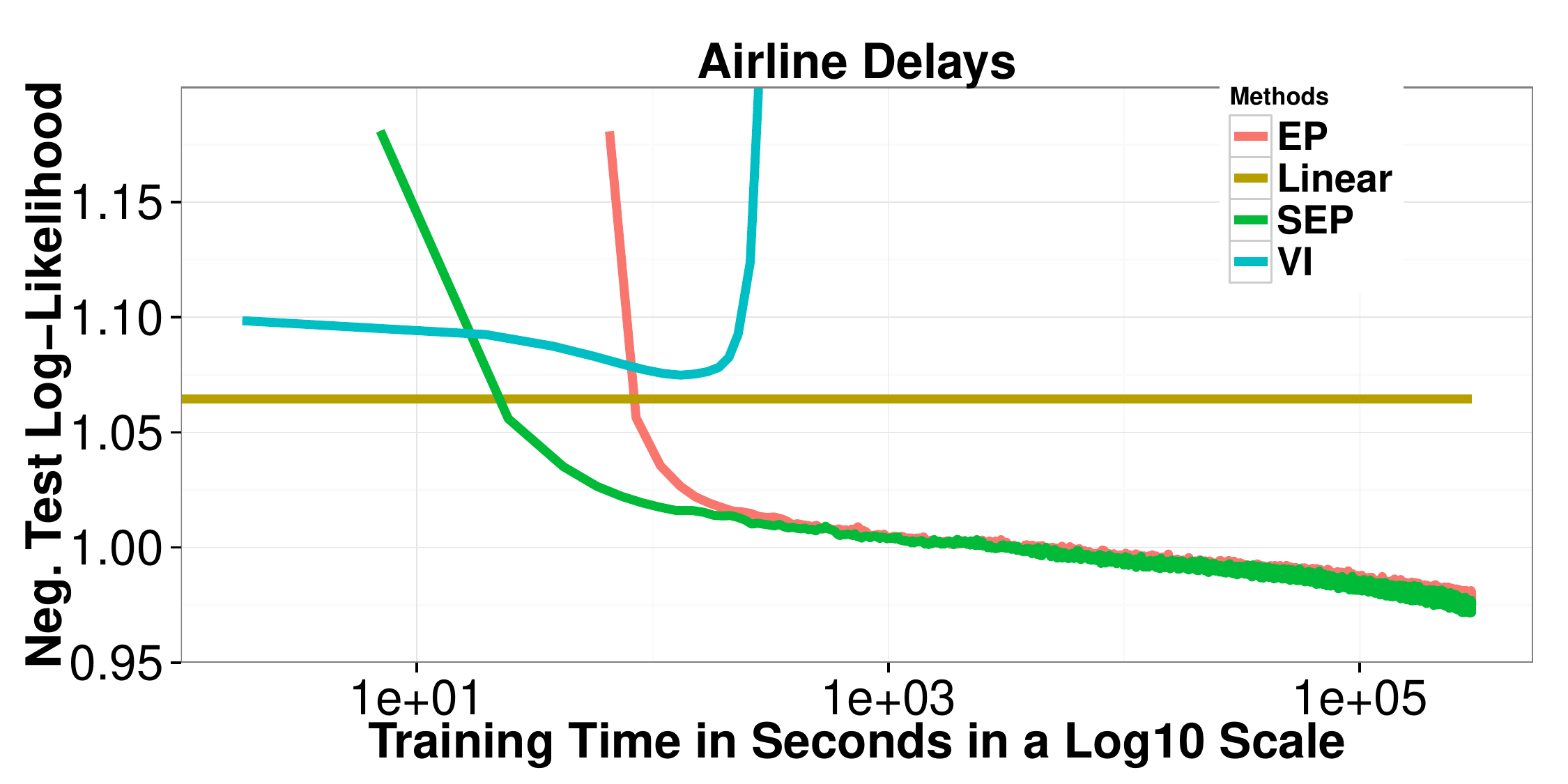}  \\
			\end{tabular}
		\end{center}
		\caption{{\small Average test error and average negative test log-likelihood for each method on the MNIST (top) and the Airline (bottom) 
				dataset. In the Airline dataset a linear model based on logistic regression is included in the comparison. Best seen in color.}}
		\label{fig:mnist_airline}
	\end{figure*}

	\subsection{Performance When Using Stochastic Gradients}
	
	In very large datasets batch training is infeasible, and one must use mini-batches 
	to update $q$ and to approximate the required gradients. We evaluate the 
	performance of each method on the MNIST dataset \citep{lecun1998gradient} with $M = 200$ inducing points 
	and mini-batches with $200$ instances. This dataset has $60,000$ instances for training and 
	$10,000$ for testing. The learning rate of each method is set using ADAM with the default 
	parameters \citep{kingma2015}. GFITC does not allow for stochastic optimization. Thus, it is ignored in 
	the comparison. The test error and the negative test log-likelihood of each method is
	displayed in Figure \ref{fig:mnist_airline} (top) as a function of the training time. In this 
	larger dataset all methods perform similarly. However, EP and SEP take less time to converge than 
	VI. SEP obtains a test error that is $2.08\%$ and average negative test 
	log-likelihood that is $0.0725$. The results of VI are $2.02\%$ and $0.0686$, respectively.
	These results are similar to the ones reported in \citep{HensmanMG15} using $M = 500$.
	
	A last experiment considers all flights within the USA between 01/2008 and 
	04/2008 ({\small \url{http://stat-computing.org/dataexpo/2009}}). The task is to classify
	the flights according to their delay using three classes: On time, more
	than $5$ minutes of delay, or more than $5$ minutes before time. 
	We consider $8$ attributes: age of the aircraft, distance covered, airtime, 
	departure time, arrival time, day of the week, day of the month and month. After removing 
	all instances with missing data $2,127,068$ instances remain, from which $10,000$ 
	are used for testing and the rest for training. We use the same settings as on the MNIST 
	dataset and evaluate each method. The results obtained are shown in Figure \ref{fig:mnist_airline} 
	(bottom). We also report the performance of a logistic regression classifier. 
	Again, all methods perform similarly in terms of test error. However, EP and SEP 
	converge faster and quickly outperform the linear model. Importantly, the negative test 
	log-likelihood of VI starts increasing at some point, which is again probably due to the 
	optimization of $\mathds{E}_{q(\mathbf{f}_i)}[\log p(y_i|\mathbf{f}_i)]$ in  (\ref{eq:lower_bound_vi}). 
	The supplementary material has further evidence supporting this.
	
	\section{Conclusions}
	
	We have proposed the first method for multi-class classification with Gaussian 
	processes, based on expectation propagation (EP), that scales well to very large datasets. 
	Such a method uses the FITC approximation to reduce the number of latent variables in 
	the model from $\mathcal{O}(N)$ to $\mathcal{O}(M)$, where $M \ll N$, and $N$ is the 
	number of data instances. For this, $M$ inducing points are introduced for 
	each latent function in the model. Importantly, the proposed method allows for stochastic 
	optimization as the estimate of the log-marginal-likelihood involves a sum across 
	the data. Moreover, we have also considered a stochastic version of EP (SEP) to reduce the 
	memory usage. When mini-batches and stochastic gradients are used for training, the computational cost 
	of the proposed approach is $\mathcal{O}(C M^3)$, with $C$ the number of classes. 
	The memory cost is $\mathcal{O}(CM^2)$.
	
	We have compared the proposed method with other approaches from 
	the literature based on variational inference (VI) \citep{hensman2015}, and
	with the model considered by \citet{Kim2006}, which has been combined with FITC 
	approximate priors (GFITC) \citep{quinonero2005unifying}. The 
	proposed approach outperforms GFITC in large datasets as this method does not 
	allow for stochastic optimization, and in small datasets it produces similar 
	results. The proposed method, SEP, is slightly faster than VI which also allows
	for stochastic optimization. In particular, VI requires one-dimensional 
	quadratures which in small datasets are expensive. We have also observed that
	SEP converges faster than VI. This is probably because the EP updates, free of 
	any learning rate, are more efficient for finding a good posterior approximation
	than the gradient ascent updates employed by VI. 
	
	An important conclusion of this work is that VI sometimes gives bad predictive 
	distributions in terms of the test log-likelihood. The EP and SEP methods 
	do not seem to have this problem. Thus, if one cares about accurate predictive 
	distributions, VI should be avoided in favor of the proposed methods. In our 
	experiments we have also observed that the proposed approaches tend to place the inducing 
	points one on top of each other, which can be seen as an inducing point pruning 
	technique \citep{bauer2016}. By contrast, VI tends to place them near the decision boundaries.
	
	\section*{Acknowledgements}

	The authors gratefully acknowledge the use of the facilities of Centro de Computaci\'on Cient\'ifica (CCC) at Universidad Aut\'onoma de Madrid. The authors also acknowledge financial support from Spanish Plan Nacional I+D+i, Grants TIN2013-42351-P, TIN2016-76406-P, TIN2015-70308-REDT and TEC2016-81900-REDT (MINECO/FEDER EU), and from Comunidad de Madrid, Grant S2013/ICE-2845.

	\bibliography{references}
	\bibliographystyle{icml2017}

\appendix
	
\input{supplementary.tex}

\end{document}

%% file: supplementary.tex
\section{Introduction}\label{ch:calculations}

In this appendix we give all the details to implement the EP algorithm for the proposed method described in the main manuscript. In particular, we describe how to reconstruct the posterior approximation from the approximate factors and how to refine these factors. We also detail the computation of the EP approximation to the marginal likelihood and its gradients. Finally, we include some additional experimental results.

\section{Reconstruction of the posterior approximation}\label{sect:recPosterior}

In this section we show how to obtain the posterior distribution by multiplying the approximate factors $\tilde{\phi}_i^k(\overline{\mathbf{f}})$ and the prior $p(\overline{\mathbf{f}})$. From the main manuscript we know that these elements have the following form:
\begin{align}
\tilde{\phi}_i^k(\overline{\mathbf{f}}) & = \tilde{s}_i^k \exp
\left\{ 
- \frac{1}{2} (\overline{\mathbf{f}}^{y_i})^\text{T} 
\tilde{\mathbf{V}}_{i,k}^{y_i}
\overline{\mathbf{f}}^{y_i} +  (\overline{\mathbf{f}}^{y_i})^\text{T} \tilde{\mathbf{m}}_{i,k}^{y_i}
\right\} \times \nonumber \\
& \phantom{~~~~} 
\exp 
\left\{
- \frac{1}{2}(\overline{\mathbf{f}}^{k})^\text{T} 
\tilde{\mathbf{V}}_{i,k}
\overline{\mathbf{f}}^{k} +  (\overline{\mathbf{f}}^{k})^\text{T} \tilde{\mathbf{m}}_{i,k}
\right\}\,, \label{eq:approx_factor_ep}\\
p(\overline{\mathbf{f}}) &= \prod_{k=1}^{C} p(\overline{\mathbf{f}}^k|\overline{\mathbf{X}}^k) = \prod_{k=1}^{C} \mathcal{N}(\overline{\mathbf{f}}^k|\mathbf{0}, \mathbf{K}_{\overline{\mathbf{X}}^k\overline{\mathbf{X}}^k}^k)\,,
\end{align}
\noindent where {\scriptsize $\mathbf{K}_{\overline{\mathbf{X}}^k \overline{\mathbf{X}}^k}^k$} is a $M \times M$ matrix with the cross covariances between {\small $\overline{\mathbf{f}}^k$}. $\tilde{\mathbf{V}}_{i,k}^{y_i}$, $\tilde{\mathbf{V}}_c$, $\tilde{\mathbf{m}}_{i,k}^{y_i}$ and $\tilde{\mathbf{m}}_{i,k}$ have the following especial form (see Section \ref{sect:updateFactors} of this document for the detailed derivation):
\begin{align}
& \tilde{\mathbf{V}}_{i,k}^{y_i} = C_{i,k}^{1,y_i} \bm{\upsilon}_{i}^{y_i} (\bm{\upsilon}_{i}^{y_i})^\text{T}\,,\\
& \tilde{\mathbf{V}}_{i,k} = C_{i,k}^{1}\bm{\upsilon}_{i}^{k}(\bm{\upsilon}_{i}^{k})^\text{T}\,,\\
&\tilde{\mathbf{m}}_{i,k}^{y_i}  = C_{i,k}^{2,y_i}\bm{\upsilon}_{i}^{y_i}\,,\\
&\tilde{\mathbf{m}}_{i,k} = C_{i,k}^{2}\bm{\upsilon}_{i}^{k}\,,
\end{align}
where {\small $\bm{\upsilon}_i^k = (\mathbf{k}_{\mathbf{x}_i\overline{\mathbf{X}}^k}^k)^\text{T} 
(\mathbf{K}_{\overline{\mathbf{X}}^k \overline{\mathbf{X}}^k}^k)^{-1} $} and 
$C_{i,k}^{1,y_i}$, $C_{i,k}^{1}$, $C_{i,k}^{2,y_i}$ and $C_{i,k}^{2}$ are parameters found by EP.

We know from the main manuscript that the posterior approximation will have the following form
\begin{align}
q(\overline{\mathbf{f}}) & = \frac{1}{Z_q} \prod_{i=1}^{N} \left[\prod_{k\neq y_i} \tilde{\phi}_i^k(\overline{\mathbf{f}}) \right]\prod_{k=1}^{C}p(\overline{\mathbf{f}}^k|\overline{\mathbf{X}}^k)\,.
\end{align}
Given that all the factors are Gaussian, a distribution that is closed under product and division, $q(\overline{\mathbf{f}})$ is also Gaussian. In particular, the posterior approximation is defined as $q(\overline{\mathbf{f}}) = \prod_{k=1}^{C}\mathcal{N}(\overline{\mathbf{f}}|, \mathbf{m}_k, \mathbf{V}_k)$. The parameters of this distribution can be obtained by using the formulas given in 
the Appendix of \cite{lobatoThesis2010} for the product of two Gaussians, leading to
\begin{align}
\mathbf{V}_k &= \left[(\mathbf{K}_{\overline{\mathbf{X}}^k \overline{\mathbf{X}}^k}^k)^{-1} + 
	\boldsymbol{\Upsilon}_k\bm{\Delta}_k\boldsymbol{\Upsilon}_k^\text{T}\right]^{-1}\,, \nonumber \\
\mathbf{m}_k & = \mathbf{V}_k \boldsymbol{\Upsilon}_k \tilde{\boldsymbol{\mu}}_k \,,
\end{align}
where $\boldsymbol{\Upsilon}_k = (\boldsymbol{\upsilon}_1^k, \ldots, \boldsymbol{\upsilon}_N^k)$ is a $M \times N$ matrix, $\bm{\Delta}_k$ is a diagonal $N\times N$ matrix where each component of the diagonal has the following form
\begin{align}
\Delta_{i,i}^k & = C_{i,k}^{1}\mathds{1}(k\neq y_i) + \left(\sum_{k'\neq y_i}C_{i,k'}^{1,y_i}\right)\mathds{1}(k=y_i)\,,
\end{align}
\noindent where $\mathds{1}(\cdot)$ is an indicator function, with value $1$ if the condition holds and zero otherwise, 
and $\tilde{\boldsymbol{\mu}}_k$ is a vector where each component is defined by
\begin{align}
\tilde{\mu}_i^k & = C_{i,k}^{2}\mathds{1}(k\neq y_i) + \left(\sum_{k'\neq y_i}C_{i,k'}^{2,y_i}\right)\mathds{1}(k=y_i)\,.
\end{align}

\section{Computation of the cavity distribution}

Here we will obtain the expressions for the parameters of the cavity distribution $q^{\backslash i,k}$. This distribution is computed by dividing the posterior approximation by the corresponding approximate factor:
\begin{align}
q(\overline{\mathbf{f}})^{\backslash i,k} & \propto \frac{q(\overline{\mathbf{f}})}{\tilde{\phi}_i^k(\overline{\mathbf{f}})}\,.
\end{align}
Given that all factors are Gaussian, the resulting distribution will also be Gaussian. 
The parameters can be obtained by using again the formulas in the Appendix of \cite{lobatoThesis2010}.
However, because $\tilde{\phi}_i^k$ only depends on $\overline{\mathbf{f}}^{y_i}$ and $\overline{\mathbf{f}}^k$, only these components
of $q(\overline{\mathbf{f}})$ will change. The corresponding parameters of $q(\overline{\mathbf{f}})^{\backslash i,k}$ are:
\begin{align}
\begin{split}
\mathbf{V}_{y_i}^{\backslash i,k} &= (\mathbf{V}_{y_i}^{-1} - \tilde{\mathbf{V}}_{i,k}^{y_i})^{-1}\\
&= (\mathbf{V}_{y_i}^{-1} - C_{i,k}^{1,y_i}\bm{\upsilon}_{i}^{y_i}(\bm{\upsilon}_{i}^{y_i})^\text{T})^{-1}\\
&= \mathbf{V}_{y_i} + \mathbf{V}_{y_i}\bm{\upsilon}_{i}^{y_i}[(C_{i,k}^{1,y_i})^{-1} - \bm{\upsilon}_{i}^{y_i}\mathbf{V}_{y_i}(\bm{\upsilon}_{i}^{y_i})^\text{T}]^{-1}(\bm{\upsilon}_{i}^{y_i})^\text{T}\mathbf{V}_{y_i}\,,
\end{split}\\
\mathbf{V}_k^{\backslash i,k} &= \mathbf{V}_{k} + \mathbf{V}_{k}\bm{\upsilon}_{i}^{k}[(C_{i,k}^{1})^{-1} - \bm{\upsilon}_{i}^{k}\mathbf{V}_{k}(\bm{\upsilon}_{i}^{k})^\text{T}]^{-1}(\bm{\upsilon}_{i}^{k})^\text{T}\mathbf{V}_{k}\,,\\
\begin{split}
\mathbf{m}_{y_i}^{\backslash i,k}  &= \mathbf{V}_{y_i}^{\backslash i,k}(\mathbf{V}_{y_i}^{-1}\mathbf{m}_{y_i} - \tilde{\mathbf{m}}_{i,k}^{y_i})\\
&=\mathbf{V}_{y_i}^{\backslash i,k}(\mathbf{V}_{y_i}^{-1}\mathbf{m}_{y_i} - C_{i,k}^{2,y_i}\bm{\upsilon}_{i}^{y_i})\\
&=\mathbf{V}_{y_i}^{\backslash i,k}\mathbf{V}_{y_i}^{-1}\mathbf{m}_{y_i} - C_{i,k}^{2,y_i} \bm{\upsilon}_{i}^{y_i}\mathbf{V}_{y_i}^{\backslash i,k}\\
&=\mathbf{V}_{y_i}\mathbf{V}_{y_i}^{-1}\mathbf{m}_{y_i}+\mathbf{V}_{y_i}\bm{\upsilon}_{i}^{y_i}[(C_{i,k}^{1,y_i})^{-1} - \bm{\upsilon}_{i}^{y_i}\mathbf{V}_{y_i}(\bm{\upsilon}_{i}^{y_i})^\text{T}]^{-1}(\bm{\upsilon}_{i}^{y_i})^\text{T}\mathbf{V}_{y_i}\mathbf{V}_{y_i}^{-1}\mathbf{m}_{y_i} \\
&\phantom{=~~~}- C_{i,k}^{2,y_i} \bm{\upsilon}_{i}^{y_i}\mathbf{V}_{y_i}^{\backslash i,k}\\
&=\mathbf{m}_{y_i} + \mathbf{V}_{y_i}\bm{\upsilon}_{i}^{y_i}[(C_{i,k}^{1,y_i})^{-1} -\bm{\upsilon}_{i}^{y_i}\mathbf{V}_{y_i}(\bm{\upsilon}_{i}^{y_i})^\text{T}]^{-1}(\bm{\upsilon}_{i}^{y_i})^\text{T}\mathbf{m}_{y_i} - \mathbf{V}_{y_i}\boldsymbol{\upsilon}^\text{T}C_{i,k}^{2,y_i}\\
&\phantom{=~~~} - \mathbf{V}_{y_i}\bm{\upsilon}_{i}^{y_i}[(C_{i,k}^{1,y_i})^{-1} - \bm{\upsilon}_{i}^{y_i}\mathbf{V}_{y_i}(\bm{\upsilon}_{i}^{y_i})^\text{T}]^{-1}(\bm{\upsilon}_{i}^{y_i})^\text{T}\mathbf{V}_{y_i}\bm{\upsilon}_{i}^{y_i}C_{i,k}^{2,y_i}\,,
\end{split}\\
\begin{split}
\mathbf{m}_k^{\backslash i,k}  &= \mathbf{m}_{k} + \mathbf{V}_{k}\bm{\upsilon}_{i}^{k}[(C_{i,k}^{1})^{-1} -\bm{\upsilon}_{i}^{k}\mathbf{V}_{k}(\bm{\upsilon}_{i}^{k})^\text{T}]^{-1}(\bm{\upsilon}_{i}^{k})^\text{T}\mathbf{m}_{k} - \mathbf{V}_{k}\boldsymbol{\upsilon}^\text{T}C_{i,k}^{2}\\
&\phantom{=~~~} - \mathbf{V}_{k}\bm{\upsilon}_{i}^{k}[(C_{i,k}^{1})^{-1} - \bm{\upsilon}_{i}^{k}\mathbf{V}_{k}(\bm{\upsilon}_{i}^{k})^\text{T}]^{-1}(\bm{\upsilon}_{i}^{k})^\text{T}\mathbf{V}_{k}\bm{\upsilon}_{i}^{k}C_{i,k}^{2}\,,
\end{split}
\end{align}
where we have used the Woodbury matrix identity and  $\bm{\upsilon}_{i}^{k}$, $\bm{\upsilon}_{i}^{y_i}$, $C_{i,k}^{1,y_i}$, $C_{i,k}^{1}$, $C_{i,k}^{2,y_i}$ and $C_{i,k}^{2}$ are the parameters specified in Section \ref{sect:recPosterior}.

\section{Update of the approximate factors}\label{sect:updateFactors}

In this section we show how to find the approximate factors $\tilde{\phi}_i^k$ once the cavity distribution $q^{\backslash i,k}$ has already been computed. We know from the main manuscript that the exact factors are:
\begin{align}
\phi_i^k(\overline{\mathbf{f}}) & = \Phi(\alpha_i^k) = \Phi\left(\frac{{m}_i^{y_i} - {m}_i^{k}}{\sqrt{{s}_i^{y_i} + {s}_i^k}}\right)\,,
\end{align}
where $\Phi(\cdot)$ is the c.d.f. of a standard Gaussian distribution and
$m_i^{y_i}$, $m_i^k$, $s_i^{y_i}$ and $s_i^k$ are defined in the main manuscript.
The normalization constant of $\phi_i^kq^{\backslash i,k}$ has the following form:
\begin{align}
\begin{split}
Z_{i,k} &= \int \Phi\left(\frac{{m}_i^{y_i} - {m}_i^{k}}{\sqrt{{s}_i^{y_i} + {s}_i^k}}\right)\mathcal{N}(\overline{\mathbf{f}}^{y_i}|\mathbf{m}^{\backslash i,k}_{y_i}, \mathbf{V}^{\backslash i,k}_{y_i})\mathcal{N}(\overline{\mathbf{f}}^c|\mathbf{m}_k^{\backslash i,k}, \mathbf{V}_k^{\backslash i,k})d\overline{\mathbf{f}}^{y_i}d\overline{\mathbf{f}}^c\\
&=\Phi\left(\frac{a_i^{y_i} - a_i^{k}}{\sqrt{{b}_i^{y_i} + {b}_i^k}}\right)\,,
\end{split}
\end{align}
where:
\begin{align}
& a_i^{y_i} = (\mathbf{k}_{\mathbf{x}_i\overline{\mathbf{X}}^{y_i}}^{y_i})^\text{T}(\mathbf{K}_{\overline{\mathbf{X}}^{y_i}\overline{\mathbf{X}}^{y_i}}^{y_i})^ {-1}\mathbf{m}^{\backslash i,k}_{y_i} = \left(\bm{\upsilon}_{i}^{y_i} \right)^\text{T} \mathbf{m}^{\backslash i,k}_{y_i}\,,\\
& a_i^{k} = (\mathbf{k}_{\mathbf{x}_i\overline{\mathbf{X}}^{y_i}}^{y_i})^\text{T}(\mathbf{K}_{\overline{\mathbf{X}}^k\overline{\mathbf{X}}^k}^{k})^ {-1}\mathbf{m}_k^{\backslash i,k} = \left( \bm{\upsilon}_{i}^{k} \right)^\text{T} \mathbf{m}_k^{\backslash i,k}\,,\\
& b_i^{y_i} = \kappa_{\mathbf{x}_i\mathbf{x}_i}^{y_i} - (\mathbf{k}_{\mathbf{x}_i\overline{\mathbf{X}}^{y_i}}^{y_i})^\text{T}(\mathbf{K}_{\overline{\mathbf{X}}^{y_i}\overline{\mathbf{X}}^{y_i}}^{y_i})^{ -1}\mathbf{k}_{\mathbf{x}_i\overline{\mathbf{X}}^{y_i}}^{y_i} +  (\mathbf{k}_{\mathbf{x}_i\overline{\mathbf{X}}^{y_i}}^{y_i})^\text{T}(\mathbf{K}_{\overline{\mathbf{X}}^{y_i}\overline{\mathbf{X}}^{y_i}}^{y_i})^{ -1}\mathbf{V}^{\backslash i,k}_{y_i}(\mathbf{K}_{\overline{\mathbf{X}}^{y_i}\overline{\mathbf{X}}^{y_i}}^{y_i})^{ -1}\mathbf{k}_{\mathbf{x}_i\overline{\mathbf{X}}^{y_i}}^{y_i}\,,\\
& b_i^k = \kappa_{\mathbf{x}_i\mathbf{x}_i}^{k} - (\mathbf{k}_{\mathbf{x}_i\overline{\mathbf{X}}^{k}}^{k})^\text{T}(\mathbf{K}_{\overline{\mathbf{X}}^k\overline{\mathbf{X}}^k}^{k})^{ -1}\mathbf{k}_{\mathbf{x}_i\overline{\mathbf{X}}^{k}}^{k} +  (\mathbf{k}_{\mathbf{x}_i\overline{\mathbf{X}}^{k}}^{k})^\text{T}(\mathbf{K}_{\overline{\mathbf{X}}^k\overline{\mathbf{X}}^k}^{k})^{ -1}\mathbf{V}_k^{\backslash i,k}(\mathbf{K}_{\overline{\mathbf{X}}^k\overline{\mathbf{X}}^k}^{k})^{ -1}\mathbf{k}_{\mathbf{x}_i\overline{\mathbf{X}}^{k}}^{k}\,.
\end{align}

Now, that we have an expression for $Z_{i,k}$, we can compute the moments of $\phi_i^kq^{\setminus i,k}$ by getting the
derivatives of $\log Z_{i,k}$ with respect to the parameters of $q^{\backslash i,k}$, as indicated in the Appendix 
of \cite{lobatoThesis2010}. These derivatives are:
\begin{align}
& \frac{\partial\log Z_{i,k}}{\partial \mathbf{m}_{y_i}^{\backslash i,k}} = \frac{\mathcal{N}(\hat{\alpha}_{i,k}|0,1)}{\Phi(\hat{\alpha}_{i,k})} \frac{1}{\sqrt{b_i^{y_i}+b_i^k}}\bm{\upsilon}_{i}^{y_i} = \frac{\beta_{i,k}}{\sqrt{b_i^{y_i}+b_i^k}}\bm{\upsilon}_{i}^{y_i}\,,\\
& \frac{\partial\log Z_{i,k}}{\partial \mathbf{m}_k^{\backslash i,k}} = \frac{-\beta_{i,k}}{\sqrt{b_i^{y_i}+b_i^k}}\bm{\upsilon}_{i}^{k}\,,\\
&\frac{\partial\log Z_{i,k}}{\partial \mathbf{V}_{y_i}^{\backslash i,k}} = -\frac{1}{2}\frac{\mathcal{N}(\hat{\alpha}_{i,k}|0,1)}{\Phi(\hat{\alpha}_{i,k})} \frac{a_i^{y_i}-a_i^{k}}{\sqrt{b_i^{y_i}+b_i^k}}\frac{1}{b_i^{y_i}+b_i^k} = -\frac{1}{2}\beta_{i,k}\hat{\alpha}_{i,k}\frac{1}{b_i^{y_i}+b_i^k}\bm{\upsilon}_{i}^{y_i}(\bm{\upsilon}_{i}^{y_i})^\text{T}\,,\\
&\frac{\partial\log Z_{i,k}}{\partial \mathbf{V}_k^{\backslash i,k}} = -\frac{1}{2}\frac{N(\hat{\alpha}_{i,k}|0,1)}{\Phi(\hat{\alpha}_{i,k})} \frac{a_i^{y_i}-a_i^{k}}{\sqrt{b_i^{y_i}+b_i^k}} \frac{1}{b_i^{y_i}+b_i^k} = -\frac{1}{2}\beta_{i,k}\hat{\alpha}_{i,k}\frac{1}{b_i^{y_i}+b_i^k}\bm{\upsilon}_{i}^{k}(\bm{\upsilon}_{i}^{k})^\text{T}\,,
\end{align}
where $\hat{\alpha}_i^k = (a_i^{y_i} - a_i^{k}) / \sqrt{{b}_i^{y_i} + {b}_i^k}$ and $\beta_i^k = \mathcal{N}(\hat{\alpha}_i^k|0,1) / \Phi(\hat{\alpha}_i^k)$. By following the Appendix of \cite{lobatoThesis2010} we can obtain the moments of $\phi_i^kq^{\backslash i,k}$ (means $\hat{\mathbf{m}}_{y_i}$, $\hat{\mathbf{m}}_{c}$ and covariances $\hat{\mathbf{V}}_{y_i}$, $\hat{\mathbf{V}}_{c}$) from the derivatives of $\log Z_{i,k}$ with respect to the parameters of $q^{\backslash i,k}$. Namely:
\begin{align}
\hat{\mathbf{m}}_{i,k}^{y_i} &= \mathbf{m}_{y_i}^{\backslash i,k} + \mathbf{V}_{y_i}^{\backslash i,k} \frac{\partial \log Z_{i,k}}{\partial \mathbf{m}_{y_i}^{\backslash i,k}} = \mathbf{m}_{y_i}^{\backslash i,k} + \mathbf{V}_{y_i}^{\backslash i,k} \frac{\beta_{i,k}}{\sqrt{b_i^{y_i}+b_i^k}}\bm{\upsilon}_{i}^{y_i}\\
\hat{\mathbf{m}}_{i,k} &= \mathbf{m}_k^{\backslash i,k} + \mathbf{V}_k^{\backslash i,k} \frac{\partial \log Z_{i,k}}{\partial \mathbf{m}_k^{\backslash i,k}} = \mathbf{m}_k^{\backslash i,k} - \mathbf{V}_k^{\backslash i,k} \frac{\beta_{i,k}}{\sqrt{b_i^{y_i}+b_i^k}}\bm{\upsilon}_{i}^{k}\\
\begin{split}
\hat{\mathbf{V}}_{i,k}^{y_i} &= \mathbf{V}_{y_i}^{\backslash i,k} - \mathbf{V}_{y_i}^{\backslash i,k} \left(\left(\frac{\partial\log Z_{i,k}}{\partial \mathbf{m}_{y_i}^{\backslash i,k}}\right)\left(\frac{\partial\log Z_{i,k}}{\partial \mathbf{m}_{y_i}^{\backslash i,k}}\right)^\text{T} - 2 \frac{\partial\log Z_{i,k}}{\partial \mathbf{V}_{y_i}^{\backslash i,k}}\right) \mathbf{V}_{y_i}^{\backslash i,k}\\
& = \mathbf{V}_{y_i}^{\backslash i,k} - \mathbf{V}_{y_i}^{\backslash i,k} \left[\frac{\beta_{i,k}^2}{b_i^{y_i}+b_i^k}\bm{\upsilon}_{i}^{y_i}(\bm{\upsilon}_{i}^{y_i})^\text{T} + \frac{\beta_{i,k}\hat{\alpha}_{i,k}}{b_i^{y_i}+b_i^k}\bm{\upsilon}_{i}^{y_i}(\bm{\upsilon}_{i}^{y_i})^\text{T}\right] \mathbf{V}_{y_i}^{\backslash i,k}\\
& = \mathbf{V}_{y_i}^{\backslash i,k} - \mathbf{V}_{y_i}^{\backslash i,k} \left[\frac{\beta_{i,k}^2 + \beta_{i,k}\hat{\alpha}_{i,k}}{b_i^{y_i}+b_i^k}\bm{\upsilon}_{i}^{y_i}(\bm{\upsilon}_{i}^{y_i})^\text{T}\right] \mathbf{V}_{y_i}^{\backslash i,k}
\end{split}\\
\begin{split}
\hat{\mathbf{V}}_{i,k} &= \mathbf{V}_k^{\backslash i,k} - \mathbf{V}_k^{\backslash i,k} \left(\left(\frac{\partial\log Z_{i,k}}{\partial \mathbf{m}_k^{\backslash i,k}}\right)\left(\frac{\partial\log Z_{i,k}}{\partial \mathbf{m}_k^{\backslash i,k}}\right)^\text{T} - 2 \frac{\partial\log Z_{i,k}}{\partial \mathbf{V}_k^{\backslash i,k}}\right) \mathbf{V}_k^{\backslash i,k}\\
& = \mathbf{V}_k^{\backslash i,k} - \mathbf{V}_k^{\backslash i,k} \left[\frac{\beta_{i,k}^2 + \beta_{i,k}\hat{\alpha}_{i,k}}{b_i^{y_i}+b_i^k}\bm{\upsilon}_{i}^{k}(\bm{\upsilon}_{i}^{k})^\text{T}\right] \mathbf{V}_k^{\backslash i,k}\,.
\end{split}
\end{align}

Now we can find the parameters of the approximate factor $\tilde{\phi}_i^k$, which is obtained as $\tilde{\phi}_i^k = Z_{i,k} q^\text{new}/q^{\backslash i,k}$, where $q^\text{new}$ is a Gaussian distribution with the parameters of $\phi_i^kq^{\backslash i,k}$ just computed. By following the equations given in the Appendix of \cite{lobatoThesis2010} we obtain the precision matrices of the approximate factor:
\begin{align}
\tilde{\mathbf{V}}_{i,k}^{y_i} & = (\hat{\mathbf{V}}_{i,k}^{y_i})^{-1} - (\mathbf{V}_{y_i}^{\backslash i,k})^{-1} \nonumber \\
&= \left(\mathbf{V}_{y_i}^{\backslash i,k} - \mathbf{V}_{y_i}^{\backslash i,k} \bm{\upsilon}_{i}^{y_i} \left[\frac{\beta_{i,k}^2+\beta_{i,k}\hat{\alpha}_{i,k}}{b_i^{y_i}+b_i^k} \right](\bm{\upsilon}_{i}^{y_i})^\text{T} \mathbf{V}_{y_i}^{\backslash i,k}\right)^{-1} - (\mathbf{V}_{y_i}^{\backslash i,k})^{-1} \nonumber \\
& \quad  (\mathbf{V}_{y_i}^{\backslash i,k})^{-1} + (\mathbf{V}_{y_i}^{\backslash i,k})^{-1}\mathbf{V}_{y_i}^{\backslash i,k}\bm{\upsilon}_{i}^{y_i} \Bigg(\left[\frac{\beta_{i,k}^2+\beta_{i,k}\hat{\alpha}_{i,k}}{b_i^{y_i}+b_i^k}\right]^{-1} -
(\bm{\upsilon}_{i}^{y_i})^\text{T}\mathbf{V}_{y_i}^{\backslash i,k}(\mathbf{V}_{y_i}^{\backslash i,k})^{-1}\mathbf{V}_{y_i}^{\backslash i,k} \bm{\upsilon}_{i}^{y_i}\Bigg)^{-1}  \nonumber \\
& \quad (\bm{\upsilon}_{i}^{y_i})^\text{T}\mathbf{V}_{y_i}^{\backslash i,k}(\mathbf{V}_{y_i}^{\backslash i,k})^{-1} - (\mathbf{V}_{y_i}^{\backslash i,k})^{-1} \nonumber \\
&= \bm{\upsilon}_{i}^{y_i} \left(\left[\frac{\beta_{i,k}^2+\beta_{i,k}\hat{\alpha}_{i,k}}{b_i^{y_i}+b_i^k}\right]^{-1} - (\bm{\upsilon}_{i}^{y_i})^\text{T}\mathbf{V}_{y_i}^{\backslash i,k} \bm{\upsilon}_{i}^{y_i}\right)^{-1}(\bm{\upsilon}_{i}^{y_i})^\text{T}\,,
\end{align}
where we have used the Woodbury matrix identity to compute $(\hat{\mathbf{V}}_{i,k}^{y_i})^{-1}$. 
Let us define $C_{i,k}^{1,y_i}$ and $C_{i,k}^{1}$ as:
\begin{align}
&C_{i,k}^{1,y_i} = \left(\left[\frac{\beta_{i,k}^2+\beta_{i,k}\hat{\alpha}_{i,k}}{b_i^{y_i}+b_i^k}\right]^{-1} - (\bm{\upsilon}_{i}^{y_i})^\text{T}\mathbf{V}_{y_i}^{\backslash i,k} \bm{\upsilon}_{i}^{y_i}\right)^{-1}\,,\\
&C_{i,k}^{1} = \left(\left[\frac{\beta_{i,k}^2+\beta_{i,k}\hat{\alpha}_{i,k}}{b_i^{y_i}+b_i^k}\right]^{-1} - (\bm{\upsilon}_{i}^{k})^\text{T}\mathbf{V}_k^{\backslash i,k} \bm{\upsilon}_{i}^{k}\right)^{-1}\,.
\end{align}
The precision matrices of the approximate factors will be then:
\begin{equation}
\begin{split}
& \tilde{\mathbf{V}}_{i,k}^{y_i} = C_{i,k}^{1,y_i}\bm{\upsilon}_{i}^{y_i}(\bm{\upsilon}_{i}^{y_i})^\text{T}\,,\\
& \tilde{\mathbf{V}}_{i,k} = C_{i,k}^{1}\bm{\upsilon}_{i}^{k}(\bm{\upsilon}_{i}^{k})^\text{T}\,.
\end{split}
\end{equation}

For the first natural parameter we proceed in a similar way
\begin{equation}
\begin{split}
\tilde{\mathbf{m}}_{i,k}^{y_i} &= (\hat{\mathbf{V}}_{i,k}^{y_i})^{-1}\hat{\mathbf{m}}_{i,k}^{y_i} - (\mathbf{V}_{y_i}^{\backslash i,k})^{-1}\mathbf{m}_{y_i}^{\backslash i,k}\\
&= ((\mathbf{V}_{y_i}^{\backslash i,k})^{-1}+\tilde{\mathbf{V}}_{i,k}^{y_i})\hat{\mathbf{m}}_{i,k}^{y_i} - (\mathbf{V}_{y_i}^{\backslash i,k})^{-1}\mathbf{m}_{y_i}^{\backslash i,k}\\
&= (\mathbf{V}_{y_i}^{\backslash i,k})^{-1}\hat{\mathbf{m}}_{i,k}^{y_i}+\tilde{\mathbf{V}}_{i,k}^{y_i}\hat{\mathbf{m}}_{i,k}^{y_i} - (\mathbf{V}_{y_i}^{\backslash i,k})^{-1}\mathbf{m}_{y_i}^{\backslash i,k}\\
&\begin{split}
= (\mathbf{V}_{y_i}^{\backslash i,k})^{-1}\Bigg[\mathbf{m}_{y_i}^{\backslash i,k} &+ \mathbf{V}_{y_i}^{\backslash i,k} \frac{\beta_{i,k}}{\sqrt{b_i^{y_i}+b_i^k}}\bm{\upsilon}_{i}^{y_i}\Bigg]\\
&+\tilde{\mathbf{V}}_{i,k}^{y_i}\Bigg[\mathbf{m}_{y_i}^{\backslash i,k} + \mathbf{V}_{y_i}^{\backslash i,k} \frac{\beta_{i,k}}{\sqrt{b_i^{y_i}+b_i^k}}\bm{\upsilon}_{i}^{y_i}\Bigg] - (\mathbf{V}_{y_i}^{\backslash i,k})^{-1}\mathbf{m}_{y_i}^{\backslash i,k}
\end{split}\\
&\begin{split}
= (\mathbf{V}_{y_i}^{\backslash i,k})^{-1}\mathbf{m}_{y_i}^{\backslash i,k} &+ (\mathbf{V}_{y_i}^{\backslash i,k})^{-1}\mathbf{V}_{y_i}^{\backslash i,k} \frac{\beta_{i,k}}{\sqrt{b_i^{y_i}+b_i^k}}\bm{\upsilon}_{i}^{y_i}\\
&+\tilde{\mathbf{V}}_{i,k}^{y_i}\left[\mathbf{m}_{y_i}^{\backslash i,k} + \mathbf{V}_{y_i}^{\backslash i,k} \frac{\beta_{i,k}}{\sqrt{b_i^{y_i}+b_i^k}}\bm{\upsilon}_{i}^{y_i}\right]- (\mathbf{V}_{y_i}^{\backslash i,k})^{-1}\mathbf{m}_{y_i}^{\backslash i,k}
\end{split}\\
&= \frac{\beta_{i,k}}{\sqrt{b_i^{y_i}+b_i^k}}\bm{\upsilon}_{i}^{y_i}
+\tilde{\mathbf{V}}_{i,k}^{y_i}\mathbf{m}_{y_i}^{\backslash i,k} + \tilde{\mathbf{V}}_{i,k}^{y_i}\mathbf{V}_{y_i}^{\backslash i,k} \frac{\beta_{i,k}}{\sqrt{b_i^{y_i}+b_i^k}}\bm{\upsilon}_{i}^{y_i}\\
&= \frac{\beta_{i,k}}{\sqrt{b_i^{y_i}+b_i^k}}\bm{\upsilon}_{i}^{y_i}
+C_{i,k}^{1,y_i}\bm{\upsilon}_{i}^{y_i}(\bm{\upsilon}_{i}^{y_i})^\text{T}\mathbf{m}_{y_i}^{\backslash i,k} +  \frac{\beta_{i,k}}{\sqrt{b_i^{y_i}+b_i^k}}C_{i,k}^{1,y_i}\bm{\upsilon}_{i}^{y_i}(\bm{\upsilon}_{i}^{y_i})^\text{T}\mathbf{V}_{y_i}^{\backslash i,k}\bm{\upsilon}_{i}^{y_i}\\
&= \left[\frac{\beta_{i,k}}{\sqrt{b_i^{y_i}+b_i^k}}
+C_{i,k}^{1,y_i}(\bm{\upsilon}_{i}^{y_i})^\text{T}\mathbf{m}_{y_i}^{\backslash i,k} +  \frac{\beta_{i,k}}{\sqrt{b_i^{y_i}+b_i^k}}C_{i,k}^{1,y_i}(\bm{\upsilon}_{i}^{y_i})^\text{T}\mathbf{V}_{y_i}^{\backslash i,k}\bm{\upsilon}_{i}^{y_i}\right]\bm{\upsilon}_{i}^{y_i}\,,
\end{split}
\end{equation}
where we have used that $(\mathbf{V}_{y_i}^\text{new})^{-1} = \mathbf{V}_{y_i}^{-1}+\tilde{\mathbf{V}}_{i,k}^{y_i}$. If we define $C_{i,k}^{2,y_i}$ and $C_{i,k}^{2}$ as:
{\footnotesize
\begin{align}
&C_{i,k}^{2,y_i} = \left[\frac{\beta_{i,k}}{\sqrt{b_i^{y_i}+b_i^k}}
+C_{i,k}^{1,y_i}(\bm{\upsilon}_{i}^{y_i})^\text{T}\mathbf{m}_{y_i}^{\backslash i,k} +  \frac{\beta_{i,k}}{\sqrt{b_i^{y_i}+b_i^k}}C_{i,k}^{1,y_i}(\bm{\upsilon}_{i}^{y_i})^\text{T}\mathbf{V}_{y_i}^{\backslash i,k}\bm{\upsilon}_{i}^{y_i}\right]\,,\\
&C_{i,k}^{2} = \left[\frac{\beta_{i,k}}{\sqrt{b_i^{y_i}+b_i^k}}
+C_{i,k}^{1}(\bm{\upsilon}_{i}^{k})^\text{T}\mathbf{m}_k^{\backslash i,k} +  \frac{\beta_{i,k}}{\sqrt{b_i^{y_i}+b_i^k}}C_{i,k}^{1}(\bm{\upsilon}_{i}^{k})^\text{T}\mathbf{V}_k^{\backslash i,k}\bm{\upsilon}_{i}^{k}\right]\,,
\end{align}
}we obtain the following expressions for the first natural parameters:
\begin{align}
\tilde{\mathbf{m}}_{i,k}^{y_i}  &= C_{i,k}^{2,y_i}\bm{\upsilon}_{i}^{y_i}\\
\tilde{\mathbf{m}}_{i,k} & = C_{i,k}^{2}\bm{\upsilon}_{i}^{k}\,.
\end{align}

Once we have these parameters we can compute the value of the normalization constant $\tilde{s}_{i,k}$, which guarantees that the approximate factor integrates the same as the exact factor with respect to $q^{\backslash i,k}$. Let $\boldsymbol{\theta}$ be the natural parameters of $q$ after the update and $\boldsymbol{\theta}^{\backslash i,k}$ the natural parameters of the cavity distribution $q^{\backslash i,k}$. Then,
\begin{equation}
\tilde{s}_{i,k} = \log Z_{i,k} + g(\boldsymbol{\theta}^{\backslash i,k}) - g(\boldsymbol{\theta})\,,
\end{equation}
where $g(\boldsymbol{\theta})$ is the log-normalizer of a multivariate Gaussian with natural parameters $\boldsymbol{\theta}$.

\section{Parallel EP updates and damping}

As indicated in the main manuscript, we update all approximate factors in parallel. This means that we compute
all the quantities required for updating each of the approximate factors at once (in particular the quantities derived 
for the cavity distribution $q^{\setminus i,k}$). Parallel updates have also been used in the context of 
multi-class Gaussian process classification in \cite{lobato2011}. These updates are faster than sequential EP updates
because there is no need to introduce a loop over the data. All computations can be carried out in terms of 
matrix vector multiplications that are often more efficient. A disadvantage of parallel updates is, however,
that they may lead to unstable EP updates. We only observed unstable EP updates in the
batch training setting. When using mini-batches, only a few factors are refined at the same time (\emph{i.e.}, 
the ones corresponding to the data instances found in the current mini-batch), so the EP updates are more stable 
in that case. 

To prevent unstable EP updates we used damped EP updates. These simply replace the EP updates of each 
approximate factor with a linear combination of old and new parameters. For example, we set 
$\tilde{C}_{i,k}^1 = (\tilde{C}_{i,k}^1)^\text{new} \rho + (\tilde{C}_{i,k}^1)^\text{old}(1-\rho)$ in the case 
of the $\tilde{C}_{i,k}$ parameter of the approximate factor (we do this with all the parameters). 
In the previous expression $\rho \in [0,1]$ a value that specifies the amount of damping.
If $\rho = 0$ no update happens. If $\rho = 1$ we obtain the original EP update.
Importantly, damping does not change the EP convergence points so it does not affect to the quality of the 
solution.

\section{Estimate of the marginal likelihood}

As we have seen in the main manuscript, the estimate of the log marginal likelihood is:
\begin{align}
\log Z_q &= g(\boldsymbol{\theta}) - g(\boldsymbol{\theta}_\text{prior}) + \sum_{i=1}^{N}\sum_{k\neq y_i}\log\tilde{s}_{i,k}\\
\log\tilde{s}_{i,k} &= \log Z_{i,k} + g(\boldsymbol{\theta}^{\backslash i,k})-g(\boldsymbol{\theta})\,,
\end{align}
where $\boldsymbol{\theta}$, $\boldsymbol{\theta}^{\backslash i,k}$ and $\boldsymbol{\theta}_\text{prior}$ are the natural parameters of $q$, $q^{\backslash i,k}$ and $p(\overline{\mathbf{f}})$ respectively and $g(\boldsymbol{\theta}')$ is the log-normalizer of a multivariate Gaussian with natural parameters $\boldsymbol{\theta}'$. If $\boldsymbol{\mu}$ and $\boldsymbol{\Sigma}$ are the mean and covariance matrix of 
that Gaussian distribution over $D$ dimensions, then
\begin{equation}
g(\boldsymbol{\theta}') = \frac{D}{2}\log 2\pi + \frac{1}{2}\log |\boldsymbol{\Sigma}| + \frac{1}{2} \boldsymbol{\mu}^\text{T}\boldsymbol{\Sigma}^{-1}\boldsymbol{\mu}\,,
\end{equation}
which leads to
\begin{equation}
\log Z_q = \sum_{k=1}^{C}\frac{1}{2}\log|\mathbf{V}_k| + \frac{1}{2}\mathbf{m}_k^\text{T}\mathbf{V}_k^{-1}\mathbf{m}_k - \frac{1}{2}|\mathbf{K}_{\overline{\mathbf{X}}^k \overline{\mathbf{X}}^k}^k| + \sum_{i=1}^{N}\sum_{k\neq y_i}\log\tilde{s}_{i,k}\,,
\end{equation}
with
\begin{align}
\log \tilde{s}_{i,k}  &=  \log Z_{i,k} + \frac{1}{2} \log \left|\mathbf{V}_{y_i}^{\setminus i,k}\right| 
	+ \frac{1}{2} \left(\mathbf{m}_{y_i}^{\setminus i,k}\right)^\text{T} \left(\mathbf{V}_{y_i}^{\setminus i,k}\right)^{-1}
	\mathbf{m}_{y_i}^{\setminus i,k}
\nonumber \\
& \quad - \frac{1}{2} \log \left|\mathbf{V}_{y_i}^{\setminus i,k}\right| 
	- \frac{1}{2} \left(\mathbf{m}_{y_i}^{\setminus i,k}\right)^\text{T} \left(\mathbf{V}_{y_i}^{\setminus i,k}\right)^{-1}
	\mathbf{m}_{y_i}^{\setminus i,k}\,.
\end{align}
This expression can be evaluated very efficiently using
the Woodbury matrix identity; the matrix determinant lemma; 
that $(\mathbf{V}_k^{\backslash i,k})^{-1} = \mathbf{V}_k^{-1} - \tilde{\mathbf{V}}_{i,k}$, 
and $(\mathbf{V}_{y_i}^{\backslash i,k})^{-1} = \mathbf{V}_{y_i}^{-1} - \tilde{\mathbf{V}}^{y_i}_{i,k}$;
that $\mathbf{m}_k^{\backslash i,k} = \mathbf{V}_k^{\backslash i,k}(\mathbf{V}_k^{-1}\mathbf{m}_k - \tilde{\mathbf{m}}_{i,k})$, 
and $\mathbf{m}_{y_i}^{\backslash i,k} = \mathbf{V}_{y_i}^{\backslash i,k}(\mathbf{V}_{y_i}^{-1}\mathbf{m}_{y_i} - 
\tilde{\mathbf{m}}^{y_i}_{i,k})$; and the special form of the parameters of the approximate 
factors $\tilde{\mathbf{V}}_{i,k}$, $\tilde{\mathbf{m}}_{i,k}$, $\tilde{\mathbf{V}}^{y_i}_{i,k}$ and $\tilde{\mathbf{m}}^{y_i}_{i,k}$.

\section{Gradient of $\boldsymbol{\log Z_q}$ after convergence and learning rate}

We derive the expression for the gradient of $\log Z_q$ after EP has converged. Let denote $\xi_j^k$ to one 
hyper-parameter of the model (\emph{i.e.}, a parameter of one of the covariance functions or a component of the inducing points) 
and $\boldsymbol{\theta}$ and $\boldsymbol{\theta}_\text{prior}$ to the natural parameters 
of $q$ and $p(\overline{\mathbf{f}})$ respectively. When EP has converged, the approximate factors can be 
considered to be fixed (it does not change with the model hyper-parameters) \cite{matthias2006}. In this case, it 
is only necessary to consider the direct dependency of $\log Z_{i,k}$ on $\xi_j^k$ \cite{matthias2006}. Then, the gradient is given by:
\begin{equation}
\small
\begin{split}
\frac{\partial \log Z_q}{\partial \xi_j^k} = &\left(\frac{\partial g(\boldsymbol{\theta})}{\partial\boldsymbol{\theta}}\right)^\text{T} \frac{\partial \boldsymbol{\theta}}{\partial\xi_j^k} - \left(\frac{\partial g(\boldsymbol{\theta}_\text{prior})}{\partial\boldsymbol{\theta}_\text{prior}}\right)^\text{T} \frac{\partial \boldsymbol{\theta}_\text{prior}}{\partial\xi_j^k} + \sum_{i=1}^{N}\sum_{k\neq y_i} \frac{\partial\log 
	\tilde{s}_{i,k}}{\partial\xi_j^k} \\
&= \boldsymbol{\eta}^\text{T} \frac{\partial \boldsymbol{\theta}}{\partial\xi_j^k} - \left(\boldsymbol{\eta}_\text{prior}\right)^\text{T} \frac{\partial \boldsymbol{\theta}_\text{prior}}{\partial\xi_j^k} + \sum_{i=1}^{N}\sum_{k\neq y_i} \frac{\partial\log \tilde{s}_{i,k}}{\partial\xi_j^k}\\
&= \boldsymbol{\eta}^\text{T} \frac{\partial \boldsymbol{\theta}_\text{prior}}{\partial\xi_j^k} - \left(\boldsymbol{\eta}_\text{prior}\right)^\text{T} \frac{\partial \boldsymbol{\theta}_\text{prior}}{\partial\xi_j^k} + \sum_{i=1}^{N}\sum_{k\neq y_i} \frac{\partial\log Z_{i,k}}{\partial\xi_j^k}\\
&= \left(\boldsymbol{\eta}^\text{T} - \boldsymbol{\eta}_\text{prior}^\text{T}\right) \frac{\partial \boldsymbol{\theta}_\text{prior}}{\partial\xi_j^k}  + \sum_{i=1}^{N}\sum_{k\neq y_i} \frac{\partial\log Z_{i,k}}{\partial\xi_j^k}\,,
\end{split}
\normalsize
\end{equation}
where we have used the chain rule of matrix derivatives \cite{petersen_matrix_2012}, the especial form of the derivatives when using inducing points \cite{SnelsonThesis07} and that $\boldsymbol{\theta} = \boldsymbol{\theta}_\text{prior} + \sum_{i=1}^{N}\sum_{k\neq y_i}\boldsymbol{\theta}_i^c$, with $\boldsymbol{\theta}_i^c$ the natural parameters of the approximate factor $\tilde{\phi}_i^k$. 
Furthermore, $\bm{\eta}$ and $\bm{\eta}_\text{prior}$ are expected sufficient statistics under the posterior approximation $q$ and
the prior, respectively. This gradient coincides with the one in the main manuscript.

It is important to note that one has to use the chain rule of matrix derivatives when trying to use the previous expression
to compute the gradient. In particular, natural parameters and expected sufficient statistics are expressed in the form of
matrices. Thus, one has to use in practice the chain rule of matrix derivatives, as indicated in \cite{petersen_matrix_2012}:
\begin{align}
(\bm{\eta} - \bm{\eta}_\text{prior})^\text{T} \frac{\bm{\theta}_\text{prior}}{\partial \xi_j^k} &= 
	- \frac{1}{2} \text{trace} \left(\mathbf{M}_k^\text{T} 
	\frac{\partial \mathbf{K}^k_{\overline{\mathbf{X}}^k\overline{\mathbf{X}}^k}}{\partial \xi_j^k} \right)
\end{align}
where 
\begin{align}
	\mathbf{M}_k &= \left(\mathbf{K}^k_{\overline{\mathbf{X}}^k\overline{\mathbf{X}}^k}\right)^{-1} -
		\left(\mathbf{K}^k_{\overline{\mathbf{X}}^k\overline{\mathbf{X}}^k}\right)^{-1} 
		\mathbf{V}_k
		\left(\mathbf{K}^k_{\overline{\mathbf{X}}^k\overline{\mathbf{X}}^k}\right)^{-1}	-
                \left(\mathbf{K}^k_{\overline{\mathbf{X}}^k\overline{\mathbf{X}}^k}\right)^{-1} 
		\mathbf{m}_k
		\mathbf{m}_k^\text{T}
                \left(\mathbf{K}^k_{\overline{\mathbf{X}}^k\overline{\mathbf{X}}^k}\right)^{-1}
		\,,
\end{align}
where $\mathbf{V}_k$ and $\mathbf{m}_k$ are the covariance matrix and mean vector of the $k$-th component of $q$.
Furthermore, several standard properties of the trace can be employed to simplify the computations. 
In particular, the trace is invariant to cyclic rotations. Namely, $\text{trace}(\mathbf{ABCD}) = \text{trace}(\mathbf{DABC})$.
The derivatives with respect to each $\log Z_{i,k}$ can be computed also efficiently using the chain rule for
matrix derivatives. 

In our experiments we use an adaptive learning rate for the batch methods. This learning rate 
is different for each hyper-parameter. The rule that we use is to increase the learning rate 
by 2\% if the sign of the estimate of the gradient for that hyper-parameter does not change 
between two consecutive iterations. If a change is observed, we multiply the learning rate by 
$1/2$. When applying stochastic optimization methods, we use the ADAM method with the default settings
to estimate the learning rate \cite{kingma2015}.

\section{Predictive distribution}

Once the training has completed, we can use the posterior approximation to make predictions for new instances. For that, we first compute an approximate posterior evaluated at the location of the new instance $\mathbf{x^\star}$, denoted by 
$\mathbf{f}^\star=(f^1(\mathbf{x}^\star),\ldots,f^C(\mathbf{x}^\star))^\text{T}$:
\begin{align}
p(\mathbf{f}^\star|\mathbf{y}) & = \int p(\mathbf{f}^\star|\overline{\mathbf{f}}) p(\overline{\mathbf{f}}|\mathbf{y}) d\overline{\mathbf{f}} = \int p(\mathbf{f}^\star|\overline{\mathbf{f}}) q(\overline{\mathbf{f}}) d\overline{\mathbf{f}}
\approx \prod_{k=1}^{C}\mathcal{N}(f^k(\mathbf{x}^\star)|m_k^\star, v_k^\star)\,,
\end{align}
where: 
\begin{align}
m_k^\star &= (\mathbf{k}_{\mathbf{x}^\star, \overline{\mathbf{X}}^k}^k)^\text{T}(\mathbf{K}_{\overline{\mathbf{X}}^k \overline{\mathbf{X}}^k}^k)^{-1}\mathbf{m}_k\\
v_k^\star &= \kappa^k_{\mathbf{x}^\star,\mathbf{x}^\star} - (\mathbf{k}_{\mathbf{x}^\star, \overline{\mathbf{X}}^k}^k)^\text{T}(\mathbf{K}_{\overline{\mathbf{X}}^k \overline{\mathbf{X}}^k}^k)^{-1}\mathbf{k}_{\mathbf{x}^\star, \overline{\mathbf{X}}^k}^k + (\mathbf{k}_{\mathbf{x}^\star, \overline{\mathbf{X}}^k}^k)^\text{T}(\mathbf{K}_{\overline{\mathbf{X}}^k \overline{\mathbf{X}}^k}^k)^{-1}\mathbf{V}_k(\mathbf{K}_{\overline{\mathbf{X}}^k \overline{\mathbf{X}}^k}^k)^{-1}\mathbf{k}_{\mathbf{x}^\star, \overline{\mathbf{X}}^k}^k\,.
\end{align}

This approximate posterior can be used to obtain an approximate predictive distribution for the class label $y^\star$:
\begin{equation}
\small
\begin{split}
p(y^\star | \mathbf{x^\star}, \mathbf{y}) &= \int p(y^\star|\mathbf{x^\star}, \mathbf{f}^\star) p(\mathbf{f}^\star|\mathbf{y})d\mathbf{f}^\star \\
&= \int p(y^\star|\mathbf{x^\star}, \mathbf{f}^\star) \prod_{k=1}^{C}\mathcal{N}(f^k(\mathbf{x}^\star)|\mathbf{m}_k^\star, \mathbf{V}_k^\star) d\mathbf{f}^\star\\
&= \int \left[\prod_{k\neq y^\star} \Theta\left(f^{y^\star}(\mathbf{x}^\star) - f^c(\mathbf{x}^\star) \right)
	\right] \prod_{k=1}^{C}\mathcal{N}(f^k(\mathbf{x}^\star)|m_k^\star, v_k^\star) d\mathbf{f}^\star\\
&=\int \left[\prod_{k\neq y^\star} \Theta\left(f^{y^\star}(\mathbf{x}^\star) - f^k(\mathbf{x}^\star)\right)\right] 
	\prod_{k\neq y^\star}\mathcal{N}(f^k(\mathbf{x}^\star)|\mathbf{m}_k^\star, \mathbf{V}_k^\star) d\mathbf{f}^\star 
	\mathcal{N}(f^{y^\star}(\mathbf{x}^\star)|\mathbf{m}_{y^\star}^\star, \mathbf{V}_{y^\star}^\star)\\
&= \int \prod_{k\neq y^\star} \Phi\left(\frac{f^{y^\star}(\mathbf{x}^\star) - \mathbf{m}_k^\star}
	{\sqrt{{v}_k^\star}}\right)  \mathcal{N}(f^{y^\star}(\mathbf{x}^\star)|{m}_{y^\star}^\star, {v}_{y^\star}^\star)
	df^{y^\star}(\mathbf{x}^\star)\,,
\end{split}
\normalsize
\end{equation}
where $\Phi(\cdot)$ is the cumulative distribution function of a Gaussian distribution. This is 
an integral in one dimension and can easily be approximated by quadrature techniques.

\section{Additional Experimental Results}

\begin{figure}[H]
	\begin{center}
		\centerline{\includegraphics[width=0.6\columnwidth]{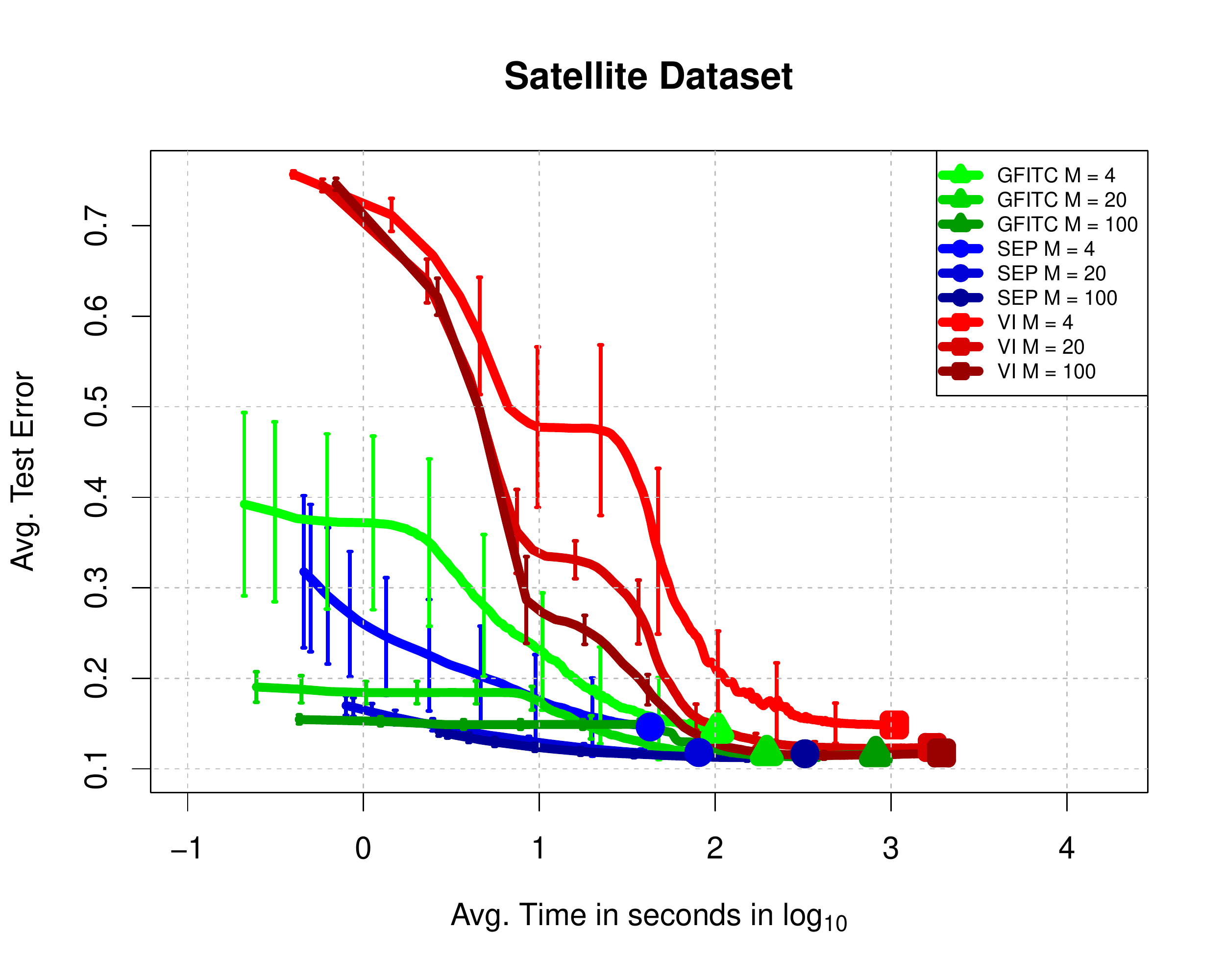}}
		\caption{{\small Test error for GFITC, SEP and VI on \emph{Satellite} as a function
				of the training time. Best seen in color.}} 
		\label{fig:err_time}
	\end{center}
\end{figure}

In this section we show additional results that did not fit in the main manuscript. 
Figure \ref{fig:err_time} shows the test error of each method as a function of 
the training time on the Satellite dataset from the UCI repository. This figure shows 
similar results to the ones observed in the main manuscript in terms of the negative 
log-likelihood. 

Table \ref{tab:err_uci} shows the average test error for all the 
methods on the UCI repository datasets. The obtained results are 
similar for all the methods in terms of the test error.
Namely, EP and SEP perform similarly to GFITC. Importantly, the prediction 
error of VI is not that bad as in terms of the test log-likelihood.  It is in fact similar 
to the one of the other methods. This gives evidence supporting that VI can provide good prediction errors, 
but it can fail to accurately model the predictive distribution.

\begin{table}[H]
	\begin{center}
		\caption{ Average test error for each method and average training time in seconds on UCI repository datasets.}
		\label{tab:err_uci}
		{\small
			\begin{tabular}{@{\hspace{0mm}}l@{\hspace{.5mm}}|@{\hspace{.1mm}}l@{\hspace{.1mm}}
					|@{\hspace{.1mm}}c@{{\tiny $\pm$}}c@{\hspace{.1mm}}
					|@{\hspace{.1mm}}c@{{\tiny $\pm$}}c@{\hspace{.1mm}}
					|@{\hspace{.1mm}}c@{{\tiny $\pm$}}c@{\hspace{.1mm}}
					|@{\hspace{.1mm}}c@{{\tiny $\pm$}}c@{\hspace{.1mm}}
				}
				\hline
				& \bf{Problem}  & \multicolumn{2}{c}{\scriptsize \bf GFITC} & \multicolumn{2}{c}{\bf EP} &\multicolumn{2}{c}{\bf SEP} & \multicolumn{2}{c}{\bf VI}  \\
				\hline
				\parbox[t]{2mm}{\multirow{8}{*}{\rotatebox[origin=c]{90}{{\scriptsize ${\bf M = 5\%}$}}}}
				& Glass  & \bf{ 0.23 } & \bf{ 0.02 } & 0.31 & 0.02 & 0.31 & 0.02 & 0.35 & 0.02 \\
				& New-thyroid  & \bf{ 0.02 } & \bf{ 0.01 } & 0.04 & 0.01 & 0.02 & 0.01 & 0.03 & 0.01 \\
				& Satellite  & 0.12 & 0 & \bf{ 0.11 } & \bf{ 0 } & 0.12 & 0 & 0.12 & 0 \\
				& Svmguide2  & 0.2 & 0.01 & 0.2 & 0.01 & 0.2 & 0.02 & \bf{ 0.19 } & \bf{ 0.01 } \\
				& Vehicle  & 0.17 & 0.01 & 0.17 & 0.01 & \bf{ 0.16 } & \bf{ 0.01 } & 0.17 & 0.01 \\
				& Vowel  & \bf{ 0.05 } & \bf{ 0.01 } & 0.09 & 0.01 & 0.09 & 0.01 & 0.06 & 0.01 \\
				& Waveform  & 0.17 & 0 & \bf{ 0.15 } & \bf{ 0 } & 0.16 & 0 & 0.17 & 0 \\
				& Wine  & 0.03 & 0.01 & \bf{ 0.03 } & \bf{ 0.01 } & 0.03 & 0.01 & 0.04 & 0.01 \\
				\hline
				& {\bf Avg. Time} & 131 & 3.11 & 53.8 & 0.19 & {\bf 48.5} & {\bf 0.97} & 157 & 0.59 \\
				\hline
				\parbox[t]{2mm}{\multirow{8}{*}{\rotatebox[origin=c]{90}{{\scriptsize ${\bf M = 10\%}$}}}}
				& Glass  & \bf{ 0.2 } & \bf{ 0.01 } & 0.29 & 0.02 & 0.3 & 0.02 & 0.35 & 0.02 \\
				& New-thyroid  & 0.03 & 0.01 & \bf{ 0.02 } & \bf{ 0.01 } & 0.03 & 0.01 & 0.03 & 0.01 \\
				& Satellite  & 0.11 & 0 & \bf{ 0.11 } & \bf{ 0 } & 0.12 & 0 & 0.12 & 0 \\
				& Svmguide2  & 0.19 & 0.02 & 0.2 & 0.02 & 0.2 & 0.02 & \bf{ 0.17 } & \bf{ 0.02 } \\
				& Vehicle  & 0.17 & 0.01 & 0.16 & 0.01 & 0.16 & 0.01 & \bf{ 0.15 } & \bf{ 0.01 } \\
				& Vowel  & \bf{ 0.03 } & \bf{ 0.01 } & 0.05 & 0.01 & 0.06 & 0.01 & 0.06 & 0 \\
				& Waveform  & 0.17 & 0 & \bf{ 0.16 } & \bf{ 0 } & 0.16 & 0 & 0.18 & 0 \\
				& Wine  & 0.04 & 0.01 & \bf{ 0.02 } & \bf{ 0.01 } & 0.03 & 0.01 & 0.03 & 0.01 \\
				\hline
				& {\bf Avg. Time} & 264 & 6.91 & 102 & 0.64 & {\bf 96.6} & {\bf 1.99} & 179 & 0.78 \\
				\hline
				\parbox[t]{2mm}{\multirow{8}{*}{\rotatebox[origin=c]{90}{{\scriptsize ${\bf M = 20\%}$}}}}
				& Glass  & \bf{ 0.2 } & \bf{ 0.02 } & 0.28 & 0.02 & 0.28 & 0.02 & 0.36 & 0.02 \\
				& New-thyroid  & 0.03 & 0 & 0.02 & 0.01 & \bf{ 0.02 } & \bf{ 0.01 } & 0.03 & 0.01 \\
				& Satellite  & 0.11 & 0 & \bf{ 0.11 } & \bf{ 0 } & 0.12 & 0 & 0.11 & 0 \\
				& Svmguide2  & 0.2 & 0.01 & 0.19 & 0.01 & 0.2 & 0.02 & \bf{ 0.19 } & \bf{ 0.02 } \\
				& Vehicle  & 0.17 & 0.01 & 0.16 & 0.01 & 0.16 & 0.01 & \bf{ 0.15 } & \bf{ 0.01 } \\
				& Vowel  & \bf{ 0.03 } & \bf{ 0 } & 0.03 & 0 & 0.05 & 0.01 & 0.03 & 0.01 \\
				& Waveform  & 0.17 & 0 & \bf{ 0.16 } & \bf{ 0 } & 0.17 & 0 & 0.18 & 0 \\
				& Wine  & 0.04 & 0.01 & \bf{ 0.01 } & \bf{ 0.01 } & 0.03 & 0.01 & 0.03 & 0.01 \\
				\hline
				& {\bf Avg. Time} & 683 & 17.3 & 228 & 0.78 & {\bf 216} & {\bf 2.88} & 248 & 0.66 \\
				\hline
			\end{tabular}
		}
	\end{center}
\end{table}

In this section we also give further evidence supporting that the VI method can sometimes fail 
to successfully provide accurate predictive distributions. For this, consider a test instance 
$(\mathbf{x}^\star,y^\star)$. We record on
airline dataset, for the VI method,  the average value of $\log \mathds{E}[p(y^\star|\mathbf{f}^\star)]$ on the test set with
respect to the training time (\emph{i.e.}, 
this is the  test log-likelihood). Similarly, we also record the average value of $\mathds{E}[\log p(y^\star|\mathbf{f}^\star)]$.
This is the quantity that appears in the lower bound that is maximized by the VI method, although evaluated on the
training data (see the main manuscript). Note that the second expression is a lower bound on the first one by using Jensen's inequality.

\begin{figure}[H]
	\begin{center}
		\centerline{\includegraphics[width=0.6\columnwidth]{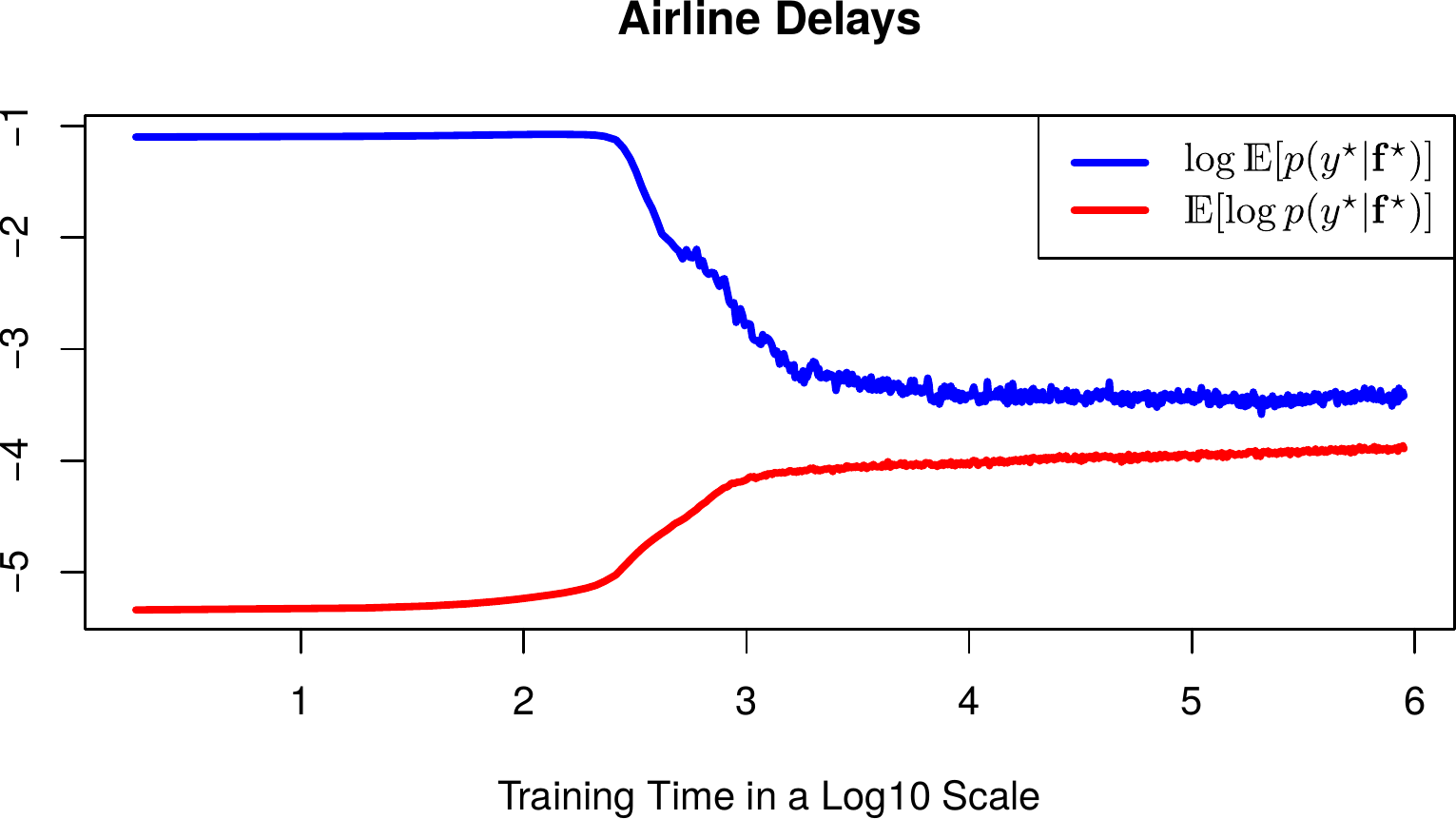}}
		\caption{{\small Average value of $\log \mathds{E}[p(y^\star|\mathbf{f}^\star)]$ for each test instance
			$(\mathbf{x}^\star,y^\star)$ alongside with the average value of $\mathds{E}[\log p(y^\star|\mathbf{f}^\star)]$. The 
			results plotted correspond to the the VI method.
			}} 
		\label{fig:vi_airline}
	\end{center}
\end{figure}

The results obtained are displayed in Figure \ref{fig:vi_airline}. In principle, the main idea behind the VI
method is that maximizing on the training data $\mathds{E}[\log p(y^\star|\mathbf{f}^\star)]$ should also 
increase $\log \mathds{E}[p(y^\star|\mathbf{f}^\star)]$ as a consequence of the inequality 
$\log \mathds{E}[p(y^\star|\mathbf{f}^\star)] \geq \mathds{E}[\log p(y^\star|\mathbf{f}^\star)]$. However, as it can be observed in Figure 
\ref{fig:vi_airline}, while the VI method successfully maximizes $\log \mathds{E}[p(y^\star|\mathbf{f}^\star)]$ on 
test data not seen by the model, the corresponding value of 
$\mathds{E}[\log p(y^\star|\mathbf{f}^\star)]$ decreases. Thus, we conclude 
that if one cares about the predictive test log-likelihood, the VI objective may not be 
the best one to use, as it is only expected to maximize $\log \mathds{E}[p(y^\star|\mathbf{f}^\star)]$ in an indirect
way, by maximizing in practice a lower bound on this quantity. As pointed out 
by Figure \ref{fig:vi_airline}, sometimes this procedure can fail and in fact 
lead to a decrease of $\log \mathds{E}[p(y^\star|\mathbf{f}^\star)]$.

